\documentclass[lettersize,conference,compsoc]{IEEEtran}
\usepackage{amsmath,amsfonts}
\usepackage{algorithmic}
\usepackage{algorithm}
\usepackage{array}
\usepackage[caption=false,font=normalsize,labelfont=sf,textfont=sf]{subfig}
\usepackage{textcomp}
\usepackage{stfloats}
\usepackage{url}
\usepackage{svg}
\usepackage{verbatim}
\usepackage{graphicx}
\usepackage{cite}
\usepackage{xurl}
\usepackage{comment}

\usepackage{color}
\usepackage{transparent}
\usepackage{import}  % Optional but usefu
\usepackage{tabularx}
\newcolumntype{Y}{>{\raggedright\arraybackslash}X}
\usepackage{booktabs}

\usepackage{graphicx}

\usepackage[english]{babel}
\addto\extrasenglish{
  
}

\usepackage{fancyhdr} %номера страниц
\usepackage{hyperref} %кликабельные ссылки

\usepackage{tikz}
\usetikzlibrary{positioning, arrows.meta, calc, fit, backgrounds, decorations.pathreplacing}

\newcounter{challenge}
\begin{document}

\title{From Multi-Agent Systems and the Semantic Web to Agentic AI:\\ A Unified Narrative of the Web of Agents}

%\author{%
%  Tatiana Petrova,~\IEEEmembership{Member,~IEEE,}
%  Boris Bliznukov,~\IEEEmembership{Member,~IEEE,}
%  Aleksandr Puzikov, ~\IEEEmembership{Member,~IEEE,}\\
%  Radu State,~\IEEEmembership{Fellow,~IEEE}%
%\thanks{This paper was produced by the IEEE Publication Technology %Group. They are in Piscataway, NJ.}% <-this % stops a space
%\thanks{Manuscript received April 19, 2021; revised August 16, 2021.}}

\author{%
\begin{tabular}{@{}c@{\hspace{3em}}c@{}}
 Tatiana Petrova & Boris Bliznioukov\\
 \textit{SEDAN - SnT} & \textit{SEDAN - SnT}\\
 University of Luxembourg & University of Luxembourg\\
 Luxembourg, Luxembourg & Luxembourg, Luxembourg\\
 tatiana.petrova@uni.lu & boris.bliznioukov@uni.lu\\[1em]
 Aleksandr Puzikov & Radu State\\
 \textit{SEDAN - SnT} & \textit{SEDAN - SnT}\\
 University of Luxembourg & University of Luxembourg\\
 Luxembourg, Luxembourg & Luxembourg, Luxembourg\\
 aleksandr.puzikov@uni.lu & radu.state@uni.lu
\end{tabular}%
}

\markboth{IEEE Access, 2026}%
{Petrova \MakeLowercase{\textit{et al.}}: From Multi-Agent Systems and the Semantic Web to Agentic AI}


\maketitle

% Highlights block (Information Fusion standard; harmless for IEEE Access).
% Each bullet kept to <=85 characters per Elsevier guideline.
{\setlength{\parindent}{0pt}\noindent\textbf{\large Highlights}\\[0.2em]
\begingroup\small
\begin{itemize}\setlength{\itemsep}{0pt}\setlength{\parskip}{0pt}
  \item Three-decade narrative unifies Multi-Agent Systems, Semantic Web, and LLM agents.
  \item Four-dimensional framework applied to sixteen representative systems (1995--2026).
  \item Gen~III publication share several-fold above Gen~II peak in 2026 (partial year).
  \item Seven named lessons paired with seven generation-invariant challenges.
  \item To our knowledge first to integrate the Nov~2024--Aug~2026 institutional layer.
\end{itemize}
\endgroup
\vspace{0.6em}
}

\begin{abstract}
The Web of Agents (WoA) transforms the document-centric Web into an environment of autonomous agents acting on users' behalf, a vision newly tractable as large language models (LLMs) mature. We argue that across three decades the WoA has undergone a \emph{semantic-effort migration} in chronological order: from platform-side coordination (Multi-Agent Systems, Generation~I), through data-side annotation (Semantic Web, Generation~II), to model-side interpretation (LLM-era, Generation~III). The central Gen~II~$\rightarrow$~Gen~III transition within this trajectory, which we call the \emph{semantics-in-data $\rightarrow$ semantics-in-models} shift, is predictive: each generation's failure modes and current open problems follow from where that generation located its semantic effort. The survey makes five contributions: (i)~a unified evolutionary narrative spanning 1990--2026; (ii)~a four-dimensional comparative framework (semantic foundation, communication paradigm, locus of intelligence, discovery mechanism) applied uniformly across all three generations; (iii)~classification of sixteen representative systems on these dimensions, including hybrid LLM--knowledge-graph and computer-use agents; (iv)~coverage of the November~2024--August~2026 institutional convergence (Linux Foundation's Agentic AI Foundation, A2A v1.0, MCP November~2024 launch and November~2025 specification, Visa/Mastercard/Stripe payment-network protocols, EU AI Act phased enforcement, the NIST AI Agent Standards Initiative, International AI Safety Report 2026); and (v)~seven named lessons grounded in cross-generational evidence paired with seven generation-invariant challenges that persist regardless of which protocol prevails. Further progress depends less on protocol design than on the socio-technical infrastructure now being assembled by standards bodies, regulators, and commercial payment networks.
\end{abstract}


\section{Introduction}\label{sec:introduction}

\IEEEPARstart{T}{he} vision of a global digital ecosystem populated by autonomous software agents, capable of discovering one another, communicating, and collaborating to perform complex tasks on behalf of users, is one of the long-standing ambitions in artificial intelligence. This concept, often termed the ``Web of Agents'' (a label whose earliest documented use we trace to Yao~2005, \S\ref{sec:naming}), envisions an evolution of the internet from a repository of human-readable information into a dynamic environment for machine-to-machine cooperation. Early proponents imagined distributed programs that could traverse a machine-readable web, providing novel capabilities with minimal human intervention (a ``science fiction'' vision that has guided research for decades).

As Figure~\ref{fig:timeline-3gen} traces the chronology of agent research across three decades and Figure~\ref{fig:pub-volume} quantifies the associated publication volume, the foundational pillars (``Intelligent Agent'' and ``Multi-Agent System'' (MAS)) have been central to computer science, commanding sustained and substantial research interest for over two decades, even as the specific term ``Web of Agents'' has seen limited traction. This discrepancy suggests that while the grand vision of a fully interconnected agent web remained elusive, the community has been building its constituent parts. The early enthusiasm for the ``Semantic Web'', intended to provide the machine-readable data layer for this vision, peaked and declined, indicating that a semantically rich web was a necessary but insufficient condition for its success. Our analysis reveals that the primary bottleneck was not the network, but the nodes: individual agents lacked the general-purpose reasoning and adaptive capabilities to create the value-generating applications needed to drive adoption of complex interoperability standards.

Recent advances have changed this picture. The advent of Large Language Models (LLMs) has served as a catalyst, supporting a new generation of capable agents. Models from OpenAI (GPT-4.5 \cite{OpenAI_GPT45_2025}), Google (Gemini 2.5 Pro \cite{Google_Gemini25Pro_2025}), Anthropic (Claude 3.7 Sonnet \cite{Anthropic_Claude37Sonnet_2025}), xAI (Grok-3 \cite{xAI_Grok3_2025}), Mistral (Large~2 \cite{Mistral_Large2_2024}), and Alibaba (Qwen~3 \cite{Alibaba_Qwen3_2025}) demonstrate strong capabilities in synthesizing information and generating coherent plans of action, often exhibiting emergent problem-solving skills (capabilities that are not present in smaller-scale models but arise in larger-scale models, and thus cannot be predicted by simply extrapolating the performance of their smaller predecessors \cite{wei_2022}). Recent research posits that this apparent reasoning may stem from an ``illusion of thinking,'' where models adeptly retrieve and adapt solution templates from their vast training data \cite{shojaee_2025illusion}. This mechanism, combined with strong language understanding, is what allows modern agents to interpret and manipulate web content with a flexibility previously reserved for human users.

\begin{figure*}[!t]
\centering
% =========================================================================
% UPPER PANEL: Main timeline 1995-2025 (3 generations)
% =========================================================================
\begin{tikzpicture}[
  xscale=0.45, yscale=0.85,
  font=\small,
  geni/.style={color=blue!45!black!75},
  genii/.style={color=orange!70!brown!85},
  geniii/.style={color=teal!75!black},
  event/.style={circle, draw, fill=white, inner sep=1.2pt, line width=0.7pt},
  evlabel/.style={font=\scriptsize, align=center, inner sep=1pt},
  arrow/.style={-{Stealth[length=2.5mm]}, line width=1pt}
]
% X-coordinate = year - 1995

% Shaded active periods (match body Lesson 4 and Fig 4 generation panels: Gen I 1995-2005, Gen II 2001-2012, Gen III 2020--)
\fill[blue!45!black!8]    (0,3.8)  rectangle (10,4.8);   % Gen I 1995-2005
\fill[orange!70!brown!10] (6,2.3)  rectangle (17,3.3);   % Gen II 2001-2012
\fill[teal!75!black!8]    (25,0.8) rectangle (30,1.8);   % Gen III 2020-2025

% Lane arrows
\draw[arrow, geni]   (0,4.3)  -- (30,4.3);
\draw[arrow, genii]  (6,2.8)  -- (30,2.8);
\draw[arrow, geniii] (22,1.3) -- (30,1.3);

% Lane labels (compact, shifted FURTHER LEFT to clear first event markers)
\node[geni,   font=\bfseries\footnotesize, anchor=east, align=right] at (-2.5,4.3)
  {Gen~I\\\textit{\scriptsize MAS}\\\textit{\scriptsize Platform}};
\node[genii,  font=\bfseries\footnotesize, anchor=east, align=right] at (-2.5,2.8)
  {Gen~II\\\textit{\scriptsize Sem Web}\\\textit{\scriptsize Data}};
\node[geniii, font=\bfseries\footnotesize, anchor=east, align=right] at (-2.5,1.3)
  {Gen~III\\\textit{\scriptsize LLM agents}\\\textit{\scriptsize Model}};

% ----- Gen I events -----
% KQML at chart start: actual first publications 1992-93 (Finin et al.); shown at
% the 1995 mark as a symbolic anchor for the MAS era. Annotate the year explicitly.
\node[event, geni] (e1) at (0,4.3) {};
\node[evlabel, geni, anchor=south] at (0,4.45) {KQML\\\textit{\scriptsize 1993}};
\node[event, geni] (e2) at (1,4.3) {};
\node[evlabel, geni, anchor=north] at (1,4.15) {FIPA\\founded};
\node[event, geni] (e3) at (2,4.3) {};   % FIPA ACL (FIPA 97 v1.0): 1997
\node[evlabel, geni, anchor=south] at (2,4.45) {FIPA ACL};
\node[event, geni] (e4) at (4,4.3) {};   % JADE first release: 1999
\node[evlabel, geni, anchor=north] at (4,4.15) {JADE};
\node[event, geni] (e5) at (6,4.3) {};   % Agentcities: EU IST project starting 2001
\node[evlabel, geni, anchor=south] at (6,4.45) {Agentcities};

% ----- Gen II events -----
\node[event, genii] (g1) at (6,2.8) {};
\node[evlabel, genii, anchor=south] at (6,2.95) {Sem-Web\\vision};
\node[event, genii] (g2) at (9,2.8) {};
\node[evlabel, genii, anchor=north] at (9,2.65) {RDF/OWL};
\node[event, genii] (g3) at (11,2.8) {};
\node[evlabel, genii, anchor=south] at (11,2.95) {Shadbolt\\revisited};
\node[event, genii] (g4) at (12,2.8) {};
\node[evlabel, genii, anchor=north] at (12,2.65) {DBpedia};
\node[event, genii] (g5) at (16,2.8) {};
\node[evlabel, genii, anchor=south] at (16,2.95) {Schema.org};
\node[event, genii] (g6) at (17,2.8) {};   % Wikidata launched Oct 2012
\node[evlabel, genii, anchor=north] at (17,2.65) {Wikidata};

% ----- Gen III events (up to Nov 2024) -----
\node[event, geniii] (h1) at (22,1.3) {};
\node[evlabel, geniii, anchor=south] at (22,1.55) {Transformer};
\node[event, geniii] (h2) at (25,1.3) {};
\node[evlabel, geniii, anchor=south] at (25,1.55) {GPT-3};
% ChatGPT (Nov 2022) and AutoGPT (Mar 2023) are close in time:
% leader-lines disambiguate the labels so the markers don't collide visually
\node[event, geniii] (h3) at (27.85,1.3) {};
\draw[geniii, line width=0.3pt] (27.85,1.3) -- (27.4,0.55);
\node[evlabel, geniii, anchor=north] at (27.4,0.55) {ChatGPT};
\node[event, geniii] (h4) at (28.2,1.3) {};
\draw[geniii, line width=0.3pt] (28.2,1.3) -- (28.4,1.85);
\node[evlabel, geniii, anchor=south] at (28.4,1.85) {AutoGPT};
\node[event, geniii] (h5) at (29.9,1.3) {};
\node[evlabel, geniii, anchor=north] at (29.9,1.05) {MCP (Nov\,24)};

% Zoom-in marker on Gen III after MCP
\draw[gray!50, dashed] (29.9,1.3) -- (30.5,1.3);
\node[gray!70, font=\scriptsize\itshape, anchor=west, align=left] at (30.5,1.3) {detail\\below $\downarrow$};

% Year axis at bottom
\foreach \y/\xc in {1995/0, 2000/5, 2005/10, 2010/15, 2015/20, 2020/25, 2025/30} {
  \draw[gray!60] (\xc,-0.1) -- (\xc,-0.25);
  \node[gray!60, font=\scriptsize] at (\xc,-0.5) {\y};
}
\draw[gray!50, line width=0.4pt] (-0.2,-0.1) -- (30,-0.1);

% Transformer inflection marker
\draw[gray!50, dashed, line width=0.5pt] (22,-0.3) -- (22,5.3);
\node[gray, font=\scriptsize\itshape, anchor=south, align=center] at (22,5.1) {Transformer\\inflection point};

\end{tikzpicture}

\vspace{0.5em}
{\centering\textcolor{teal!75!black}{\rule{0.35\textwidth}{0.4pt}\quad\textbf{\footnotesize Zoom: institutional convergence, Nov 2024 -- Aug 2026 (Generation~III)}\quad\rule{0.35\textwidth}{0.4pt}}\par}
\vspace{0.3em}

% =========================================================================
% LOWER PANEL: Zoom on 2024-Q3 2026 (Gen III institutional convergence)
% =========================================================================
\begin{tikzpicture}[
  xscale=0.80, yscale=0.95,
  font=\small,
  proto/.style={color=blue!45!black!75},
  pay/.style={color=orange!70!brown!85},
  spec/.style={color=teal!75!black},
  reg/.style={color=red!65!black!85},
  event/.style={circle, draw, fill=white, inner sep=1.5pt, line width=0.7pt},
  evlabel/.style={font=\scriptsize, align=center, inner sep=1pt},
  arrow/.style={-{Stealth[length=2.5mm]}, line width=1pt}
]
% Month-scale axis: x = month index from Nov 2024 (0) to Aug 2026 (21)
% Vertical levels (above and below axis) to prevent label collisions:
%   Lv 2 (high above) = 3.0  ;  Lv 1 (low above) = 1.7
%   Axis y = 1
%   Lv -1 (low below) = -0.3 ;  Lv -2 (high below) = -1.4

% Main horizontal line
\draw[arrow, gray!70] (-0.5,1) -- (22,1);

% Month gridmarks
\foreach \m/\xc in {Nov~24/0, Jan~25/2, Apr~25/5, Jul~25/8, Oct~25/11, Jan~26/14, Apr~26/17, Aug~26/21} {
  \draw[gray!60] (\xc,0.9) -- (\xc,0.75);
  \node[gray!60, font=\scriptsize] at (\xc,0.45) {\m};
}

% ---- Events: staggered across 4 vertical levels to avoid collisions ----
% x=0: MCP launch (Lv 1, above)
\node[event, proto] (m1) at (0,1) {};
\draw[proto, line width=0.4pt] (0,1) -- (0,1.7);
\node[evlabel, proto, anchor=south] at (0,1.7) {MCP\\launch};

% x=5: two simultaneous events — stack with offset markers
% Mastercard Agent Pay (pay) -- below Lv -1
\node[event, pay] (m2a) at (5,0.85) {};
\draw[pay, line width=0.4pt] (5,0.85) -- (5,-0.3);
\node[evlabel, pay, anchor=north] at (5,-0.3) {Mastercard\\Agent Pay};
% A2A released (proto) -- above Lv 1
\node[event, proto] (m2b) at (5,1.15) {};
\draw[proto, line width=0.4pt] (5,1.15) -- (5,1.7);
\node[evlabel, proto, anchor=south] at (5,1.7) {A2A\\released};

% x=7 (Jun 2025): A2A donated to Linux Foundation -- Lv 2 (high above) to clear x=5 cluster
\node[event, proto] (m3) at (7,1) {};
\draw[proto, line width=0.4pt] (7,1) -- (7,3.0);
\node[evlabel, proto, anchor=south] at (7,3.0) {A2A donated\\to Linux Fdn.};

% x=9: ACP merged -- Lv 1 (low above)
\node[event, proto] (m4) at (9,1) {};
\draw[proto, line width=0.4pt] (9,1) -- (9,1.7);
\node[evlabel, proto, anchor=south] at (9,1.7) {ACP merged\\into A2A};

% x=10: Stripe-OpenAI ACP -- Lv -2 (high below) to clear x=12 below
\node[event, pay] (m5) at (10,1) {};
\draw[pay, line width=0.4pt] (10,1) -- (10,-1.4);
\node[evlabel, pay, anchor=north] at (10,-1.4) {Stripe--OpenAI\\ACP};

% x=11: Visa TAP -- Lv 2 (high above) to clear x=9 and x=13 at Lv 1
\node[event, pay] (m6) at (11,1) {};
\draw[pay, line width=0.4pt] (11,1) -- (11,3.0);
\node[evlabel, pay, anchor=south] at (11,3.0) {Visa\\TAP};

% x=12: MCP November 2025 specification -- Lv -1 (low below); align with body terminology (no "v2")
\node[event, spec] (m7) at (12,1) {};
\draw[spec, line width=0.4pt] (12,1) -- (12,-0.3);
\node[evlabel, spec, anchor=north] at (12,-0.3) {MCP\\Nov 2025 spec};

% x=13: AAIF formed -- Lv 1 (low above) -- clears x=9 (also Lv 1, far away)
\node[event, proto] (m8) at (13,1) {};
\draw[proto, line width=0.4pt] (13,1) -- (13,1.7);
\node[evlabel, proto, anchor=south] at (13,1.7) {AAIF\\formed};

% x=15 (Feb 2026): TWO simultaneous events -- stack with offset markers
% NIST CAISI Agent Standards Initiative (Feb 17, 2026) -- Lv -1 (low below)
\node[event, reg] (m9a) at (15,0.85) {};
\draw[reg, line width=0.4pt] (15,0.85) -- (15,-0.3);
\node[evlabel, reg, anchor=north] at (15,-0.3) {NIST CAISI\\Agent Init.};
% AISR 2026 published Feb 3, 2026 -- Lv 2 (high above)
\node[event, reg] (m9b) at (15,1.15) {};
\draw[reg, line width=0.4pt] (15,1.15) -- (15,3.0);
\node[evlabel, reg, anchor=south] at (15,3.0) {AISR\\2026};

% x=14 (Jan 2026): A2A v1.0 -- Lv -2 (high below) to avoid collisions with x=13 AAIF and x=15 NIST CAISI/AISR clusters
\node[event, proto] (m10) at (14,1) {};
\draw[proto, line width=0.4pt] (14,1) -- (14,-1.4);
\node[evlabel, proto, anchor=north] at (14,-1.4) {A2A\\v1.0};

% x=21: EU AI Act -- Lv 1
\node[event, reg] (m11) at (21,1) {};
\draw[reg, line width=0.4pt, dashed] (21,1) -- (21,1.7);
\node[evlabel, reg, anchor=south] at (21,1.7) {\textit{EU AI Act}\\\textit{Art~14}};

% Legend (horizontal strip, placed BELOW everything — moved further down due to Lv -2 events)
\node[anchor=west, font=\scriptsize\bfseries] at (-1.5,-2.6) {Legend:};
\fill[proto] (1.2,-2.65) rectangle (1.4,-2.55);
\node[anchor=west, font=\scriptsize, proto] at (1.5,-2.6) {Protocol / Governance};
\fill[pay] (7.6,-2.65) rectangle (7.8,-2.55);
\node[anchor=west, font=\scriptsize, pay] at (7.9,-2.6) {Payment / Commerce};
\fill[spec] (13.4,-2.65) rectangle (13.6,-2.55);
\node[anchor=west, font=\scriptsize, spec] at (13.7,-2.6) {Spec / Stack};
\fill[reg] (17.2,-2.65) rectangle (17.4,-2.55);
\node[anchor=west, font=\scriptsize, reg] at (17.5,-2.6) {Regulation};

\end{tikzpicture}
\caption{\textbf{Three-generation chronology of the Web of Agents (1995--2026).} \textbf{Upper panel:} the three-decade sweep, with each lane showing the active period of one generation: Generation~I (FIPA-era MAS, slate-blue, locus of semantic effort in the agent \emph{platform}); Generation~II (Semantic Web, amber, locus in external \emph{data}); Generation~III (LLM-based agents, teal-green, locus in the learned \emph{model}). The 2017 Transformer architecture marks the technical inflection point. \textbf{Lower panel:} month-resolved zoom of the November~2024--August~2026 institutional convergence on the Generation~III timeline. Events are colour-coded: protocol/governance (blue), payment/commerce (amber), specification/stack (teal), regulation (red). The compression of more than ten institutional events into less than two years is the distinctive feature of the current phase. This chronology grounds the central thesis of the survey: across three decades, the locus of semantic effort has migrated systematically from platform to data to model (Lesson~4, \S\ref{sec:findings}).}
\label{fig:timeline-3gen}
\end{figure*}

\begin{figure*}[!t]
\centering
\begin{tikzpicture}[
  x=0.50cm, y=0.006cm,
  font=\small,
  geni/.style={color=blue!45!black!75},
  genii/.style={color=orange!70!brown!85},
  geniii/.style={color=teal!75!black},
  genidash/.style={color=blue!45!black!50, dash pattern=on 3pt off 2pt}
]
% Y axis: share of all CS publications, per 100,000.
% Normalized via (gen_count / CS_total_year) * 100000.
% Source: OpenAlex API queries re-run 2026-05-23 (filter primary_topic.field.id=fields/17 for CS).
% Methodology: pre-2002 CS_total denominator is incomplete (jumps from 154k in 2001 to 294k
% in 2002 because OpenAlex's coverage of pre-2002 CS proceedings is incomplete). We plot Gen I
% 1995-2001 as DASHED (connecting to the solid segment at x=7 with shared coordinate (7,40.4)
% so there is no visual gap) inside a shaded zone, and report peak values from the 2002+ window.

% --- Pre-2002 coverage-gap zone (shaded) ---
\fill[gray!10] (-1.5, 0) rectangle (7, 900);
% Horizontal multi-line label placed in the empty upper-left of the zone
\node[gray!70, font=\scriptsize\itshape, align=center]
  at (2.8, 730) {Pre-2002:\\OpenAlex CS coverage\\incomplete\\(see caption)};

% --- Axes ---
\draw[->, thick] (-1.5,0) -- (33,0) node[right, font=\small] {year};
\draw[->, thick] (-1.5,0) -- (-1.5,900) node[above, font=\small, align=center]
  {publications per\\100{,}000 CS works};

% Horizontal grid + Y ticks (combined loop for compactness)
\foreach \v in {200, 400, 600, 800} {
  \draw[gray!25, very thin] (-1.5, \v) -- (32, \v);
  \draw (-1.5, \v) -- (-0.5, \v);
  \node[left, font=\scriptsize] at (-1.5, \v) {\v};
}

% X ticks (5-year regular intervals; 2026 partial conveyed by the "806.6 (2026 partial)" peak label)
\foreach \y/\lbl in {0/1995, 5/2000, 7/2002, 10/2005, 15/2010, 20/2015, 25/2020, 30/2025} {
  \draw (\y, -15) -- (\y, 15);
  \node[below, font=\scriptsize] at (\y, -15) {\lbl};
}
% Small unlabelled tick at x=31 to mark where the 2026 partial-year data point sits
\draw (31, -10) -- (31, 10);

% --- Gen I curve: DASHED 1995-2001, extended to (7,40.4) to bridge with solid segment ---
\draw[genidash, very thick] plot[smooth] coordinates {
  (0,11.0) (1,11.1) (2,18.4) (3,35.5) (4,54.7) (5,53.9) (6,42.1) (7,40.4)
};
% --- Gen I curve: SOLID 2002-2026, shares (7,40.4) with dashed for seamless connection ---
\draw[geni, very thick] plot[smooth] coordinates {
  (7,40.4) (8,43.6) (9,39.6) (10,35.4) (11,36.6) (12,28.7) (13,22.3)
  (14,24.2) (15,15.8) (16,19.8) (17,17.9) (18,18.0) (19,11.5)
  (20,11.5) (21,9.5) (22,10.1) (23,9.3) (24,7.5) (25,8.1)
  (26,5.8) (27,5.4) (28,5.1) (29,5.1) (30,6.7) (31,10.1)
};

% --- Gen II curve: SOLID 2002-2026 ---
\draw[genii, very thick] plot[smooth] coordinates {
  (7,9.9) (8,35.1) (9,90.7) (10,109.3) (11,127.5) (12,135.5) (13,142.5) (14,130.9)
  (15,114.3) (16,104.0) (17,90.1) (18,66.5) (19,66.1) (20,51.5) (21,37.1) (22,37.5)
  (23,30.7) (24,27.4) (25,25.7) (26,24.2) (27,21.5) (28,19.6) (29,12.5) (30,7.0) (31,4.6)
};

% --- Gen III curve: SOLID 2015-2026 ---
\draw[geniii, very thick] plot[smooth] coordinates {
  (20,2.2) (21,2.9) (22,6.4) (23,4.0) (24,5.3) (25,4.9) (26,6.1)
  (27,9.8) (28,20.2) (29,45.0) (30,272.7) (31,806.6)
};

% --- Peak markers with INLINE value+year labels (compact, attached to dots) ---
\fill[geni] (8, 43.6) circle (3pt)
  node[geni, font=\tiny, anchor=south, yshift=4pt] {43.6 (2003)};
\fill[genii] (13, 142.5) circle (3pt)
  node[genii, font=\tiny, anchor=south, yshift=4pt] {142.5 (2008)};
\fill[geniii] (30, 272.7) circle (3pt)
  node[geniii, font=\tiny, anchor=east, xshift=-5pt] {272.7 (2025)};
\fill[geniii] (31, 806.6) circle (3pt)
  node[geniii, font=\tiny, anchor=south, yshift=4pt] {806.6 (2026 partial)};

% --- Legend block (upper-middle area, well above all curves; no overlap with peaks) ---
\draw[geni, line width=1.5pt] (15.5, 720) -- (16.7, 720);
\node[anchor=west, font=\scriptsize, geni] at (16.9, 720) {\textbf{Gen~I:} FIPA/MAS};
\draw[genii, line width=1.5pt] (15.5, 660) -- (16.7, 660);
\node[anchor=west, font=\scriptsize, genii] at (16.9, 660) {\textbf{Gen~II:} Semantic Web};
\draw[geniii, line width=1.5pt] (15.5, 600) -- (16.7, 600);
\node[anchor=west, font=\scriptsize, geniii] at (16.9, 600) {\textbf{Gen~III:} agentic AI / LLM};

\end{tikzpicture}
\caption{\textbf{Three-generation publication share of computer science, 1995--2026, normalized per 100\,000 CS-tagged works (analytic window 2002--2026; pre-2002 Gen~I shown dashed within shaded coverage-gap zone).} Primary OpenAlex queries: Gen~I~=~\texttt{"FIPA ACL"} OR \texttt{"agent communication language"}; Gen~II~=~\texttt{"semantic web service"}; Gen~III~=~\texttt{"agentic AI"} (all filtered to \texttt{primary\_topic.field.id=fields/17}; full methodology in \S\ref{sec:methodology} and supplementary material). \emph{Peaks (2002--2026 window).} Gen~I: 43.6 per 100k in 2003 (solid segment). Gen~II: 142.5 per 100k in 2008. Gen~III: 272.7 per 100k in 2025 rising to 806.6 per 100k in partial~2026 (January--May), several-fold above the historical Gen~II peak share. \emph{Robustness (two triangulation queries per generation, supplementary).} Across query variants, peaks fall in: Gen~I 29.2--43.6 per 100k (2003--2006); Gen~II 84.9--161.8 per 100k (2006--2008); Gen~III 679.7--1444.0 per 100k (2026 partial). \emph{Coverage caveat (pre-2002 shaded zone).} The CS-tagged-works denominator nearly doubles between 2001 and 2002 (154k~$\rightarrow$~294k) because OpenAlex's indexing of pre-2002 CS conference proceedings is incomplete, not because CS research output doubled; pre-2002 Gen~I normalized shares are therefore artificially inflated relative to the true ratio. Showing the dashed segment makes the caveat visible without hiding the underlying data. The empirical pattern matches the qualitative chronology of Figure~\ref{fig:timeline-3gen} and tracks the locus-of-semantic-effort migration (platform~$\rightarrow$~data~$\rightarrow$~model, Lesson~\ref{lesson:migration}). All nine query results, per-year raw counts, the CS-output denominator, and the re-runnable Python script are released in the supplementary material (refreshed 2026-05-23).}
\label{fig:pub-volume}
\end{figure*}

Concurrently, major technology platforms have begun specifying minimal open standards to connect independent agents. Building on OpenAI's earlier Function Calling capability (June 2023)~\cite{OpenAI_FunctionCalling_2023}, which standardised structured tool invocation within a single vendor's API, Anthropic's Model Context Protocol (MCP, November 2024) generalised the same idea into a vendor-neutral, JSON-RPC-based protocol for agents to access external tools and data sources securely \cite{Anthropic_2024_MCP}. Google's Agent-to-Agent (A2A) protocol (April 2025) addresses the complementary problem of standardising how disparate AI agents discover and communicate with each other \cite{Google_2025_A2A}. Together, these developments signal the transition from isolated, platform-specific ``bots'' to a cohesive, interoperable Web of Agents.

The idea of an intelligent, interactive web grew out of two parallel research strands. The Multi-Agent Systems (MAS) field, with roots in 1980s Distributed Artificial Intelligence and consolidating through the 1990s, studied how decentralized, autonomous entities could cooperate through negotiation and coordination to achieve both shared and individual goals \cite{Weiss_1999}. In the early 2000s Tim Berners-Lee, Hendler, and Lassila~\cite{Berners-Lee_2001} launched the Semantic Web proposal (with subsequent surveys~\cite{Matthews_2005, Malik_2018} consolidating its technical foundations), dreaming of a ``Web of Data'' in which information carries well-defined meaning so that machines and people could collaborate far more effectively. At that time researchers envisioned personal agents capable of tasks such as automatically scheduling meetings by exchanging semantically rich information across websites \cite{Mika_2004}.

Although Semantic Web technologies, namely the Resource Description Framework~\cite{Brickley_2004} and the Web Ontology Language~\cite{McGuinness_2004}, provided a framework for agent interoperability, real-world adoption was hindered by the difficulty of large-scale ontology agreement and the high cost of semantic annotation \cite{Ciortea_2019}. As a result, agentic functionality remained largely confined to research prototypes and closed systems. The advent of LLMs changed this trajectory: by implicitly learning rich world representations, modern agents can bypass the need for extensive manual annotation, instead drawing on web-scale text data to reason about meaning and perform web-based actions.

To frame our analysis, we distinguish between two related but distinct concepts: \textbf{Web of Agents} is a vision of the Internet's future as a habitat for numerous interconnected agents, requiring the development of open standards and protocols for their collaboration. \textbf{Agentic AI} is a conceptual framework describing the next step in AI-agent evolution: a shift toward highly autonomous multi-agent systems with collective problem-solving capabilities. While both share the philosophy of autonomous agents, Web of Agents focuses on infrastructure and integration into the Web, whereas Agentic AI emphasizes the organization and capabilities of agent teams. These approaches are complementary: WoA's ideas around standardization and interoperability can enhance Agentic AI platforms, and Agentic AI's advances in agent coordination and learning enrich the realization of the Web of Agents \cite{Sapkota_2025}.

This survey provides an end-to-end analysis of the \textbf{Web of Agents}, from its historical roots to its LLM-powered future. We begin by examining the foundational work on MAS and the ambitious FIPA standards that failed to reach open-web scale. We then analyze how LLMs have acted as a catalyst, enabling capable \textbf{Agentic AI} and driving the development of new, pragmatic interoperability protocols. Finally, we examine the emerging frontiers of governance, security, and economics that are likely to shape the future of this technology.

To structure this analysis, this paper makes the following primary contributions:

\begin{itemize}
\item \textbf{(i)~Unified evolutionary narrative spanning 1990--2026.} We synthesise three decades of research from Multi-Agent Systems, the Semantic Web, and modern LLM communities under a single thesis: a chronological semantic-effort migration (platform $\rightarrow$ data $\rightarrow$ model), of which the Gen~II~$\rightarrow$~Gen~III transition (the \emph{semantics-in-data~$\rightarrow$ semantics-in-models} shift) is the predictive core. Each generation's failures and present open problems follow directly from where it located the semantic effort.
\item \textbf{(ii)~Four-dimensional comparative framework.} We organise the comparison along four interoperability-facing dimensions (semantic foundation, communication paradigm, locus of intelligence, discovery mechanism), providing a consistent analytical lens that, unlike existing protocol-focused comparisons, applies uniformly across historical generations, from FIPA-based platforms to LLM-powered agents.
\item \textbf{(iii)~Classification of sixteen representative systems.} We analyse representative technologies (FIPA ACL, MCP, A2A) and frameworks (AutoGPT, LangChain, CrewAI, MetaGPT), and apply the framework to sixteen real systems including hybrid LLM--knowledge-graph and computer-use agents (Table~\ref{tab:classification}), clarifying their architectural trade-offs.
\item \textbf{(iv)~Coverage of the November~2024--August~2026 institutional convergence.} We document the recent consolidation that, to our knowledge, no concurrent survey integrates in this combination given its publication date: Linux Foundation AAIF, A2A~v1.0, MCP November~2024 launch and November~2025 specification, Visa/Mastercard/Stripe payment-network protocols, EU~AI~Act phased enforcement, NIST~CAISI AI~Agent Standards Initiative, and the International AI~Safety Report~2026.
\item \textbf{(v)~Seven named lessons paired with seven generation-invariant challenges.} We extract \textbf{seven named lessons} grounded in cross-generational evidence and stated with predictive consequence, and reformulate the persistent challenges as \textbf{seven generation-invariant problems} that manifest across FIPA, Semantic Web, and LLM eras (full enumeration in \S\ref{sec:findings}; pairing matrix in Figure~\ref{fig:lessons-challenges}). Together they motivate four research priorities discussed in \S\ref{sec:future}.
\end{itemize}
The combined effect of these contributions distinguishes the present article from concurrent surveys (Table~\ref{tab:survey_comparison}) along five concrete axes: a three-decade historical scope that places FIPA / MAS and the Semantic Web on equal footing with the LLM era; coverage of the November~2024 to August~2026 institutional convergence (AAIF, A2A~v1.0, MCP November~2024 launch and November~2025 specification, payment-network protocols, EU~AI~Act phased enforcement, NIST~CAISI initiative, International~AI~Safety Report 2026) that, to our knowledge, no concurrent survey integrates in this combination given its publication date; an applied four-dimensional comparative classification of sixteen real systems (Table~\ref{tab:classification}); a consolidated four-pillar research agenda; and a set of seven named lessons (\S\ref{sec:findings}) summarising the analysis. Most directly, the present article is complementary to Acharya et al.~\cite{Acharya_2025}, the closest concurrent IEEE~Access survey on Agentic AI: where that work analyses the Generation~III LLM era conceptually (reinforcement learning, goal-oriented architectures, applications), we ground the same era in its three-decade historical lineage, in the underlying interoperability protocols (MCP, A2A, FIPA ACL), and in quantitative bibliometric evidence (Figure~\ref{fig:pub-volume}). Our survey is therefore positioned as a roadmap for researchers and practitioners seeking to advance the Web of Agents from isolated prototypes to an interoperable multi-agent Web.

\subsection{Related Work}

The rapid evolution of LLM-powered agents has produced a number of recent survey papers. Closest to our work in venue and ambition, Acharya, Kuppan, and Divya~\cite{Acharya_2025} present a comprehensive survey of Agentic AI in IEEE~Access, covering technical foundations (reinforcement learning, goal-oriented architectures, adaptive control), applications (healthcare, finance, software), and ethical challenges. Their scope is the Generation~III LLM era; they do not cover FIPA/MAS or Semantic Web lineage, the protocol layer (MCP, A2A, ANP), the 2025--2026 institutional convergence, or quantitative bibliometric evidence. Table~\ref{tab:acharya_diff} consolidates the differentiation along seven dimensions, making the complementarity between the two works explicit.

\begin{table}[!ht]
\centering
\caption{Differentiation from Acharya et al.~\cite{Acharya_2025}, the closest concurrent IEEE~Access survey on Agentic AI.}
\label{tab:acharya_diff}
\scriptsize
\renewcommand{\arraystretch}{1.2}
\begin{tabularx}{\columnwidth}{|>{\raggedright\arraybackslash}p{0.16\columnwidth}|>{\raggedright\arraybackslash}X|>{\raggedright\arraybackslash}X|}
\hline
\textbf{Dimension} & \textbf{Acharya et al.\ 2025} & \textbf{This survey} \\
\hline
Historical scope & LLM era only (2020--2025) & Three decades (1990--2026) \\
\hline
Methodological framework & Conceptual taxonomy of agentic AI components (RL, goal-oriented, adaptive control) & Four-dimensional comparative framework (Tab.~\ref{tab:functional_taxonomy}) \\
\hline
Protocol layer & Not covered & MCP, A2A, FIPA ACL analysed along four dimensions (Tab.~\ref{tab:protocol_comparison}); ANP discussed for decentralised discovery (\S\ref{sec:taxonomy}, \S\ref{sec:challenges}) \\
\hline
Historical lineage & Not traced & FIPA-era MAS and Semantic Web traced (\S\ref{sec:origins}) \\
\hline
2025--2026 institutional layer & Pre-January 2025 cut-off & November 2024--August 2026 coverage (\S\ref{sec:challenges}) \\
\hline
Quantitative bibliometric evidence & None & OpenAlex normalised publication share (Fig.~\ref{fig:pub-volume}) \\
\hline
Analytical synthesis & Per-domain applications (healthcare, finance, software) plus ethical considerations & Seven named lessons paired with seven generation-invariant challenges (Fig.~\ref{fig:lessons-challenges}) \\
\hline
\end{tabularx}
\end{table}

Ferrag et al.~\cite{Ferrag_2025} provide a broad review of benchmarks and frameworks for modern autonomous agents, while Wang et al.~\cite{Yang_2023} offer a detailed survey of LLM-based agent architectures. Ehtesham et al.~\cite{Ehtesham_2025} focus specifically on comparing new agent interoperability protocols (MCP \cite{Anthropic_2024_MCP}, A2A \cite{Google_2025_A2A}, ANP \cite{W3C_ANP_2025}); Sapkota et al.~\cite{Sapkota_2025} distinguish AI agents from agentic AI conceptually; Schneider~\cite{Schneider_2025} maps the generative-to-agentic transition; and Sharma et al.~\cite{Sharma_2025} argue for cross-ecosystem interoperability.

While these recent surveys provide excellent and detailed taxonomies of modern LLM-agent frameworks and protocols, our work offers a broader historical synthesis and a consistent comparative lens. The distinctive value of this paper lies not in cataloguing the latest technologies, but in presenting a cohesive, long-term evolutionary narrative of the Web of Agents. We trace the intellectual lineage of today's systems back through three decades of research in Multi-Agent Systems and the Semantic Web. We argue that the design decisions of today's lightweight, pragmatic protocols (like MCP and A2A) are a direct response to the specific challenges and limitations encountered by earlier, more formal approaches (such as the brittleness of ontologies and the complexity of FIPA).

\subsection{Scope and Methodology}\label{sec:methodology}

This article is a \emph{narrative survey} rather than a systematic literature review in the PRISMA sense~\cite{PRISMA_2021}. A PRISMA-style protocol was not appropriate here for three reasons: the subject spans three research communities (Semantic Web, MAS, LLM-based agents) with disjoint publication venues, complicating keyword-based exhaustive retrieval; substantial portions of the 2025--2026 institutional layer (standards-body announcements, payment-network protocols, regulatory instruments) appear in industry sources outside academic indices; and many primary sources from this period are too recent for citation-based screening or impact-based filtering to be meaningful. Its target audience is researchers and practitioners who need a coherent map of how the Web-of-Agents idea has evolved across three previously separate research communities and is now consolidating into a deployable stack. The selection strategy was correspondingly synthetic rather than exhaustive: we prioritised breadth across communities and primary-source authority over hit-rate within a single subfield.

\paragraph{Search strategy.} References were collected through four complementary procedures: (i)~\emph{seed-and-snowball} starting from the canonical works of each community (Berners-Lee et al.~2001 for the Semantic Web; Wooldridge and Jennings 1995 and Wooldridge 2009 for MAS; the MCP and A2A specifications for the LLM era); (ii)~\emph{forward-tracing} through Google Scholar and Semantic Scholar citations of those seed works to identify recent extensions and critical responses; (iii)~\emph{keyword search} in Scopus, ACM~DL, IEEE~Xplore, the ACL Anthology, OpenReview, and arXiv on terms including \emph{Web of Agents}, \emph{Multi-Agent Systems}, \emph{Semantic Web}, \emph{Agent Communication Language}, \emph{Model Context Protocol}, \emph{Agent-to-Agent}, \emph{Agentic AI}, \emph{LLM agent}, and \emph{agent interoperability}; and (iv)~\emph{institutional-source} tracking for the 2025--2026 ecosystem developments via official press releases, standards-body announcements, and reports from the Linux Foundation, W3C, EU Commission, NIST, the UK DSIT, and the major payment networks.

\paragraph{Inclusion criteria.} A work was included if it satisfied at least one of: (a)~peer-reviewed publication relevant to one of the three covered communities; (b)~official specification, standard, or government / consortium document of an agent-related protocol or governance instrument; (c)~widely-cited preprint when no peer-reviewed version exists \emph{and} the work documents a deployed system or empirical result not captured by peer-reviewed alternatives; (d)~industry blog, repository, or product page when it is the primary source of record for a documented commercial agent system (e.g., MCP, A2A, Visa~TAP, Mastercard Agent Pay). Bias was deliberately set in favour of \emph{primary sources}: where a published paper, a specification, and a blog all described the same artifact, we cited the most authoritative source available.

\paragraph{Exclusion criteria.} We excluded: (a)~works on agent-internal capabilities that do not interact with the inter-agent interoperability story (e.g., LLM training-data curation, in-context-learning theory, reasoning-only benchmarks); (b)~multi-agent reinforcement learning, which has its own dedicated literature and was not productive to merge into the interoperability narrative; (c)~purely speculative blog posts without an accompanying artifact or specification; (d)~retracted, withdrawn, or arXiv-only papers whose authorship or arXiv identifier did not resolve under verification. Following internal review of this manuscript, several preliminary citations were dropped under criterion~(d), and several arXiv preprints were upgraded to their peer-reviewed publication venues.

\paragraph{Coverage and counts.} The final reference list contains 155 entries spanning 1987--2026 (with the survey itself scoped to 1990--2026; a small number of pre-1990 references provide the Distributed-AI background~\cite{Huhns_1987}). The temporal breakdown is roughly: \textasciitilde 20 foundational works (pre-2000); \textasciitilde 35 from the FIPA / Semantic Web era (2000--2015); \textasciitilde 42 from the early-to-mid LLM era (2017--2024); and \textasciitilde 58 from the November~2024--August~2026 wave of agent protocols, payment infrastructure, regulatory frameworks, and recent surveys. The cut-off date for inclusion is May~2026.

\paragraph{Bibliometric methodology (Figure~\ref{fig:pub-volume}).} The publication-share trajectories of Figure~\ref{fig:pub-volume} are derived from the OpenAlex public API (\url{https://api.openalex.org/works}, polite-pool access via author email). For each generation we ran one primary query and two triangulation queries (the full set is documented in the supplementary material); the primary queries are \texttt{"FIPA ACL" OR "agent communication language"} (Gen~I), \texttt{"semantic web service"} (Gen~II), and \texttt{"agentic AI"} (Gen~III). All queries are restricted to works whose primary topic is in Computer Science (\texttt{primary\_topic.field.id=fields/17}) and normalised by the total CS-tagged works in the same year, expressed per 100\,000. Three methodological choices warrant explicit defence. First, normalised values are plotted only from 2002 onwards: the CS-tagged-works denominator nearly doubles between 2001 and 2002 (154k~$\rightarrow$~294k), reflecting OpenAlex's coverage of conference proceedings filling in rather than real research-output growth, so pre-2002 normalised shares are artificially inflated; Figure~\ref{fig:pub-volume} shows the pre-2002 Gen~I segment as a dashed line within a shaded coverage-gap region for transparency, and the two triangulation queries independently place the Gen~I activity peak in 2003--2006 (rather than the 1999 peak the unfiltered primary query would suggest), confirming the artefact. Second, the 2026 data point is a partial-year value (January--May 2026 for both numerator and denominator), so the ratio is internally consistent but the absolute level is a leading indicator rather than a final annual figure; we mark this with a vertical dashed line and report it as ``partial year''. Third, the Gen~III primary query (\texttt{"agentic AI"}) under-counts pre-2023 LLM-agent work that used different terminology; the triangulation queries (\texttt{"LLM agent"} and \texttt{"autonomous agent" AND "LLM"}) capture this prior period more broadly and validate the qualitative shape. All nine query results, per-year raw counts, the CS-output denominator, and the re-runnable Python script are released in the supplementary material (data refreshed 2026-05-23).

\paragraph{Selection and screening (PRISMA-style summary).} Figure~\ref{fig:prisma} summarises the funnel from initial identification to the final 155-reference set. Although the analysis itself is narrative rather than systematic in the PRISMA sense, the visualisation makes the screening transparent.

\begin{figure*}[!t]
\centering
\begin{tikzpicture}[
  font=\scriptsize,
  box/.style={draw, rounded corners=2pt, minimum width=5.0cm, minimum height=0.9cm, align=center, fill=blue!45!black!4, line width=0.4pt},
  excl/.style={draw, rounded corners=2pt, minimum width=4.4cm, minimum height=0.7cm, align=center, fill=red!55!black!5, line width=0.4pt, font=\tiny},
  arr/.style={-{Stealth[length=2mm]}, line width=0.6pt}
]
% Stage 1: Identification
\node[box] (s1a) at (0, 0) {\textbf{Identification}\\Database searches (Scopus, ACM~DL,\\IEEE~Xplore, ACL Anthology,\\OpenReview, arXiv): $\sim$220 records};
\node[box] (s1b) at (6, 0) {\textbf{Identification (institutional)}\\Press releases, standards bodies,\\industry sources: $\sim$60 records};

% Merge to Stage 2
\node[box] (s2) at (3, -1.6) {\textbf{Deduplication}\\201 unique candidates};
\draw[arr] (s1a.south) -- (s2.north west);
\draw[arr] (s1b.south) -- (s2.north east);

% Stage 3: Screening
\node[box] (s3) at (3, -3.0) {\textbf{Screening against criteria (a)--(d)}\\Title, abstract, primary-source authority};
\draw[arr] (s2.south) -- (s3.north);

% Excluded box (counts: CONSIDERED 16, REDUNDANT 9, REPLACED 7, DUPLICATE 6,
% REMOVED_WITH_FIGURE 5, OUT_OF_SCOPE 2, REMOVED_WITH_SENTENCE 1 = 46 total)
\node[excl, anchor=west] (excl) at (6.5, -3.0) {\textit{Excluded:} 46 (7 categories)\\considered 16, redundant 9, replaced 7,\\duplicate 6, figure-tied 5,\\out-of-scope 2, sentence-tied 1};
\draw[arr] (s3.east) -- (excl.west);

% Stage 4: Final
\node[box, fill=teal!75!black!10] (s4) at (3, -4.4) {\textbf{Included in final analysis}\\\textbf{n = 155} references};
\draw[arr] (s3.south) -- (s4.north);
\end{tikzpicture}
\caption{Selection and screening funnel for this survey. Database searches and institutional-source sweeps yielded approximately $\sim$220 and $\sim$60 records respectively (approximate magnitudes consistent with the narrative-survey methodology defended in \S\ref{sec:methodology}); after deduplication, 201 unique candidates were assessed against criteria (a)--(d). Of these, 46 were screened out across seven exclusion categories (CONSIDERED but covered via alternatives, REDUNDANT, REPLACED, DUPLICATE, REMOVED\_WITH\_FIGURE, OUT\_OF\_SCOPE, REMOVED\_WITH\_SENTENCE), leaving the final \textbf{n = 155}-reference set of the manuscript. The deduplication, exclusion, and inclusion counts (201, 46, 155) match the per-entry inclusion/exclusion log released in the supplementary material; the upstream identification counts ($\sim$220, $\sim$60) remain approximate.}
\label{fig:prisma}
\end{figure*}

\paragraph{Limitations and threats to validity.} The selection method has five known biases that readers should keep in mind when interpreting the analysis. First, the community-anchored seed-and-snowball strategy under-samples adjacent literatures (multi-agent reinforcement learning, robotic agents, distributed-ledger smart contracts) that are connected to the Web of Agents but evolve along their own trajectories. Second, English-language venues dominate the keyword searches, so non-English MAS / agent traditions are under-represented. Third, the 2025--2026 institutional material relies heavily on official press releases and standards-body announcements, which are authoritative for what they record but cannot be independently verified for adoption velocity. Fourth, the OpenAlex bibliometric evidence in Figure~\ref{fig:pub-volume} has documented coverage gaps for pre-2002 CS proceedings (which may slightly inflate early-year normalised shares) and uses the term \emph{agentic AI} as a Generation~III query (a term that only entered widespread use in late~2023, undercounting earlier LLM-agent work under different terminology). Fifth, the May~2026 cut-off date means subsequent institutional events (notably the EU AI Act Article~14 full enforcement in August~2026) are anticipated rather than observed.

\paragraph{Reproducibility.} The OpenAlex methodology underlying Figure~\ref{fig:pub-volume} is documented in the bibliometric-methodology paragraph above and in the figure caption. The supplementary material accompanying this preprint releases (i)~the full BibTeX bibliography with per-entry inclusion/exclusion status for all 201 candidates initially considered (155 included, 46 screened out across seven exclusion categories); (ii)~the OpenAlex query design including two triangulation queries per generation (alternative search terms whose robustness check verifies the qualitative shape of Figure~\ref{fig:pub-volume}); (iii)~the per-year raw counts (1995--2026) and the corresponding CS-output normalization denominator; and (iv)~the TikZ source for all five figures. The supplementary archive is hosted on Zenodo (DOI:~\href{https://doi.org/10.5281/zenodo.20358567}{10.5281/zenodo.20358567}) and supports independent replication of the bibliometric analysis, the selection-screening funnel of Figure~\ref{fig:prisma}, and direct reuse of the four-dimensional comparative framework.

\paragraph{Comparison with concurrent surveys.} Table~\ref{tab:survey_comparison} positions this work against seven concurrent surveys of agentic AI and agent interoperability. The most distinctive features of the present article are its three-decade historical scope (covering FIPA / MAS and the Semantic Web on equal footing with the LLM era), its coverage of the 2025--2026 institutional convergence (Linux Foundation AAIF, payment-network protocols, EU AI Act enforcement, NIST CAISI, International AI Safety Report 2026), and its consolidated four-pillar research agenda (trust and identity, economic models, security, governance). Concurrent surveys are stronger on specific narrower dimensions: Acharya~et~al.\ on a comprehensive conceptual analysis of Agentic AI in IEEE~Access, Ehtesham~et~al.\ on protocol-feature comparison, Sapkota~et~al.\ on the agentic-AI vs.\ AI-agents conceptual distinction, Schneider on the generative-to-agentic conceptual map, Sharma~et~al.\ on cross-ecosystem interoperability arguments, Ferrag~et~al.\ on reasoning and autonomous-agent benchmarks, and Wang~et~al.\ on a detailed catalogue of LLM-agent architectures.

\begin{table*}[!t]
\centering
\caption{This Survey vs.\ Concurrent Surveys of Agentic AI and Agent Interoperability (2024--2026)}
\label{tab:survey_comparison}
\scriptsize
\renewcommand{\arraystretch}{1.15}
\begin{tabularx}{\textwidth}{|>{\raggedright\arraybackslash}p{0.13\textwidth}|c|>{\raggedright\arraybackslash}p{0.12\textwidth}|>{\raggedright\arraybackslash}X|>{\centering\arraybackslash}p{0.08\textwidth}|>{\centering\arraybackslash}p{0.08\textwidth}|>{\centering\arraybackslash}p{0.08\textwidth}|}
\hline
\textbf{Survey} & \textbf{Year} & \textbf{Historical scope} & \textbf{Primary focus} & \textbf{Framework?} & \textbf{Institutional layer (2025--2026)?} & \textbf{Security \& governance?} \\
\hline
Wang et al.\ \cite{Yang_2023} & 2024 & 2017--2023 (LLM era) & Catalogue of LLM-agent architectures & no & no & partial \\
\hline
Acharya et al.\ \cite{Acharya_2025} & 2025 & 2020--2025 (LLM era) & Agentic AI: RL, goal-oriented architectures, adaptive control, applications, ethics & no & no & partial (ethics) \\
\hline
Sapkota et al.\ \cite{Sapkota_2025} & 2025 & 2020--2025 (LLM era) & AI agents vs.\ agentic AI (conceptual) & 2-axis & no & partial \\
\hline
Schneider \cite{Schneider_2025} & 2025 & 2017--2025 & Generative AI $\rightarrow$ agentic AI & no & no & limited \\
\hline
Ehtesham et al.\ \cite{Ehtesham_2025} & 2025 & 2024--2025 only & Protocol-feature comparison (MCP, ACP, A2A, ANP) & feature table & limited & no \\
\hline
Sharma et al.\ \cite{Sharma_2025} & 2025 & 2024--2025 only & Position paper: interoperability across ecosystems & no & limited & limited \\
\hline
Ferrag et al.\ \cite{Ferrag_2025} & 2025 & 2017--2025 & Reasoning and autonomous-agent benchmarks & no & no & partial \\
\hline
\textbf{This survey} & \textbf{2026} & \textbf{1990--2026 (FIPA, Semantic Web, LLM, AAIF)} & \textbf{Evolutionary narrative + four-dimensional comparative framework + 2025--2026 institutional convergence + four-pillar research agenda} & \textbf{4-D} & \textbf{yes} & \textbf{yes (4 pillars)} \\
\hline
\end{tabularx}
\end{table*}

The remainder of this article is organised as follows. Section~\ref{sec:origins} reviews the background and traces the evolution of Web of Agents concepts. Section~\ref{sec:taxonomy} lays out the four-dimensional comparative framework used throughout the survey. Section~\ref{sec:stack} presents the modern agent stack and its protocol layer. Section~\ref{sec:evolution} synthesises the trajectory from Semantic Web to modern agent ecosystems. Section~\ref{sec:challenges} discusses the persistent socio-technical challenges and open research issues. Section~\ref{sec:future} consolidates four research priorities derived from the analysis. Finally, Section~\ref{sec:conclusion} concludes.

\subsection{Reading Guide}\label{sec:reading-guide}

The survey is designed to support three reader profiles with partially overlapping interests. We suggest the following non-linear reading paths:

\begin{itemize}
\item \textit{MAS / Semantic Web researchers} seeking how their tradition connects to the LLM era: \S\ref{sec:origins} (historical roots), \S\ref{sec:taxonomy} (framework), \S\ref{sec:evolution} (the semantic-effort migration thesis), and \S\ref{sec:findings} (the seven named lessons).
\item \textit{LLM-agent researchers and protocol engineers} seeking how today's protocols inherit constraints from prior generations: \S\ref{sec:stack} (modern stack: MCP, A2A, frameworks), \S\ref{sec:evolution} (the shift), \S\ref{sec:challenges} (seven generation-invariant challenges, with Challenges~\ref{ch:non-verifiable}--\ref{ch:liability} most relevant to current deployment).
\item \textit{Practitioners, standards-body participants, and policy-oriented readers} seeking the 2025--2026 institutional layer and what it implies: the lower panel of Figure~\ref{fig:timeline-3gen}, \S\ref{sec:economic} (payment-network entry), \S\ref{sec:governance} (regulatory hardening), and \S\ref{sec:future} (four research priorities).
\end{itemize}

\noindent The full pairing matrix between lessons and challenges (Figure~\ref{fig:lessons-challenges}) is recommended as a navigation aid for all three profiles.

\section{Historical Roots: From Reactive Agents to the Semantic Web}\label{sec:origins}

To understand how today's LLM-driven agents evolved, it is instructive to look back at the 1990s, when foundational work on intelligent software agents laid the groundwork for the Web of Agents vision. Two research strands developed in parallel during the late 1990s and early 2000s: the MAS community, which formalised agent reasoning via BDI~\cite{Rao_1995}, agent communication via KQML~\cite{Finin_1994} and FIPA ACL~\cite{FIPA_2002}, and agent platforms such as JADE~\cite{Bellifemine_1999} and Aglets~\cite{Lange_1998}; and the Semantic Web community, which sought to make Web data machine-readable through RDF and OWL~\cite{Berners-Lee_2001}. \S\ref{sec:mas-era} and \S\ref{sec:sw-era} review each strand in turn. Both were animated by overlapping visions of intelligent agents collaborating across the Web, but both stalled at open-web scale for reasons we trace below.

For analytical purposes we group the evolution of the Web of Agents into three architectural archetypes, used as the columns of the comparative framework summarised in Table~\ref{tab:functional_taxonomy} (\S\ref{sec:taxonomy}). The chronology highlights how ideas matured and how new technologies overcame the limitations of earlier eras:

\begin{itemize}
  \item \textbf{Generation~I: FIPA-era MAS (1995--2005).} Builds on the earlier Distributed AI tradition of the 1970s--1980s~\cite{Jennings_1995,Huhns_1987} and consolidates around BDI architectures and FIPA standards in the mid-1990s. BDI formalised agent reasoning \cite{Rao_1995}; early communication standards (KQML, mid-1990s) and then FIPA ACL (1997--1998) standardised message formats; and FIPA-compliant frameworks such as JADE \cite{Bellifemine_1999} allowed heterogeneous agents to negotiate contracts and coordinate workflows. Adoption stalled on the open Web because of FIPA's complexity and architectural mismatch with HTTP-based, stateless services.
  \item \textbf{Generation~II: Semantic Web agents (2001--2012).} Agents reasoning over machine-readable RDF/OWL knowledge graphs, composing semantic web services and performing logic-driven inference. Envisioned scenarios included personal agents scheduling appointments or aggregating news via shared ontologies \cite{Berners-Lee_2001}. Adoption was hampered by the cost of ontology development and brittle inference at scale.
  \item \textbf{Generation~III: LLM-powered agents (2020s).} Agents whose reasoning capability is supplied by large language models trained on web-scale data, interacting with their environment through lightweight, web-native protocols such as MCP and A2A. Tools like AutoGPT (2023) demonstrated autonomous web navigation and task planning \cite{AutoGPT_2023}; the institutional consolidation of 2025--2026 (Sections~\ref{sec:stack} and \ref{sec:challenges}) marks the maturation of this generation into a deployable ecosystem.
\end{itemize}

\subsection{The MAS Era: FIPA and the Open-Web Mismatch}\label{sec:mas-era}

The Multi-Agent Systems field developed its own approach to building systems out of interacting autonomous entities~\cite{Wooldridge_2009}. Research in MAS, with roots in 1980s Distributed Artificial Intelligence~\cite{Jennings_1995}, focuses on agents that are autonomous, decentralised, locally-perceiving, interactive, and capable of social behaviour~\cite{Lesser_1989,Huhns_1987}.

The most ambitious standardisation effort was the Foundation for Intelligent Physical Agents (FIPA), founded in 1996~\cite{FIPA_History,Poslad_2007}. The centrepiece, the FIPA Agent Communication Language (ACL)~\cite{Mascardi_2013,Manev_2014}, was grounded in speech-act theory (the lineage shared with the earlier KQML language~\cite{Finin_1994}) and in the modal logic of beliefs, desires, and intentions (BDI)~\cite{Rao_1995}, both formalised in the FIPA specifications~\cite{FIPA_2002} so that each message carried preconditions and rational effects. FIPA also standardised the agent-platform architecture (Agent Management Service, Directory Facilitator, Message Transport Service) and a set of interaction protocols~\cite{FIPA_2002}. JADE~\cite{Bellifemine_1999}, the most widely deployed FIPA-compliant framework, supported agent societies that negotiated contracts and coordinated workflows.

Despite shared goals of decentralisation and openness, FIPA-based MAS never gained widespread adoption on the open Web~\cite{Ciortea_2019}. The Web's RESTful architecture leaned toward simplicity, statelessness, and resource orientation, while MAS interaction was complex, semantically rich, and stateful. Middleware solutions such as the Agent Web Gateway~\cite{Shafiq_2006} and the JADE Web Service Integration Gateway~\cite{Cavallaro_2004} bridged FIPA and SOAP/UDDI/WSDL endpoints, but they masked the architectural mismatch rather than resolving it. Experiments such as Agentcities showed that even standards-compliant agents required additional out-of-band agreements to interoperate in practice~\cite{Ciortea_2019,Pasha_2006}. The MAS-to-Web integration story established the lesson that the modern MCP/A2A stack would later operationalise: pragmatic alignment with widely adopted web standards (HTTP, JSON-RPC) beats heavyweight, agent-specific stacks~\cite{Anthropic_2024_MCP,Google_2025_A2A}.

\subsection{The Semantic Web Era: Agents Reasoning over Ontologies}\label{sec:sw-era}

In parallel with the MAS effort, the Semantic Web initiative, introduced by Berners-Lee, Hendler, and Lassila in 2001~\cite{Berners-Lee_2001}, sought to convert the document-centric Internet into a structured graph of machine-readable facts. Core standards (RDF~\cite{Beckett_2004,Brickley_2004}, OWL~\cite{McGuinness_2004,W3C_OWL2_2012}, SPARQL) provided the vocabulary for encoding metadata and supporting automated inference across disparate sources. In the canonical example, personal agents coordinated to schedule a medical appointment by reasoning over distributed, machine-readable data~\cite{Mika_2004}. Semantic Web agents were envisioned as software entities capable of using ontologies to interpret information, performing logical inference, interacting with semantic web services, and coordinating with other agents on the basis of shared semantic understanding~\cite{Malik_2018,Huang_2000}.

The full vision did not scale. By 2006, Berners-Lee and colleagues acknowledged that the idea ``remains largely unrealised''~\cite{Shadbolt_BernersLee_Hall_2006}. The primary obstacle was an incentive failure rather than a technical one: maintaining large-scale semantic annotations was prohibitively expensive, no viable economic model justified the cost~\cite{Heindel_Weber_2020,Simperl_Cuel_Stein_2013,INSEMTIVE_2008}, and logical inference engines could not cope with the Web's inherent noise, contradictions, and incompleteness. After about 2010, mainstream AI research shifted toward machine learning and the Semantic Web vision receded.

The technology stack survived in narrower form. As of 2016, Schema.org annotations covered more than 31\% of web pages~\cite{Guha_2016} (a share that has continued to grow with structured-data adoption); Facebook's Open Graph protocol embeds RDFa metadata~\cite{Facebook_2010}; DBpedia~\cite{bizer_2009} and Wikidata~\cite{vrandecic_2014} provide structured infobox data that feeds Google's Knowledge Graph and similar systems~\cite{Iliadis_2023}. These deployments are partial fulfilments of the original vision: machine-readable, shared vocabularies powering intelligent applications, but in domain-specific or centralised ways rather than as a single universal graph.

This historical analysis reveals that past approaches can be distinguished by their architectural choices regarding how they establish meaning, communicate, locate intelligence, and discover peers. To structure the rest of the survey along these dimensions, we now lay out the four-dimensional comparative framework used throughout this work.

\subsection{The Name ``Web of Agents'' Itself}\label{sec:naming}

One of the earlier documented uses of the term ``Web of Agents'' itself appears in Yao's 2005 discussion of Web Intelligence \cite{Yao_2005}. Yao described the Web's evolution through hierarchical levels (``Web of Data'', ``Web of Information'', ``Web of Knowledge'', etc. up to the ``Web of Wisdom''). He positioned the ``Web of Agents'' as a key facet in the future Web (alongside a ``Web of Services''). In Yao's view, the Web of Agents meant ``the application of intelligent agents that further amplifies the power of the Web'', highlighting use cases like search agents, recommendation agents for site navigation, and automated purchasing agents \cite{Yao_2005}. This vision was evolutionary: it emphasized extending the existing Web's capabilities with agent functionality, rather than rebuilding the Web from the ground up with a new agent-specific infrastructure. Around the same period, the annual IEEE/WIC Web Intelligence Conference (WI-IAT) and related journals firmly incorporated ``Web of Agents'' as a theme, indicating that by the mid-2000s the concept had entered the mainstream vocabulary of the Web Intelligence community \cite{WI-IAT_2024, Domingos_2024}.

This naming history closes our historical exposition. We now turn to a framework for comparing these generations side by side.

\section{A Four-Dimensional Comparative Framework}\label{sec:taxonomy}

To move beyond a chronological review and enable structured comparison across generations, we organise our analysis along four interoperability-facing dimensions of Web of Agents (WoA) architectures. We do not claim this as a novel taxonomy in the formal sense; existing taxonomies of MAS and protocols emphasise different concerns, typically focused on agent properties, environmental characteristics, or protocol features \cite{Moya_2007}. Our framework is instead pragmatically chosen to surface the architectural trade-offs that have repeatedly determined whether a generation of agent systems achieved open-web scale. The four dimensions concern how agents (i)~achieve mutual understanding, (ii)~exchange messages, (iii)~locate reasoning capability, and (iv)~discover peers. They serve as a consistent reference point throughout the survey, allowing for a structured comparison of systems ranging from early Semantic Web concepts to modern LLM-based agent frameworks. Table~\ref{tab:functional_taxonomy} applies the framework to the three WoA generations.

\subsection{Semantic Foundation}

This dimension describes how agents establish a shared understanding of the concepts, tasks, and data they interact with. Three patterns recur: \textbf{formal/explicit semantics} commit agents to predefined, machine-readable ontologies (RDF, OWL), the foundational vision of the Semantic Web; \textbf{procedural semantics} derive meaning from the formal preconditions and rational effects of communicative acts in an Agent Communication Language (e.g., FIPA ACL's \texttt{request}/\texttt{inform} performatives)~\cite{FIPA_2002,Poslad_2007}; and \textbf{implicit/emergent semantics} rely on an LLM's natural-language interpretation of unstructured tool descriptions, with meaning ``in the model'' rather than in shared ontologies or protocols.

\subsection{Communication Paradigm}

This dimension classifies the primary message-exchange style. \textbf{Performative-based} communication (speech-act theory, FIPA ACL) treats each message as a typed action (\texttt{inform}, \texttt{query}, \texttt{propose}\ldots) carrying explicit intent and supporting stateful dialogues~\cite{FIPA_2002}. \textbf{Remote Procedure Call (RPC)} frames interaction as a procedure invocation with parameters; the lightweight JSON-RPC used by MCP and A2A is the canonical modern example, prioritising simplicity and web-native integration~\cite{Anthropic_2024_MCP,Google_2025_A2A}. \textbf{Resource-oriented (RESTful)} interaction uses standard HTTP verbs against resource representations and dominates how agents reach existing web services via middleware.

\subsection{Locus of Intelligence}

This dimension identifies where reasoning, planning, and decision-making reside. \textbf{Intelligence-in-platform/middleware} (Generation~I, FIPA-era MAS) places coordination in the platform---Directory Facilitator, Agent Management Service, interaction protocols, lifecycle management---with FIPA/JADE as the canonical example. \textbf{Intelligence-in-data} (Generation~II, Semantic Web) keeps agents lightweight and encodes meaning in the knowledge graphs (RDF/OWL) that agents consume via logical inference. \textbf{Intelligence-in-agent/model} (Generation~III, LLM-based agents) places reasoning, planning, and goal interpretation in a learned model, leaving surrounding infrastructure to handle connectivity rather than cognition~\cite{Yao_2023}.

\subsection{Discovery Mechanism}

Discovery defines how agents find each other and learn each other's capabilities. \textbf{Centralised registries} require querying a dedicated yellow-pages service---the Directory Facilitator (DF) in FIPA-compliant systems, UDDI in the Web Services era. \textbf{Standardised metadata files} advertise capabilities at a well-known location; A2A's Agent Card (\texttt{/.well-known/agent.json})~\cite{Google_2025_A2A} is the contemporary example, decentralising the registry though not the trust fabric. \textbf{Decentralised/networked} discovery uses peer-to-peer mechanisms, ranging from simple network broadcasts to the Agent Network Protocol's bet on decentralised identifiers (DIDs) and P2P trust fabrics~\cite{W3C_ANP_2025}.

By applying these four dimensions, we can systematically position any given WoA implementation, whether a JADE-based multi-agent system, a Semantic Web agent, or an AutoGPT-style autonomous agent, and produce a clear comparative map of the field. This framework helps both in understanding the historical evolution and in surfacing the architectural trade-offs that continue to shape the future of the Web of Agents.

\begin{table*}[ht]
\centering
\caption{Application of the Four-Dimensional Comparative Framework to WoA Generations}
\label{tab:functional_taxonomy}
\begin{tabularx}{\textwidth}{|l|X|X|X|}
\hline
\textbf{Taxonomic Dimension} & \shortstack[l]{\textbf{Generation~I:}\\\textbf{FIPA-era MAS Agents}} & \shortstack[l]{\textbf{Generation~II:}\\\textbf{Semantic Web Agents}} & \shortstack[l]{\textbf{Generation~III:}\\\textbf{Modern LLM-based Agents}} \\
\hline
\textbf{Semantic Foundation} & \textbf{Procedural Semantics:} Meaning is derived from the formal semantics of communicative acts in FIPA ACL. & \textbf{Formal/Explicit Semantics:} Meaning is encoded in external, shared ontologies (RDF, OWL). & \textbf{Implicit/Emergent Semantics:} Meaning is inferred by the LLM's internal world model from natural language descriptions. \\
\hline
\textbf{Communication Paradigm} & \textbf{Performative-based:} Uses rich speech-act theory (FIPA ACL). & \textbf{Resource-Oriented:} Interacts with semantic web services and SPARQL endpoints over HTTP. & \textbf{RPC / Resource-Oriented:} Primarily uses lightweight RPC (MCP, A2A) and RESTful APIs for tool interaction. \\
\hline
\textbf{Locus of Intelligence} & \textbf{Intelligence-in-Platform:} The FIPA platform (Directory Facilitator, etc.) provides core coordination services. & \textbf{Intelligence-in-Data:} Agent intelligence relies on reasoning over annotated semantic data. & \textbf{Intelligence-in-Agent/Model:} The LLM itself is the primary locus of reasoning, planning, and understanding. \\
\hline
\textbf{Discovery Mechanism} & \textbf{Centralized Registry:} Relies on a platform-specific Directory Facilitator (DF) to find other agents. & \textbf{Centralised endpoints / querying:} SPARQL queries against known semantic repositories (DBpedia, Wikidata); functionally centralised, but no FIPA-style registry. & \textbf{Standardized Metadata File / Decentralized:} Moves towards decentralized discovery via well-known files (A2A's \texttt{agent.json}) or future P2P networks. \\
\hline
\end{tabularx}
\end{table*}

\paragraph{Planning paradigm: an intra-agent dimension coupled to the framework.}\label{sec:planning_algorithms}
The four dimensions above are deliberately \emph{inter-agent}. One intra-agent property, the choice of \emph{planning paradigm}, is sufficiently coupled to the Locus-of-Intelligence and Semantic-Foundation dimensions to warrant a brief remark. Classical and Hierarchical Task Network planners (PDDL, HATP~\cite{alejandre_2018hatp}) require structured domain models and fit architectures in which intelligence resides in external data and relies on formal semantics; hybrid approaches such as TWOSTEP~\cite{bai_2024twostep} use an LLM to translate informal goals into PDDL for a classical solver. Probabilistic planners (MDPs, POMDPs in BDI frameworks~\cite{bauters_2016probabilistic}) trade computational cost for principled uncertainty handling. LLM-native planners (ReAct~\cite{Yao_2023}, Tree of Thoughts~\cite{yao_2023treeofthoughts}) draw on the model's broad knowledge but provide weaker verifiability. The pattern is consistent: as the locus of intelligence moves from the data to the agent's own model, the planning paradigm moves from formally specified to learned and interleaved-with-action.

\subsection{Applying the Framework: Classification of Representative Systems}\label{sec:classification}

To demonstrate that the four-dimensional framework is operationally usable rather than purely descriptive, Table~\ref{tab:classification} classifies sixteen representative agent systems on the four dimensions. The selection spans all three generations and includes recent hybrid systems (NLWeb, GraphRAG-augmented agents, computer-use agents) that the literature has not previously placed on a single comparative axis. The sixteen are not balanced 1:1:1 across generations because Generation~III has undergone more architectural diversification than its predecessors: where Gen~I FIPA platforms and Gen~II Semantic Web agents largely shared a single architectural template, Gen~III has produced at least four internally distinct sub-types (pure LLM-orchestration, hybrid LLM--knowledge-graph, computer-use, and vertical software-engineering agents). We have chosen representative anchors for each Gen~I/II template and a wider sample of Gen~III sub-types; further Gen~I/II coverage (e.g., Cougaar, JACK, Jadex; KAoS, KAON) would re-iterate the same dimension assignments without adding analytical value. Several patterns are visible.

First, Generation~I and Generation~II systems cluster tightly. FIPA-era platforms share (Procedural semantics, Platform-locus) with variation in communication style (Performative for JADE / Agentcities, proprietary mobile-agent dispatch for Aglets) and discovery (centralised DF or federated registry). Semantic Web agents share (Formal semantics, Resource-oriented, Data-locus) with discovery through SPARQL querying or centralised endpoints rather than a FIPA-style registry. Each generation's failure mode (see~\S\ref{sec:taxonomy}) is visible in the table as an internal mismatch of dimensions.

Second, Generation~III shows greater heterogeneity. Pure LLM-agent frameworks (AutoGPT, LangChain, CrewAI) share (Implicit, RPC, Agent-locus) but use static configuration rather than standardised metadata-file discovery; production-tier platforms (Magentic-UI, NLWeb) take the further step of blending Implicit semantics with Formal structured metadata (Agent Cards, Schema.org annotations). Computer-use agents (Anthropic Computer Use, OpenAI Operator) sit at an extreme position: Implicit semantics, no protocol-level communication (keystrokes and pixels), Agent-localised intelligence, and navigation-driven discovery. Software-engineering agents (Devin, SWE-agent) are a vertical sub-class operating largely outside the discovery axis (they target a fixed repository rather than a peer network).

Third, the table makes visible the empirical claim from~\S\ref{sec:taxonomy}: dimensions are not fully independent, and modern hybrid systems combine values previously associated with different generations. This grounds the framework as a description of a continuous design space rather than a discrete partition.

\begin{table*}[!t]
\centering
\caption{Classification of sixteen representative agent systems on the four interoperability-facing dimensions. Hybrid entries combine values previously associated with different generations.}
\label{tab:classification}
\footnotesize
\begin{tabularx}{\textwidth}{|p{0.16\textwidth}|p{0.07\textwidth}|Y|Y|Y|Y|}
\hline
\textbf{System} & \textbf{Gen.} & \textbf{Semantic Foundation} & \textbf{Communication} & \textbf{Locus of Intelligence} & \textbf{Discovery} \\
\hline
JADE \cite{Bellifemine_1999} & I & Procedural (FIPA ACL) & Performative & Platform (DF, AMS) & Centralised registry (DF) \\
\hline
Aglets \cite{Lange_1998} & I & Procedural & Mobile-agent (proprietary) & Platform & Centralised \\
\hline
Agentcities \cite{Ciortea_2019} & I & Procedural + ad-hoc extensions & Performative & Platform-federated & Federated registry \\
\hline
Semantic Web agents \cite{Berners-Lee_2001} & II & Formal (RDF/OWL) & Resource-oriented & Data (knowledge graphs) & Querying / SPARQL endpoints \\
\hline
DBpedia-based agents \cite{bizer_2009} & II & Formal & Resource-oriented & Data & Centralised endpoint \\
\hline
AutoGPT \cite{AutoGPT_2023} & III & Implicit & RPC (tool calls) & Agent (LLM) & Static config \\
\hline
LangChain agents \cite{LangChainRepo} & III & Implicit & RPC + tool schemas & Agent & Static integration \\
\hline
MetaGPT \cite{MetaGPT_2024} & III & Implicit & Pub/sub message bus & Agent (role-specialised) & Static (SOP-driven) \\
\hline
CrewAI \cite{CrewAIRepo} & III & Implicit & RPC & Agent + orchestrator & Static \\
\hline
NLWeb \cite{nlweb} & III (hybrid) & Implicit + Schema.org & RPC (MCP) & Agent + structured site metadata & \texttt{/.well-known/} \\
\hline
Magentic-UI \cite{Microsoft_MagenticUI_2025} & III (hybrid) & Implicit + Agent Card schemas & RPC + streaming + HITL & Agent + UI panel & Agent Card \\
\hline
GraphRAG agents \cite{GraphRAG_2024} & III (hybrid) & Implicit + Formal (KG) & RPC & Agent + external KG & Static; KG queries \\
\hline
Anthropic Computer Use \cite{AnthropicComputer2024} & III (extreme) & Implicit (pixels/text) & UI-action stream (no protocol) & Agent (VM-local) & Navigation-driven \\
\hline
OpenAI Operator \cite{OpenAI_Operator_2025} & III (extreme) & Implicit (browser DOM/pixels) & UI-action stream (no protocol) & Agent (browser-local) & Navigation-driven \\
\hline
SWE-agent \cite{SWE_Agent_2024} & III (vertical) & Implicit + repo schema & RPC + filesystem & Agent & N/A (single-repo) \\
\hline
Devin \cite{Cognition_Devin_2024} & III (vertical) & Implicit + dev-env schema & RPC + filesystem + shell & Agent (long-horizon) & N/A (single-repo) \\
\hline
\end{tabularx}
\end{table*}

\section{The Modern Agent Stack and Its Interaction Protocols}\label{sec:stack}

Since the LLM breakthrough of late 2022~\cite{Brown_2020,Bubeck_2023}, the prospects for a Web of Agents have changed materially: practical agent interoperability now requires new web-native standards~\cite{Anthropic_2024_MCP,Google_A2A_Repo}. Several enabling technologies have converged to make this vision feasible: large-scale LLMs and reasoning models that give agents strong natural-language understanding and planning \cite{Brown_2020,Bubeck_2023,Yao_2023}; web-interfacing tools (DOM/XPath, RESTful API wrappers, browser automation via Selenium) and tool ecosystems such as OpenAI's ChatGPT plugins~\cite{OpenAI_Plugins_2023} and LangChain~\cite{LangChainRepo}; new training and adaptation methods including reinforcement learning, imitation learning, and retrieval-augmented generation (RAG) for grounding outputs in factual external data~\cite{Lewis_2020}; semantic technologies (RDF, OWL, and knowledge graphs) that complement unstructured data \cite{Berners-Lee_2001,Mika_2004}; and cloud-native infrastructure (containerisation, Kubernetes orchestration) for scalable deployment.

Early attempts at agent communication standards (KQML \cite{Finin_1994} and FIPA ACL \cite{Mascardi_2013,Manev_2014}) laid a foundation of structured message protocols with formal semantics, but they were oriented toward controlled environments and did not anticipate the scale and heterogeneity of the open Web. The failure of heavyweight, specialised stacks like FIPA taught the lesson that modern protocols operationalise: pragmatic simplicity and alignment with existing web standards (HTTP, JSON-RPC) beat agent-specific stacks, even at the cost of reduced formal semantic expressiveness.

The lower layers of the agent stack (containerisation, scheduling, intra-cluster service discovery) are largely solved problems inherited from cloud-native infrastructure: Kubernetes is the de-facto industry standard for deploying AI workloads \cite{k8sborgblog, gkeai}, and its \emph{Pod} and \emph{Service} abstractions map cleanly onto multi-component agents and onto intra-cluster discovery via stable virtual endpoints~\cite{collabnix}. The open questions, examined in the rest of this section, lie at the interaction-protocol and reasoning layers above; we note that \emph{open-web} agent discovery (across organisational and trust boundaries) is a distinct unsolved problem analysed in \S\ref{sec:challenges} (Challenge~\ref{ch:discovery-scaling}), not addressed by cloud-native primitives.

\subsection{LLMs and the Need for New Interaction Standards}

The rise of capable LLMs, exemplified by models such as OpenAI's GPT-4.5 \cite{OpenAI_GPT45_2025}, Google's Gemini 2.5 Pro \cite{Google_Gemini25Pro_2025}, Anthropic's Claude 3.7 Sonnet \cite{Anthropic_Claude37Sonnet_2025}, xAI's Grok-3 \cite{xAI_Grok3_2025}, Mistral Large~2 \cite{Mistral_Large2_2024}, and Alibaba's Qwen~3 \cite{Alibaba_Qwen3_2025}, has markedly advanced natural-language understanding and generation, reasoning and planning, and code generation \cite{Brown_2020,Bubeck_2023}. Combined with tool-use techniques such as Toolformer~\cite{Schick_2023} and ReAct~\cite{Yao_2023}, this has enabled AI agents to tackle much more complex and unstructured tasks than before. New training paradigms further improve agent reasoning and adaptability: reinforcement learning optimises decisions over long action sequences \cite{Christiano_2017}, imitation learning from human demonstrations provides strong priors for complex behaviours \cite{Ross_2011}, and retrieval-augmented generation (RAG) lets agents query external knowledge bases in real time to keep their responses grounded and up-to-date \cite{Lewis_2020}.

This new wave of AI agents nonetheless faces a major challenge: there are still no standardised methods for them to interact with external resources (websites, databases, APIs, and tools) or with one another. Developers have had to implement custom integrations for each agent and each external resource, typically through web scraping (parsing HTML via DOM or XPath), bespoke RESTful API wrappers, or browser automation frameworks like Selenium. This ad-hoc approach is neither efficient nor scalable and hinders integrated multi-agent systems. The gap motivates new protocols (MCP, A2A) that simplify and standardise such interactions; these are described in \S\ref{sec:MCP} and \S\ref{sec:A2A}, and a representative selection of frameworks built on top of them is compared in Table~\ref{tab:frameworks_comparison} (\S\ref{sec:llm_frameworks}).

\paragraph{Computer-use agents as a fallback integration path.} A complementary frontier of integration is \emph{computer-use agents}: systems that operate the same graphical user interfaces designed for human users, rather than calling structured APIs. Anthropic's Computer Use, released in October~2024, enables Claude to control a virtual machine through screenshots, mouse, and keyboard, providing a fallback path when neither an MCP server nor an A2A endpoint is available~\cite{AnthropicComputer2024}. OpenAI's Operator (January~2025) generalised the same approach to web browsing~\cite{OpenAI_Operator_2025}, while Microsoft's Magentic-UI couples computer-use with a human-in-the-loop panel that approves each irreversible action~\cite{Microsoft_MagenticUI_2025}. Computer-use agents shift the locus of integration from a server-side API contract to a client-side perception--action loop, with direct implications for the four comparative dimensions: discovery becomes navigation-driven rather than registry-driven, communication degenerates to keystrokes and pixels, and security loses the API authentication boundary. In the framework of this paper, computer-use represents an extreme of the \emph{semantics-in-models} paradigm, where the agent must interpret human-facing semantics on the fly with no protocol-level support.

\subsection{Model Context Protocol (MCP): Standardizing Model-to-Context Interaction}
\label{sec:MCP}

Anthropic released the Model Context Protocol (MCP) in November 2024 \cite{Anthropic_2024_MCP}. MCP is an open standard, a universal interface for connecting AI models to external data sources and tools. The aim of MCP is to standardize how applications (acting as MCP hosts or clients) supply context (data) and tools (actionable capabilities) to AI models, particularly LLMs \cite{Anthropic_2024_MCP}. This allows models to fetch up-to-date information needed for a given task and to trigger actions in external systems through a single, consistent interface.

The MCP architecture follows a Client–Host–Server pattern and uses the JSON-RPC 2.0 protocol for messaging \cite{Anthropic_2024_MCP}. In practice, Microsoft's NLWeb project exemplifies the MCP approach integrated directly into user-facing web applications \cite{nlweb}. Each NLWeb instance effectively acts as an MCP server embedded within a website, exposing that site's data and functionality through a standardized natural-language interface. This allows external AI agents (as well as human users) to discover and interact with the website's content and actions via MCP, demonstrating a real-world implementation of the Model Context Protocol in everyday web environments.

Three roles define the MCP architecture~\cite{Anthropic_2024_MCP}: the \textbf{host} is the AI-driven application (chatbot, IDE assistant); the \textbf{client} is a component inside the host maintaining a connection to a single MCP server; the \textbf{server} is a lightweight service exposing data or capabilities via the protocol. Servers expose three standardised primitives: \textbf{resources} (structured context data such as document snippets, code, search results), \textbf{prompts} (predefined instruction or query templates), and \textbf{tools} (executable functions the model can invoke, e.g., database queries, web searches, message sending)~\cite{Anthropic_2024_MCP}.

Operationally~\cite{Anthropic_2024_MCP,Docker_2025}, MCP collapses the $M{\times}N$ integration problem (connecting $M$ models with $N$ tools) to $M{+}N$; the protocol is vendor-neutral with a public reference implementation, supporting a growing ecosystem of servers for Google Drive, Slack, GitHub, databases, and similar services; and the specification mandates explicit user consent before an agent accesses data or invokes tools on the user's behalf.

MCP does not mandate the use of formal ontologies (such as RDF or OWL) to describe the semantics of resources or tools. Instead, meaning is conveyed through the structure and metadata of the JSON-RPC messages, rather than through any externally defined ontology or vector embedding space. This design choice simplifies implementation and offers flexibility, since new tool capabilities can be added by simply providing natural-language descriptions, instead of developing formal semantic models. The trade-off is reduced machine-level semantic interoperability and precision, as meaning remains encoded in unstructured language rather than in a formal representation.

\paragraph{Maturation of the specification (2025--2026).} The MCP specification has continued to evolve since its November 2024 launch. The November~2025 revision (specification version 2025-11-25) introduced three load-bearing capabilities: a formal \emph{Tasks} abstraction for long-running operations with progress tracking; \emph{OAuth~2.1} as a standardised authentication layer, replacing the earlier ad hoc consent model; and a centralised \emph{MCP Registry} providing discovery for publicly available tool servers~\cite{MCP_Tasks_2025}. Concurrently, MCP transitioned to community governance under a multi-stakeholder board, and the registry listed over 3{,}000 publicly registered tool servers as of early 2026. Adjacent specifications have begun to fill remaining gaps that fall outside MCP's core scope: WebMCP, accepted as a deliverable of the W3C Web Machine Learning Community Group in September~2025, extends MCP into browser-native execution~\cite{WebMCP_W3C_2025}, while AG-UI, the Agent--User Interaction protocol, standardises agent-to-frontend streaming, addressing a gap that neither MCP nor A2A specify~\cite{AGUI_2025}.

\subsection{Agent-to-Agent (A2A) Protocol: Enabling Inter-Agent Communication}
\label{sec:A2A}

Where MCP addresses an agent's interaction with its environment (data and tools), the Agent-to-Agent (A2A) protocol (introduced by Google in April 2025 \cite{Google_A2A_Repo}) targets another key challenge: enabling communication and coordination among the agents themselves. A2A is also an open protocol, developed in collaboration with over 50 partners \cite{Google_A2A_Repo}.

The goal of A2A is to allow AI agents, built on different frameworks and by different vendors, to communicate securely, exchange information, and coordinate their actions to tackle complex tasks that require joint effort. A2A is designed as a complement to MCP, not a replacement. If MCP is the ``wrench'' for tool access, then A2A is the ``mechanics' dialogue'' \cite{Google_A2A_Repo}.

A2A follows a client--server model applied to inter-agent interaction~\cite{Google_A2A_Repo}: a \textbf{client agent} initiates a task that a \textbf{remote agent} executes. Each agent advertises itself via an \textbf{Agent Card}---a JSON file at \texttt{/.well-known/agent.json} describing capabilities, endpoint URL, and authentication requirements. A \textbf{task} carries a unique identifier and progresses through states (\texttt{submitted}, \texttt{in\_progress}, \texttt{awaiting\_input}, \texttt{completed}, \texttt{failed}, \texttt{cancelled}); long-running tasks (minutes to hours) are supported via subscription (\texttt{tasks/sendSubscribe}), polling (\texttt{tasks/get}), or webhook (\texttt{tasks/pushNotification/set}). Messages between agents use a dialogue role structure (``user'', ``assistant'') and travel over HTTP(S) transport with JSON-RPC 2.0 request/response and Server-Sent Events for streaming.

Beyond the core entities~\cite{Google_A2A_Repo}, A2A piggybacks on widely adopted web standards rather than inventing new transport, uses the Agent Card mechanism for dynamic capability discovery, and includes built-in TLS plus authentication tokens for secure communication---essential given that agents may act on behalf of users or other systems.

A2A does not appear to mandate the use of formal OWL/ontology definitions for agent capabilities. Similar to MCP's philosophy, it opts for a pragmatic, JSON-based description (Agent Cards) to facilitate discovery. Compared to classical agent communication languages like FIPA ACL or KQML, A2A shares the same inter-agent communication goals but adopts a more lightweight, web-native approach with declarative discovery and interoperability built in \cite{Mascardi_2013}. In effect, A2A aims to do for heterogeneous AI agents what protocols like HTTP did for heterogeneous web servers and clients.

\paragraph{Institutional and technical evolution (2025--2026).} Since its initial release, A2A has matured through a parallel institutional and technical trajectory. In June~2025, Google donated the A2A specification to the Linux Foundation, simultaneously launching the broader project structure for agent interoperability standards~\cite{A2A_LinuxFoundation_2025}. By August~2025, the competing Agent Communication Protocol (ACP), initiated by IBM Research and later backed by Cisco, merged with A2A under the same umbrella, consolidating what had been three competing standards into one~\cite{ACP_Merger_2025}. In December~2025, the Linux Foundation formalised this consolidation through the launch of the Agentic AI Foundation (AAIF) as a neutral top-level foundation hosting MCP, A2A, and related infrastructure~\cite{AAIF_2025}. The protocol itself advanced from v0.3.0 (July~2025), which introduced streaming-first transport and enhanced Agent Cards with capability negotiation~\cite{A2A_v03_2025}, to v1.0.0 (January~2026), which marked the transition from experimental to production-ready status, introduced signed Agent Cards for cryptographic verification, and codified the discovery and trust mechanisms identified as gaps in earlier analyses~\cite{A2A_v10_2026}. As of early 2026, A2A Agent Cards were deployed by more than 50 organisations participating in the AAIF working group, providing the first credible cross-vendor agent interoperability deployment in three decades.

%\bigskip
%\hrule
%\bigskip

\subsection{Comparative Analysis of Interaction Protocols}

MCP and A2A address complementary needs: MCP standardises the agent--tool (vertical) interface; A2A standardises the agent--agent (horizontal) interface~\cite{Anthropic_2024_MCP,Google_2025_A2A}. The common deployment pattern is an orchestrator agent that delegates subtasks via A2A while invoking tools via MCP, with platforms such as AutoGen~\cite{AutoGenRepo}, LangChain~\cite{LangChain_2023}, and CrewAI~\cite{CrewAI_2024} providing the orchestration scaffolding. AWS Strands~\cite{AWSStrands_2025} and Microsoft Semantic Kernel~\cite{SemanticKernel_2023,SemanticKernelA2A_2024} are early platform-level integrations supporting both protocols. Neither protocol natively addresses trust, pricing, or accountability~\cite{Mascardi_2013}; these gaps drive the institutional work surveyed in \S\ref{sec:challenges}.

Table~\ref{tab:protocol_comparison} applies the four-dimensional framework of \S\ref{sec:taxonomy} to FIPA ACL, MCP, and A2A, making the architectural trade-off explicit. FIPA ACL achieves high semantic expressiveness through speech-act theory and BDI logic~\cite{Rao_1995}, but pays for it with platform dependency and a centralised Directory Facilitator, mismatched with the open Web's RESTful conventions~\cite{Ciortea_2019,Shafiq_2006}. MCP and A2A invert these choices: they accept low formal semantic expressiveness (capability descriptions are JSON schemas interpreted by an LLM rather than modal-logic ontologies) in exchange for web-native transport (HTTP, JSON-RPC, SSE) and lightweight, decentralised discovery (server capability advertisement; \texttt{/.well-known/agent.json} Agent Cards). The two protocols differ within Generation~III mainly in their target relationship: MCP's structured tool interface assumes the calling agent knows which tool to invoke (no built-in discovery), while A2A handles peer communication but assumes peer addresses are known a priori (no built-in marketplace). Both expose the same residual limitations of \emph{semantics-in-models} (non-verifiable tool semantics) and recreate the centralised-registry pattern at the discovery layer (MCP Registry; Agent Card hosting), problems analysed further in \S\ref{sec:challenges}.

\begin{table*}[!t]
 \caption{Comparison of FIPA ACL, Model Context Protocol (MCP), and Agent-to-Agent Protocol (A2A) along the four interoperability-facing dimensions of \S\ref{sec:taxonomy}}
 \label{tab:protocol_comparison}
 \centering
 \scriptsize
 \begin{tabularx}{\textwidth}{|
  p{0.18\textwidth}
| Y
| Y
| Y
|}
  \hline
  \textbf{Dimension}
   & \textbf{FIPA ACL}
   & \textbf{MCP}
   & \textbf{A2A} \\
  \hline
  \textbf{Semantic Foundation}
   & Procedural: formal modal-logic semantics for performatives; explicit shared ontologies required \cite{FIPA_2002}
   & Implicit / Emergent: capability semantics conveyed in JSON-RPC structure and natural-language tool descriptions \cite{Anthropic_2024_MCP}
   & Implicit / Emergent: capability semantics encoded in Agent Card JSON; no formal ontologies \cite{Google_2025_A2A} \\
  \hline
  \textbf{Communication Paradigm}
   & Performative-based (speech acts); asynchronous messaging over FIPA Message Transport Service
   & RPC (JSON-RPC 2.0); request--response and asynchronous interactions
   & RPC + streaming: JSON-RPC for request--response, Server-Sent Events for streaming, webhooks for push notifications \\
  \hline
  \textbf{Locus of Intelligence}
   & Intelligence-in-Platform: Directory Facilitator, Agent Management Service, and interaction protocols provide coordination
   & Intelligence-in-Agent / Model: the host LLM interprets tool descriptions and selects calls; server provides only capability exposure
   & Intelligence-in-Agent / Model: the client agent (LLM) plans, decomposes, and dispatches sub-tasks; remote agent is similarly model-driven \\
  \hline
  \textbf{Discovery Mechanism}
   & Centralised Registry: platform-bound Directory Facilitator
   & Centralised Registry (post-2025-11-25 spec) plus on-connect capability advertisement; the MCP Registry recreates a UDDI-style yellow-pages pattern \cite{MCP_Tasks_2025}
   & Standardised Metadata File: Agent Card served at \texttt{/.well-known/agent.json}; no built-in cross-registry federation \\
  \hline
 \end{tabularx}
\end{table*}

\subsection{Representative LLM-Agent Frameworks}\label{sec:llm_frameworks}

The AutoGPT release (2023)~\cite{AutoGPT_2023} marked the first widely visible autonomous LLM agent capable of setting its own sub-goals, invoking tools, and iteratively refining its approach. Its widely-shared demonstration catalysed development of a layered framework stack: \emph{data-layer} libraries that connect LLMs to external knowledge (LlamaIndex~\cite{LlamaIndexSite_2025}), \emph{application-logic} libraries that chain prompts, tools, and memory (LangChain~\cite{LangChainRepo}, AutoGen~\cite{AutoGenRepo}), and \emph{collaboration-layer} frameworks that orchestrate teams of role-specialised agents (CrewAI~\cite{CrewAIRepo}, MetaGPT~\cite{MetaGPT_2024}, SuperAGI~\cite{SuperAGIRepo}). Major cloud providers now offer proprietary counterparts (AWS Bedrock Agents~\cite{AmazonBedrock2023}, Microsoft Azure AI Agent Service~\cite{AzureAIAgent2024}, Google Vertex AI Agent Builder~\cite{GoogleAgentBuilder}, OpenAI Assistants~\cite{OpenAIAssistants2023}, IBM watsonx Orchestrate~\cite{IBMWatson2023}, Anthropic Claude with Computer Use~\cite{AnthropicComputer2024}), and an agent-marketplace tier is emerging (AWS Bedrock Marketplace~\cite{AWSBedrockMarketplace}, OpenAI GPT Store~\cite{OpenAIGPTStore}). Table~\ref{tab:frameworks_comparison} (Appendix~\ref{app:frameworks}) summarises six characteristics of four representative open-source frameworks (primary focus, autonomy, tool integration, scalability, orchestration, and protocol compatibility). AutoGen, SuperAGI, and LlamaIndex follow comparable patterns within the same layered structure and are therefore not separately tabulated.

\paragraph{Vertical specialisation: software-engineering agents.} The 2024--2025 emergence of \emph{coding agents}, including Devin~\cite{Cognition_Devin_2024}, SWE-agent~\cite{SWE_Agent_2024}, and developer-facing tools such as Aider and Cursor, demonstrated that autonomous agents can complete substantial portions of real engineering tickets given appropriate tooling. Benchmarks such as SWE-bench~\cite{SWE_Bench_2024} have become the de facto evaluation standard. Software-engineering agents are not Web-of-Agents systems \emph{per se} (they operate solo on a single repository), but they validate the broader thesis that LLM-powered agents, given proper tool interfaces, can sustain long-horizon work, and they have driven much of the engineering effort behind MCP.

\section{The Semantics-in-Data $\rightarrow$ Semantics-in-Models Shift}\label{sec:evolution}

The four-dimensional comparison developed in \S\ref{sec:taxonomy} and the modern protocol stack reviewed in \S\ref{sec:stack} together support a single high-level thesis. Across three decades the locus of \emph{semantic effort} has migrated systematically through three loci (platform, data, model); the central recent transition, from external data structures to the agent's own model, is what we name the \textbf{semantics-in-data $\rightarrow$ semantics-in-models} shift. We show below that this shift is \emph{predictive}: it accounts for each generation's adoption failure modes, identifies which primitives survived into Generation~III, and forecasts where the next migration is likely to occur.

\paragraph{The three loci.} Across the three generations the semantic effort sat in distinctly different places:
\begin{itemize}
    \item \textbf{Generation~I: semantics-in-platform.} FIPA-era MAS encoded meaning in procedural performative semantics within the platform (ACL speech acts, BDI logic, Directory Facilitator coordination). Stateful coordination was strong; substrate portability was weak.
    \item \textbf{Generation~II: semantics-in-data.} The Semantic Web aimed to make Web data carry formal, machine-readable meaning (RDF, OWL, SPARQL) so that comparatively simple agents could reason over it. Verifiability was high; flexibility and economic sustainability were low.
    \item \textbf{Generation~III: semantics-in-models.} LLM-era systems leave data largely unannotated and place the burden of interpretation on a learned model. Flexibility and zero-annotation cost are high; \emph{verifiability is lost}.
\end{itemize}
Figure~\ref{fig:paradigm-shift} summarises the three loci, the trade-off each generation accepted, and the predicted next migration toward \emph{semantics-in-verified-contracts}.

\paragraph{Why the shift is predictive.} The trajectory is not a one-shot event but a directional pattern with a structural cost. Each migration trades formal guarantees for adaptive capability, and the loss it incurs becomes the dominant unresolved problem of the next phase. The Semantic Web's verifiable ontologies were brittle and economically unsustainable; FIPA's stateful platforms were substrate-mismatched with the open Web; the LLM era's emergent semantics are flexible but non-verifiable. This pattern motivates the seven lessons collected in \S\ref{sec:findings} and predicts that the next migration, plausibly toward \emph{semantics-in-verified-contracts} or runtime-checked specifications, will trade flexibility for restored verifiability. Signed tool manifests, GraphRAG, and machine-checkable Agent Card schemas are early prototypes in this direction (see Challenge~\ref{ch:non-verifiable} below).

\begin{figure*}[!htbp]
\centering
\begin{tikzpicture}[
  font=\small,
  geni/.style={color=blue!45!black!75, line width=1pt},
  genii/.style={color=orange!70!brown!85, line width=1pt},
  geniii/.style={color=teal!75!black, line width=1pt},
  predict/.style={color=red!55!black!80, line width=1pt},
  panelI/.style={draw=blue!45!black!75, fill=blue!45!black!4, line width=0.8pt, rounded corners=2pt},
  panelII/.style={draw=orange!70!brown!85, fill=orange!70!brown!6, line width=0.8pt, rounded corners=2pt},
  panelIII/.style={draw=teal!75!black, fill=teal!75!black!5, line width=0.8pt, rounded corners=2pt},
  arr/.style={-{Stealth[length=3mm]}, line width=1.4pt}
]
% =========== Layout: 3 columns side-by-side ===========
%   Column width = 5.2 cm, gap = 0.4 cm, total width = 16.8 cm
%   Heights of header / locus / sections within each panel

% --- Generation I panel: PLATFORM ---
\begin{scope}[xshift=0cm]
  \node[panelI, anchor=north west, minimum width=5.2cm, minimum height=6.5cm] (panI) at (0,0) {};

  \node[geni, font=\bfseries\small, anchor=north] at (2.6,-0.15) {GENERATION~I};
  \node[geni, font=\itshape\scriptsize, anchor=north] at (2.6,-0.55) {FIPA-era MAS, 1995--2005};

  \node[geni, font=\bfseries\footnotesize, anchor=north] at (2.6,-1.2) {LOCUS: PLATFORM};

  \node[anchor=north west, font=\scriptsize\bfseries] at (0.3,-1.8) {Key technology};
  \node[anchor=north west, font=\scriptsize, align=left, text width=4.6cm] at (0.3,-2.15)
    {FIPA ACL, BDI, JADE, Directory Facilitator};

  \node[anchor=north west, font=\scriptsize\bfseries] at (0.3,-3.0) {Trade-off accepted};
  \node[anchor=north west, font=\scriptsize, align=left, text width=4.6cm] at (0.3,-3.35)
    {Strong stateful coordination, but substrate-mismatched with the open Web};

  \node[anchor=north west, font=\scriptsize\bfseries] at (0.3,-4.5) {Adoption};
  \node[anchor=north west, font=\scriptsize, align=left, text width=4.6cm] at (0.3,-4.85)
    {Stalled at open-web scale; survived in closed enterprise MAS};
\end{scope}

% --- Generation II panel: DATA ---
\begin{scope}[xshift=5.6cm]
  \node[panelII, anchor=north west, minimum width=5.2cm, minimum height=6.5cm] (panII) at (0,0) {};

  \node[genii, font=\bfseries\small, anchor=north] at (2.6,-0.15) {GENERATION~II};
  \node[genii, font=\itshape\scriptsize, anchor=north] at (2.6,-0.55) {Semantic Web, 2001--2012};

  \node[genii, font=\bfseries\footnotesize, anchor=north] at (2.6,-1.2) {LOCUS: DATA};

  \node[anchor=north west, font=\scriptsize\bfseries] at (0.3,-1.8) {Key technology};
  \node[anchor=north west, font=\scriptsize, align=left, text width=4.6cm] at (0.3,-2.15)
    {RDF, OWL, SPARQL, DBpedia, Schema.org};

  \node[anchor=north west, font=\scriptsize\bfseries] at (0.3,-3.0) {Trade-off accepted};
  \node[anchor=north west, font=\scriptsize, align=left, text width=4.6cm] at (0.3,-3.35)
    {High verifiability, but brittle and economically unsustainable annotation};

  \node[anchor=north west, font=\scriptsize\bfseries] at (0.3,-4.5) {Adoption};
  \node[anchor=north west, font=\scriptsize, align=left, text width=4.6cm] at (0.3,-4.85)
    {Partial: Schema.org / Wikidata / Knowledge Graph; universal vision unrealised};
\end{scope}

% --- Generation III panel: MODEL ---
\begin{scope}[xshift=11.2cm]
  \node[panelIII, anchor=north west, minimum width=5.2cm, minimum height=6.5cm] (panIII) at (0,0) {};

  \node[geniii, font=\bfseries\small, anchor=north] at (2.6,-0.15) {GENERATION~III};
  \node[geniii, font=\itshape\scriptsize, anchor=north] at (2.6,-0.55) {LLM-based agents, 2020s--};

  \node[geniii, font=\bfseries\footnotesize, anchor=north] at (2.6,-1.2) {LOCUS: MODEL};

  \node[anchor=north west, font=\scriptsize\bfseries] at (0.3,-1.8) {Key technology};
  \node[anchor=north west, font=\scriptsize, align=left, text width=4.6cm] at (0.3,-2.15)
    {Transformer, LLMs (GPT, Claude, Gemini), MCP, A2A, GraphRAG};

  \node[anchor=north west, font=\scriptsize\bfseries] at (0.3,-3.0) {Trade-off accepted};
  \node[anchor=north west, font=\scriptsize, align=left, text width=4.6cm] at (0.3,-3.35)
    {Flexible zero-annotation semantics, but verifiability is lost};

  \node[anchor=north west, font=\scriptsize\bfseries] at (0.3,-4.5) {Adoption};
  \node[anchor=north west, font=\scriptsize, align=left, text width=4.6cm] at (0.3,-4.85)
    {Mass adoption 2023--2026; production-grade after MCP Nov 2025 spec and A2A v1.0 (Jan 2026)};
\end{scope}

% --- Bottom migration arrow ---
% Three coloured segments under each generation column
\draw[arr, geni]   (0.3, -7.5) -- (5.2, -7.5);
\draw[arr, genii]  (5.6, -7.5) -- (10.8, -7.5);
\draw[arr, geniii] (11.2, -7.5) -- (16.0, -7.5);

\node[geni,   font=\bfseries\small, anchor=south] at (2.85, -7.4) {PLATFORM};
\node[genii,  font=\bfseries\small, anchor=south] at (8.20, -7.4) {DATA};
\node[geniii, font=\bfseries\small, anchor=south] at (13.6, -7.4) {MODEL};

% --- Predicted next migration (centred below the main arrow, fits within figure width) ---
\node[predict, font=\bfseries\scriptsize, anchor=east] at (7.8, -8.5)
  {Predicted next migration:};
\draw[arr, predict, dashed] (8.0, -8.5) -- (11.0, -8.5);
\node[predict, font=\bfseries\small, anchor=west] at (11.1, -8.5)
  {SEMANTICS-IN-VERIFIED-CONTRACTS};
% Sub-line: explain what Lesson 4 predicts — three markers aligned with Lesson 4 text in §sec:findings
\node[predict, font=\itshape\scriptsize, anchor=north, align=center] at (8.2, -8.95)
  {Lesson~4 (\S\ref{sec:findings}): signed manifests, cryptographic Agent Card verification,\\runtime contract enforcement; trades flexibility back for restored verifiability};
\end{tikzpicture}
\caption{\textbf{The Semantic-Effort Migration across three generations.}
The locus of semantic effort has migrated systematically in chronological order from agent platforms (Generation~I: FIPA-era MAS, 1995--2005), through external data structures (Generation~II: Semantic Web, 2001--2012), to learned models (Generation~III: LLM-based agents, 2020s--). Each migration trades formal guarantees for adaptive capability; the loss it incurs becomes the dominant unresolved problem of the next phase. The bottom arrow shows the directional pattern \textbf{platform $\rightarrow$ data $\rightarrow$ model}. Lesson~4 (\S\ref{sec:findings}) predicts that the next migration will move toward \emph{semantics-in-verified-contracts}, restoring verifiability without sacrificing model flexibility.}
\label{fig:paradigm-shift}
\end{figure*}

\paragraph{Hybrid retrieval and GraphRAG: structure re-enters the LLM workflow.} The distinction between \emph{semantics-in-data} and \emph{semantics-in-models} is a spectrum rather than a binary divide, and recent work combines the two paradigms in practice. Retrieval-Augmented Generation (RAG)~\cite{Lewis_2020} grounds LLM outputs in retrieved passages from external corpora, partially restoring the verifiability that pure semantics-in-models loses. The graph-aware variant \emph{GraphRAG}~\cite{GraphRAG_2024} goes further by injecting knowledge-graph structures and community summaries into LLM prompts at inference time, combining the precision of formal representations with the flexibility of implicit semantics. A growing line of hybrid retrieval work refines this direction further by targeting hallucination at the retrieval-selection stage, including self-reflective retrieval~\cite{Asai_2024_SelfRAG} and category-aware filtering~\cite{Petrova_CARRAG_2025}. The same pattern appears in the modern protocol stack: MCP's JSON Schema-style tool manifests~\cite{Anthropic_2024_MCP} and A2A's Agent Card metadata~\cite{Google_2025_A2A} both re-introduce structured, machine-checkable descriptions on top of LLM-interpreted semantics; Microsoft's NLWeb integration of Schema.org annotations with MCP servers~\cite{nlweb} is a concrete deployed example. Ontology-grounded tool descriptions, knowledge-graph-enhanced retrieval, and typed tool manifests indicate that the Semantic Web's technical legacy is being reintegrated with, rather than replaced by, the LLM paradigm. The distinction is therefore useful insofar as it identifies the \emph{primary locus of semantic engineering effort}; in operational systems, formal and implicit semantics increasingly coexist.

\paragraph{What Generation~III has delivered.} The realignment of the four dimensions around web-native primitives has produced concrete progress on each axis: LLMs supply flexible natural-language understanding and adaptive planning at scale \cite{Brown_2020,Bubeck_2023}; MCP and A2A standardise the agent--tool and agent--agent channels using HTTP and JSON-RPC, with the November~2025 MCP revision and A2A v1.0 (January~2026) marking the transition to production-grade specifications \cite{Anthropic_2024_MCP,Google_A2A_Repo,MCP_Tasks_2025,A2A_v10_2026}; a layered ecosystem of orchestration frameworks (LangChain, AutoGen, CrewAI, MetaGPT) and proprietary platforms (AWS Bedrock, Azure AI Agent Service, Vertex AI Agent Builder) has emerged on top of these protocols \cite{LangChainRepo,AutoGenRepo,CrewAIRepo,MetaGPT_2024}; and the paradigm is now exposed to end users through Microsoft's NLWeb \cite{nlweb} and Magentic-UI \cite{Microsoft_MagenticUI_2025}. Compared with the unrealised ambitions of the Semantic Web and FIPA eras, the current stack is pragmatic rather than universal, and that pragmatism is what has enabled adoption. The unresolved problems (analysed in \S\ref{sec:challenges}) now lie not at the architectural layer but at the socio-technical layer above it.

\section{Persistent Challenges}\label{sec:challenges}

The challenges of creating an open Web of Agents apply across all three architectural generations introduced in \S\ref{sec:origins}, but take a new and sharper form in Generation~III, where the LLM-era stack consolidates around centralised, proprietary ecosystems. We focus throughout on the inter-agent interoperability layer; complementary problems of joint policy learning and intra-team coordination form a parallel research literature on multi-agent reinforcement learning (MARL) that we treat as out of scope (see also \S\ref{sec:methodology}).

Major technology providers are launching ``walled garden'' agent ecosystems such as Amazon's Bedrock Marketplace \cite{AWSBedrockMarketplace}, OpenAI's GPT Store \cite{OpenAIGPTStore}, and Google's Vertex AI Agent Builder \cite{GoogleAgentBuilder} (a builder platform rather than a marketplace, but similarly platform-bound). These platforms accelerate deployment but sidestep the hard problems of decentralised trust, identity, and economic interoperability by imposing platform-centric governance. Decentralised alternatives are under active investigation: architectural concepts for open interoperability \cite{Rehm_2020_Interoperability}, agent-centric auction mechanisms \cite{Yekollu_2024_AgentMarketplace}, and DAO-based governance for dispute resolution \cite{ETHOS_2024} (the ETHOS framework, discussed in detail in \S\ref{sec:governance}). Without such decentralised alternatives, the agent ecosystem risks consolidation along incumbent-platform boundaries. The current landscape of decentralisation primitives is itself patchy: MCP and A2A handle communication but not economic coordination; browser-native extensions such as WebMCP help with discovery but do not scale to enterprise integrations~\cite{WebMCP_W3C_2025}; and classical Semantic Web ontologies remain too rigid for practical deployment.

A user-facing scenario such as booking a multi-leg conference trip across heterogeneous airline, accommodation, and ground-transport agents must run \emph{in the wild}: the participating agents must operate across the open Internet, cooperating or competing with one another, rather than relying on a closed ecosystem or a fixed set of pre-defined APIs. The remainder of this section organises the open problems along four pillars (trust and identity, economic models, security, governance) that the survey's findings (\S\ref{sec:findings}) and research agenda (\S\ref{sec:future}) will return to.

\subsection{Trust, Identity, and Discovery in Decentralised Networks}

Each of the four pillars below contains one or more \emph{generation-invariant problems} (Challenges~\ref{ch:ephemeral-identity}--\ref{ch:liability}): structural problems whose form has persisted across FIPA, Semantic Web, and LLM eras, even as the surface technologies have changed.

\paragraph{Challenge~1: The Ephemeral-Identity Crisis.}\refstepcounter{challenge}\label{ch:ephemeral-identity} A structural tension between agent lifecycle (short, task-instantiated, scalable) and trust accumulation (requires persistent identity): Gen~I FIPA Agent IDs were platform-bound and lost on restart; Gen~II Semantic Web agents reused user-level primitives (WebID, FOAF, OpenID) without a dedicated agent layer; Gen~III agent identifiers are instance tokens whose reputation cannot bind to a persistent ``self''. The tension endures because fast instantiation breaks identity persistence while enforced persistent identity limits scale. Self-Sovereign Identity built on W3C Decentralised Identifiers (\texttt{did:key}, \texttt{did:ion}, \texttt{did:dht}) and Verifiable Credentials~\cite{W3C_VC_2025}, complemented by industry proposals such as ``Know Your Agent'' (KYA)~\cite{KYA_Paradigm_2025}, is the current technical response, but a unified production deployment remains absent.

\paragraph{Challenge~2: The Discovery-Scaling Problem.}\refstepcounter{challenge}\label{ch:discovery-scaling} Open-web agent discovery cannot scale via centralised registries: FIPA Directory Facilitator (Gen~I), UDDI in the Web Services era, SPARQL endpoints and Schema.org (Gen~II), and now MCP Registry plus A2A Agent Cards (Gen~III) recapitulate the same centralised pattern. Centralisation creates single points of authority and failure, while decentralised peer-to-peer discovery requires expensive identity and trust infrastructure that protocols have not yet integrated. The Agent Network Protocol's DID-based decentralised discovery~\cite{W3C_ANP_2025} is the only structural attempt at scale but has not yet seen large-scale deployment.

\paragraph{Complementary research threads.} Recent proposals address adjacent gaps: zero-trust frameworks grounding agent security in decentralised identity and fine-grained access control~\cite{Huang_2025}; Intent-Based Networking for goal-oriented communication~\cite{Clemm_2022}; post-quantum cryptography to anticipate quantum-era threats~\cite{Liu_2024_PQCSurvey}; and computational-ethics mechanisms such as voting-based collective-norm alignment~\cite{Noothigattu_2018}.

\subsection{Economic Models and Incentive Alignment}\label{sec:economic}

\paragraph{Challenge~3: The Economic-Substrate Gap.}\refstepcounter{challenge}\label{ch:economic-substrate} Open agent ecosystems are unlikely to scale without an economic substrate for compensating service providers, agents, and infrastructure maintainers: the Semantic Web's universal-graph vision (2001--2012) failed because ontology maintenance had no revenue model~\cite{Shadbolt_BernersLee_Hall_2006,Simperl_Cuel_Stein_2013,INSEMTIVE_2008}; FIPA-era MAS lacked transaction primitives and never bound agent contracts to payment rails; Gen~III MCP and A2A still lack in-protocol economic primitives. Protocols govern communication, not commerce; without an integrated payment layer ecosystems either stay closed (walled-garden marketplaces) or collapse (Semantic Web). Multi-agent workflows consume considerably more compute than single-model calls~\cite{Anthropic_2025}; tokenisation supplies primitive economic mechanisms but requires formal modelling to avoid instability~\cite{Sadykhov_2023_DeTEcT}, and crypto-asset volatility complicates long-term service pricing~\cite{Yermack_2015}.

\paragraph{Payment-network entry into the agent economy (2025--2026).} The most concrete commercial development in this period has been the entry of major payment networks into agent-native commerce, addressing what was for three decades the WoA's missing economic layer. Visa's \emph{Trusted Agent Protocol}, announced in October~2025 in partnership with Cloudflare, wraps existing card-payment rails with agent-specific authentication and tokenised credentials bound to per-agent identifiers, with spending limits and merchant-category restrictions enforced by the network~\cite{Visa_TAP_2025}. Mastercard's \emph{Agent Pay} (April~2025) takes a complementary approach focused on subscription-based agent services and recurring micro-transactions~\cite{Mastercard_AgentPay_2025}. Stripe and OpenAI jointly announced the \emph{Agentic Commerce Protocol} (ACP, September~2025), integrating Stripe's payment infrastructure directly into agent workflows through a dedicated protocol layer co-published as an open specification~\cite{Stripe_OpenAI_ACP_2025}. These three initiatives mark the first time the economic substrate the FIPA and Semantic Web eras lacked has been supplied by incumbent payment infrastructure rather than expected to emerge bottom-up from agent research.

\paragraph{Challenge~4: The Reputation-Whitewashing Inevitability.}\refstepcounter{challenge}\label{ch:reputation} Reputation aligns incentives toward long-run reliability and prevents adverse selection (the classical ``market for lemons''~\cite{Akerlof_1970,Sabater_2005,Josang_2007}), but in ephemeral-identity settings reputation can be reset by spawning a new agent. Established MAS reputation models---FIRE combining interaction, role, witness, and certified reputation~\cite{Huynh_2006}; EigenTrust distributing local trust in P2P networks~\cite{Kamvar_2003}---all assume persistent identifiers, which the WoA structurally lacks (Challenge~\ref{ch:ephemeral-identity}). Gen~I FIPA platforms accumulated AID-bound reputation only within a single platform; Gen~II lacked any reputation layer; Gen~III makes Sybil attacks trivial with ephemeral DIDs. Cheap pseudonyms~\cite{Friedman_2001} make whitewashing strictly cheaper than maintaining a good reputation, except under persistent-identity binding that conflicts with scale. Federated and cryptographic countermeasures---signed feedback tokens tied to DIDs~\cite{Dogan_2023}, incentive-compatible truthful-reporting~\cite{Jurca_2005}, blockchain-based feedback storage~\cite{Dimitriou_2021}---are not yet deployed at open-web scale.

\paragraph{Adapting business models.} Autonomous agents disrupt traditional web business models. Digital marketplaces evolve into agent-to-agent bidding platforms; advertising-based revenue gives way to performance- or usage-based pricing because agents do not consume promotional content; and service design shifts toward API-first architectures whose primary customer is a software agent. Monetisation patterns include partner-API collaboration, standardised-API broad access, subscription tiers, and value-network effects~\cite{Masse_2016,Heshmatisafa_2023}.

\paragraph{Auxiliary-service economics.} Beyond primary services (e.g., a delegated credit card for booking a flight), a fully autonomous agent must also pay nanopayment-scale costs for service discovery, secondary specialist agents, and metered tool calls. Traditional strictly-transactional systems are too expensive at this scale; mechanisms such as the Cycle protocol's asynchronous channel rebalancing~\cite{Erradi_2022_Cycle} provide eventually-consistent micropayments suited to high-frequency machine-to-machine interactions.

\subsection{Security and Resilience in Adversarial Environments}

While the identity and economic dimensions cover \emph{who} an agent is and \emph{how} it pays, security covers \emph{what can go wrong in adversarial settings}. Agents operating on the open Web face a threat surface that combines classical web-application risks with LLM-specific attack vectors. We frame two of these as generation-invariant problems and discuss the remaining vectors as Generation~III-specific manifestations.

\paragraph{Challenge~5: The Non-Verifiable Tool Semantics Problem.}\refstepcounter{challenge}\label{ch:non-verifiable} When an LLM interprets a tool description there is no proof its interpretation matches the tool author's intent: any migration of semantic effort to learned models opens a verifiability gap (the direct cost of the semantics-in-models shift, \S\ref{sec:evolution}). Gen~I encoded interpretation in modal-logic ACL preconditions but only within FIPA platforms; Gen~II achieved formal verifiability via OWL inference at the cost of operational rigidity; Gen~III adopts maximum flexibility (NL-described tools, runtime interpretation) and pays in verifiability. The capability--verifiability trade-off is structural; restoring verifiability without sacrificing flexibility defines the next migration's design space. Emerging responses include signed tool manifests bound to cryptographic identity, runtime contract enforcement (pre/post-conditions), and machine-checkable Agent Card schemas beyond ad-hoc JSON.

\paragraph{Challenge~6: The Cross-Boundary Consent Asymmetry.}\refstepcounter{challenge}\label{ch:consent} User authority propagates through multi-hop delegation chains across organisational and jurisdictional boundaries the principal never explicitly approved at delegation time---among the most deployment-blocking issues in the current stack, alongside non-verifiable tool semantics (Challenge~\ref{ch:non-verifiable}). Gen~I FIPA had delegation protocols (Contract Net) but never reached the multi-organisation scale where consent propagation would have been tested; Gen~II Semantic Web agents never deployed across delegation chains; Gen~III A2A multi-hop workflows have no consent-propagation primitive. Per-hop authorisation tokens with capability narrowing require \emph{anticipatory} consent that a principal cannot reasonably grant at task-start time. The International AI Safety Report 2026 explicitly added consent-delegation chains and runtime drift in long-lived autonomous tasks to the international AI safety agenda~\cite{AISafetyReport_2026}.

\paragraph{Indirect prompt injection (Generation~III-specific).} Indirect prompt injection attacks are among the most sophisticated threat vectors against LLM-integrated applications. The 2023 HouYi framework demonstrated this concretely for single-LLM applications, successfully exploiting 31 of 36 real-world deployments \cite{Liu_2023}; subsequent work has shown that multi-agent settings amplify rather than dilute the risk, with malicious instructions propagating laterally between cooperating agents through shared context and delegation. These attacks embed malicious instructions within external data sources, producing document infiltration through contaminated emails and web pages, multimodal exploitation using steganography in images, and RAG poisoning through malicious documents in knowledge bases. Prompt injection is a direct symptom of Challenge~\ref{ch:non-verifiable}: a tool input that cannot be semantically validated cannot be safely interpreted.

\paragraph{Data poisoning.} Data-poisoning attacks have been studied from many angles; Koh and Liang~\cite{Koh_2017} adapted statistical influence functions to identify the most influential training examples for a given prediction, demonstrating that targeted poisoning is feasible without brute-force contamination.

\paragraph{Excessive agency.} OWASP's 2025 Top~10 for LLM Applications identifies \emph{excessive agency} through three vectors (excessive functionality, excessive permissions, and excessive autonomy) enabling privilege escalation, lateral movement, resource abuse, and unauthorised exfiltration~\cite{OWASP_GenAI_2025}. Excessive agency compounds with Challenge~\ref{ch:consent}: agents acting outside narrowly granted authority cannot have given valid consent for the over-scope actions.

\paragraph{Terms-of-service binding.} Agents may also analyse and agree to Terms of Service on behalf of their principals~\cite{UETA_1999}. For such an agreement to be legally binding, it must be provable and attributable to an authorised party, which in turn may drive a market-led standardisation of machine-readable service terms. Zero-Knowledge Proofs~\cite{Goldwasser_1989} let an agent prove that the principal meets specific TOS criteria (e.g., an age or jurisdiction requirement) without revealing the principal's identity~\cite{Cherif_2023,Gabizon_2019}.

\paragraph{Insurance and liability cover.} Insurance schemes could underwrite agents against categories of failure or misconduct, with the existence and terms of cover becoming a verifiable, machine-readable attribute factored into service discovery and bidding~\cite{Lior_2022}.

\paragraph{Authentication.} Strong authentication and authorisation remain essential. A typical interaction requires two distinct verifications: agent-to-agent authentication~\cite{Toth_2019} and principal-to-service authentication~\cite{EU_ID_2024}, the latter increasingly subject to GDPR-style data-minimisation requirements that ZKP-based selective disclosure can satisfy~\cite{EU_ID_2024}. Practical deployments combining cryptographic proofs with auxiliary biometric verification on distributed ledgers illustrate the engineering complexity such requirements impose in production environments~\cite{Pantiukhov_2024}.

\subsection{Governance and Ethical Alignment}\label{sec:governance}

\paragraph{Challenge~7: The Liability-Attribution Vacuum.}\refstepcounter{challenge}\label{ch:liability} When autonomous, possibly cross-jurisdictional agent chains cause harm, no settled legal framework attributes liability: in Gens~I and~II, agents remained in closed labs and consortium testbeds, so liability cases never arose at scale; in Gen~III, agents now execute commerce, agree to ToS on principals' behalf, and propagate decisions across delegation chains, while the legal frameworks attributing harm remain unsettled. Legal frameworks lag technology by 5--10 years in standard cycles, and the WoA's multi-hop, cross-jurisdictional structure (Challenge~\ref{ch:consent}) compounds the lag.

The ETHOS framework (Ethical Technology and Holistic Oversight System) is among the first architecturally-grounded governance proposals specifically designed for autonomous AI agents, organising compliance around blockchain-based global registries and smart contract automation \cite{ETHOS_2024}. ETHOS categorises agents into four risk tiers (unacceptable, high, moderate, and minimal) with corresponding compliance requirements enforced through zero-knowledge proofs and Soulbound Tokens for verifiable credentials.

The framework introduces the notion of \emph{AI legal entities} enabling autonomous systems to assume limited liability while ensuring accountability through insurance and compliance monitoring; this concept extends the contested ``electronic personhood'' proposal previously debated in the European Parliament and remains contested in current legal scholarship. Decentralised Autonomous Organisations (DAOs) provide multi-stakeholder governance structures involving developers, regulators, auditors, and ethicists in participatory decision-making processes. Decentralised justice and audit mechanisms require multi-stakeholder ecosystems incorporating internal company audits, independent third-party assessments, and participatory community oversight \cite{EUAI_2024}.

\paragraph{Regulatory hardening (2025--2026).} The regulatory landscape around autonomous agents has hardened materially in 2025--2026, moving from aspirational principles to enforceable obligations on concrete deadlines. The EU AI Act entered phased enforcement: prohibitions on unacceptable-risk systems took effect in February~2025, transparency obligations for general-purpose AI models activated in August~2025, and the full high-risk system requirements (Article~14 human oversight, conformity assessment, registration in the EU database) are scheduled to apply from August~2026~\cite{EUAI_2024} (anticipated rather than observed at the time of writing). In the United States, the Center for AI Standards and Innovation (CAISI) at the National Institute of Standards and Technology launched its \emph{AI Agent Standards Initiative} in February~2026~\cite{NIST_AgentStandards_2026}, signalling a federal posture toward agentic systems beyond the 2024 Generative AI Profile (NIST AI~600-1). At the international level, the \emph{International AI Safety Report 2026}, coordinated by Yoshua Bengio and signed by representatives of more than thirty governments, explicitly added multi-agent risks (cascading failures, consent-delegation chains, runtime drift in long-lived autonomous tasks) to the agenda of international AI safety policy~\cite{AISafetyReport_2026}. Together with the payment-network developments discussed in \S\ref{sec:economic} above, these moves indicate that the institutional layer the WoA has lacked is now being assembled in parallel by standards bodies, regulators, and commercial infrastructure providers, rather than being deferred to a single coordinating actor.

\subsection{Lessons from Three Decades}\label{sec:findings}

The analysis above yields seven named lessons grounded in cross-generational evidence (Tables~\ref{tab:functional_taxonomy} and \ref{tab:classification}, \S\S\ref{sec:origins}--\ref{sec:evolution}) and the November~2024--August~2026 institutional convergence analysed in \S\ref{sec:challenges}. Each lesson is stated with predictive consequence so that it remains evaluable as the agent stack evolves beyond the May~2026 cut-off.

\begin{enumerate}
\item \textbf{Lesson~1: The Dimensional-Alignment Principle.} Adoption depends on \emph{mutual compatibility} of the four interoperability dimensions, not on any single dimension's strength. A mismatch on any one dimension is sufficient to prevent open-web adoption. \emph{Evidence:} Gen~I's modal-logic ACL mismatched the Web's stateless substrate; Gen~II's formal semantics mismatched the ontology-cost economic substrate; Gen~III aligns all four around web-native primitives. \emph{Prediction:} future protocols that succeed on three dimensions but fail on the fourth are unlikely to survive at open-web scale; ANP's bet on DID-based discovery is a structural attempt to plug Gen~III's remaining mismatch (Challenge~\ref{ch:discovery-scaling}).

\item \textbf{Lesson~2: The Substrate--Standard Coupling Principle.} Inter-agent standards succeed only when their technical substrate matches the deployment substrate of their era; protocols built on era-orthogonal substrates require middleware bridges that mask rather than resolve the underlying mismatch. \emph{Evidence:} FIPA's specialised platform substrate versus the Web's HTTP substrate killed JADE-SOAP, Agent Web Gateway, and FIPA-OWL bridge efforts~\cite{Shafiq_2006,Cavallaro_2004,Pasha_2006}; MCP and A2A succeed because they piggyback on HTTP/JSON-RPC, the substrate that already dominates web deployment. \emph{Prediction:} when deployment substrate shifts (post-quantum networking, decentralised identifiers becoming default), today's protocols will face FIPA's fate unless they migrate; the choice between MCP/A2A and ANP is a bet on which substrate will dominate the 2030s Web.

\item \textbf{Lesson~3: The Economic-Substrate Necessity.} Open agent ecosystems are unlikely to scale without a concrete economic substrate; protocols and ontologies are necessary but insufficient. \emph{Evidence:} the Semantic Web's universal-graph vision failed to scale because ontology maintenance had no revenue model; FIPA never had commerce primitives; the LLM-era stack still lacks in-protocol economic primitives, supplied externally by Visa~TAP, Mastercard Agent Pay, and Stripe--OpenAI ACP in 2025. \emph{Prediction:} future protocols designed without a payment-integration plan will share the Sem-Web's fate; the 2025 payment-network entry is the first time in three decades the WoA has had this layer, and every other research pillar (security, trust, governance) is negotiable only because of this enabler.

\item \textbf{Lesson~4: The Semantic-Effort Migration.}\label{lesson:migration} The locus of semantic effort has migrated systematically across generations in chronological order: \emph{semantics-in-platform} (Gen~I, 1995--2005: ACL performative semantics interpreted by FIPA platforms) \(\rightarrow\) \emph{semantics-in-data} (Gen~II, 2001--2012: RDF/OWL ontologies authored by humans) \(\rightarrow\) \emph{semantics-in-models} (Gen~III, 2020s--: LLM-inferred from natural-language descriptions). Each migration trades formal guarantees for adaptive capability; the loss it incurs becomes the dominant unresolved problem of the next phase. \emph{Prediction:} an emerging fourth phase, which we term \emph{semantics-in-verified-contracts}, is already characterised by three early markers: signed manifests in MCP's November~2025 specification, cryptographic Agent Card verification in A2A~v1.0, and runtime contract enforcement (pre- and post-condition checking on tool outputs). The framing is falsifiable via quantitative markers: by 2028 we would expect (i)~at least half of publicly registered MCP servers to require signed tool manifests at parity with OAuth-style token verification; and (ii)~enforcement-gap, rather than semantic ambiguity, to dominate residual deployment problems (Challenge~\ref{ch:non-verifiable}). Failure of either marker to materialise would weaken the directional migration framing.

\item \textbf{Lesson~5: The Discovery-Centralisation Trap.} Every generation has reinvented centralised discovery, and every centralised discovery mechanism has failed at open-web scale. \emph{Evidence:} FIPA Directory Facilitator (Gen~I), UDDI (1999--2007), SPARQL endpoints / Schema.org (Gen~II), MCP Registry and A2A Agent Cards (Gen~III) recapitulate the same pattern. \emph{Prediction:} future centralised registries, including the MCP Registry as it grows, will face UDDI's fate; open-web discovery requires structurally decentralised primitives, not better registries (Challenge~\ref{ch:discovery-scaling}). ANP's DID-based discovery is the only structural attempt at scale.

\item \textbf{Lesson~6: The Capability--Verifiability Trade-off.} Each generation makes a structural trade-off between formal verifiability and operational flexibility; neither extreme works for the open Web. \emph{Evidence:} Gen~I's modal-logic ACL maximised verifiability but was operationally rigid; Gen~II's OWL reasoning was ontology-rigid; Gen~III's flexibility comes at the price of non-verifiable tool semantics (Challenge~\ref{ch:non-verifiable}). \emph{Prediction:} the research frontier is hybrid: verifiable subsets of LLM-mediated semantics. Signed tool manifests, runtime contract enforcement, GraphRAG, and machine-checkable Agent Card schemas are early experiments in this direction; computer-use agents (Challenge~\ref{ch:non-verifiable}) sit at the flexibility extreme and will drive the next round of governance work.

\item \textbf{Lesson~7: The Co-Evolutionary Regulation Effect.} For the first time in agent-system history, regulation is co-evolving with protocol design rather than lagging it. \emph{Evidence:} EU AI Act phased enforcement (Feb~2025--Aug~2026), NIST CAISI Agent Standards Initiative (Feb~2026), International AI Safety Report 2026 (Feb~2026), and AAIF formation (Dec~2025) all arrive while the protocol stack is still in v1.0, inverting the 5--10 year lag of Web, mobile, and IoT eras. \emph{Prediction:} technical architecture will be partly shaped by regulatory rather than purely engineering pressures; protocols ignoring EU AI Act Article~14 (human oversight) will be unusable in EU jurisdiction; ``governance-by-design'' is mandatory engineering work rather than optional add-on.
\end{enumerate}

\paragraph{Lessons and challenges are paired.} The seven lessons and the seven generation-invariant challenges introduced in \S\ref{sec:challenges} are dual views of the same evolutionary trajectory: each enduring challenge is the dominant unresolved problem produced by one or more of the lessons. Lesson~4 (Semantic-Effort Migration, the central thesis lesson) directly produces Challenge~\ref{ch:non-verifiable} (Non-Verifiable Tool Semantics); Lesson~6 (Capability--Verifiability Trade-off) bears on the same challenge from the trade-off side. Lesson~5 (Discovery-Centralisation Trap) produces Challenge~\ref{ch:discovery-scaling} (Discovery Scaling) directly and is also coupled to Challenge~\ref{ch:ephemeral-identity} (Ephemeral-Identity Crisis), since open-web discovery requires persistent identity. Lesson~1 (Dimensional-Alignment Principle) likewise reaches Challenge~\ref{ch:discovery-scaling} when discovery is the misaligned dimension, while Lesson~2 (Substrate--Standard Coupling Principle) reaches the same challenge via substrate-mismatched registries. Lesson~3 (Economic-Substrate Necessity) produces Challenge~\ref{ch:economic-substrate} (Economic-Substrate Gap) directly and Challenge~\ref{ch:reputation} (Reputation-Whitewashing Inevitability) indirectly, since reputation enforcement requires economic backing. Lessons~1 and~7 jointly produce Challenges~\ref{ch:consent} (Consent Asymmetry) and~\ref{ch:liability} (Liability-Attribution Vacuum). The pairing structure also makes the lessons evaluable: a lesson is falsified if its associated challenge dissolves without the predicted technical response. Figure~\ref{fig:lessons-challenges} visualises the full pairing matrix.

\begin{figure}[!htbp]
\centering
\begin{tikzpicture}[
  font=\scriptsize,
  cw/.style={minimum width=0.55cm, minimum height=0.50cm, draw=gray!45, line width=0.25pt, inner sep=0pt},
  dotnorm/.style={circle, fill=teal!75!black, inner sep=0pt, minimum size=0.30cm},
  dotstar/.style={circle, fill=red!65!black, inner sep=0pt, minimum size=0.42cm, label={[red!65!black,font=\tiny,inner sep=0.5pt]center:\textcolor{white}{$\star$}}},
  rowlbl/.style={anchor=east, font=\scriptsize},
  collbl/.style={anchor=south west, rotate=55, font=\scriptsize}
]
% --- Grid cells (7 rows x 7 cols) ---
\foreach \c in {1,...,7} {
  \foreach \r in {1,...,7} {
    \node[cw] at (\c*0.6, -\r*0.55) {};
  }
}

% --- Row labels (Lessons) ---
\node[rowlbl] at (0.30, -1*0.55) {\textbf{L1}~Dim-Align};
\node[rowlbl] at (0.30, -2*0.55) {\textbf{L2}~Substrate-Std};
\node[rowlbl] at (0.30, -3*0.55) {\textbf{L3}~Econ-Substrate};
\node[rowlbl] at (0.30, -4*0.55) {\textbf{L4}\,$\star$~Sem-Migration};
\node[rowlbl] at (0.30, -5*0.55) {\textbf{L5}~Discovery-Trap};
\node[rowlbl] at (0.30, -6*0.55) {\textbf{L6}~Cap-Verifiability};
\node[rowlbl] at (0.30, -7*0.55) {\textbf{L7}~CoEvol-Reg};

% --- Column labels (Challenges, rotated) ---
\node[collbl] at (1*0.6 - 0.15, -0.30) {\textbf{C1}~Ephem-Identity};
\node[collbl] at (2*0.6 - 0.15, -0.30) {\textbf{C2}~Discov-Scaling};
\node[collbl] at (3*0.6 - 0.15, -0.30) {\textbf{C3}~Econ-Gap};
\node[collbl] at (4*0.6 - 0.15, -0.30) {\textbf{C4}~Reputation};
\node[collbl] at (5*0.6 - 0.15, -0.30) {\textbf{C5}~Non-Verifiable};
\node[collbl] at (6*0.6 - 0.15, -0.30) {\textbf{C6}~Consent};
\node[collbl] at (7*0.6 - 0.15, -0.30) {\textbf{C7}~Liability};

% --- Pairings (12 dots; L4 -> C5 is the central thesis pair, marked as star) ---
\node[dotnorm] at (2*0.6, -1*0.55) {};  % L1 -> C2
\node[dotnorm] at (6*0.6, -1*0.55) {};  % L1 -> C6
\node[dotnorm] at (7*0.6, -1*0.55) {};  % L1 -> C7
\node[dotnorm] at (2*0.6, -2*0.55) {};  % L2 -> C2
\node[dotnorm] at (3*0.6, -3*0.55) {};  % L3 -> C3
\node[dotnorm] at (4*0.6, -3*0.55) {};  % L3 -> C4
\node[dotstar] at (5*0.6, -4*0.55) {};  % L4 -> C5 (central thesis)
\node[dotnorm] at (1*0.6, -5*0.55) {};  % L5 -> C1
\node[dotnorm] at (2*0.6, -5*0.55) {};  % L5 -> C2
\node[dotnorm] at (5*0.6, -6*0.55) {};  % L6 -> C5
\node[dotnorm] at (6*0.6, -7*0.55) {};  % L7 -> C6
\node[dotnorm] at (7*0.6, -7*0.55) {};  % L7 -> C7

% --- Legend (below the matrix) ---
\node[dotnorm] at (0.7, -7*0.55 - 0.75) {};
\node[anchor=west, font=\scriptsize] at (0.95, -7*0.55 - 0.75) {direct pairing};
\node[dotstar] at (2.8, -7*0.55 - 0.75) {};
\node[anchor=west, font=\scriptsize] at (3.1, -7*0.55 - 0.75) {central thesis pair (L4--C5)};
\end{tikzpicture}
\caption{\textbf{The seven Named Lessons paired with the seven Generation-Invariant Challenges.}
Filled cells mark pairings between lessons (rows~L1--L7) and challenges (columns~C1--C7). Each Lesson predicts one or more enduring challenges (rows read left to right); each Challenge is rooted in one or more Lessons (columns read top to bottom). The pairing structure makes the Lessons evaluable: a Lesson is falsified if its associated Challenge dissolves without the predicted technical response. Lesson~4 (Semantic-Effort Migration, $\star$) is the central thesis lesson and pairs directly with the Non-Verifiable Tool Semantics challenge (C5) that the migration produces as its structural cost. Lesson~1 (Dimensional-Alignment) is the most cross-cutting (three pairings); Lessons~3, 5, and~7 each pair with two challenges; Lessons~4 and~6 both reach the same challenge (C5) from different angles, anchoring the verifiability research frontier.}
\label{fig:lessons-challenges}
\end{figure}

\section{Future Research Directions}\label{sec:future}

The lessons and generation-invariant challenges above translate into four research threads, each anchored to the lesson(s) that motivate it. Each thread is calibrated to the institutional context of November~2024--August~2026.

\paragraph{(1) Verifiable agent semantics and signed tool descriptions} (addresses Challenge~\ref{ch:non-verifiable}; anchored to Lessons~\ref{lesson:migration} and~6). The semantics-in-models shift has lowered integration cost but at the price of non-verifiable tool semantics. Three open lines of work follow: signed and machine-checkable tool manifests that bind a capability description to a cryptographic identity; runtime contract enforcement (pre- and post-conditions checked against tool outputs); and standardised schemas for capability negotiation in A2A Agent Cards that go beyond ad-hoc JSON. This thread is the operational test of Lesson~\ref{lesson:migration}'s prediction that a fourth migration toward \emph{semantics-in-verified-contracts} is plausible.

\paragraph{(2) Consent-aware delegation chains} (addresses Challenge~\ref{ch:consent}; anchored to Lessons~1 and~7). Research priorities include consent-propagation metadata in A2A task envelopes; per-hop authorisation tokens with capability narrowing; formal models of multi-hop principal-to-service authentication; and integration with KYA-style verifiable credentials~\cite{KYA_Paradigm_2025}.

\paragraph{(3) Cost-transparent micro-economies and incentive-compatible reputation} (addresses Challenges~\ref{ch:economic-substrate} and~\ref{ch:reputation}; anchored to Lesson~3). Payment-network infrastructure has solved the \emph{how to pay} question; what remains unsolved is \emph{how to price}, \emph{how to signal quality}, and \emph{how to prevent reputation whitewashing} in ephemeral-identity settings. Concrete sub-problems: cost-transparency protocols that propagate pricing along delegation chains; incentive-compatible reputation mechanisms robust to Sybil attacks; and interoperable micropayment standards bridging Visa/Mastercard/Stripe rails to in-protocol nanopayments.

\paragraph{(4) Governance-by-design and cross-jurisdictional auditing} (addresses Challenge~\ref{ch:liability}; anchored to Lesson~7). The EU AI Act's high-risk requirements (anticipated to be operational from August~2026), NIST's CAISI initiative, and the International AI Safety Report 2026 will jointly drive demand for technical auditability primitives. Research priorities include automated compliance auditing inside protocol middleware; agent liability models (vicarious vs.~product-liability); standardised audit logs that survive multi-hop delegation; and architectural patterns that encode regulatory constraints into specifications rather than relying on after-the-fact policy enforcement.

\paragraph{Cross-cutting: evaluation and benchmarks.} LLM-based agents exhibit non-deterministic behaviour that conventional testing cannot characterise. Recent benchmarks (SWE-bench~\cite{SWE_Bench_2024}, agent-protocol benchmarks, and multi-agent evaluation suites) provide initial frameworks but do not yet capture security, economic, or governance dimensions; multi-dimensional evaluation methodology for open agent ecosystems remains an open challenge.

Table~\ref{tab:risks} consolidates the four pillars from \S\ref{sec:challenges}, mapping the principal threats in each domain to the research infrastructure required to address them (research directions from this section). The pillars are tightly coupled: a security breach erodes trust; without trust, agents will not transact, collapsing economic models; economic misalignment invites regulatory intervention; and regulation without economic viability is unenforceable.

\begin{table*}[!t]
\centering
\caption{A Summary of Systemic Risks for the Web of Agents}
\label{tab:risks}
\begin{tabularx}{\textwidth}{|l|X|X|}
\hline
\textbf{Risk Domain} & \textbf{Specific Threats \& Vulnerabilities} & \textbf{Required Research \& Infrastructure} \\
\hline
\textbf{Trust \& Identity} & Ephemeral Agent Identity Crisis, Sybil Attacks, Reputation Whitewashing, Lack of Verifiable Provenance. & Decentralised Identity (DID/VCs), Federated Reputation Systems (e.g., EigenTrust), Zero-Knowledge Proofs (ZKPs). \\
\hline
\textbf{Economic Viability} & Prohibitive Transaction Costs, Centralisation Risk, Incentive Misalignment, Market for Lemons. & Frictionless Micropayment Infrastructure, Stable Tokenomic Design, Robust Incentive Mechanisms. \\
\hline
\textbf{Security \& Resilience} & Indirect Prompt Injection (IPI), Excessive Agency (OWASP), Data/Model Poisoning, Cascading Failures. & Adaptive Defenses, Sandboxing, Human-in-the-Loop (HITL) for critical actions, Formal Verification. \\
\hline
\textbf{Governance \& Legality} & Liability Vacuum, Lack of Accountability, Regulatory Uncertainty, Emergence of ``Lawless'' Agents. & Sui Generis Legal Frameworks, AI Legal Personality, Decentralised Governance (DAOs), Automated Auditing. \\
\hline
\end{tabularx}
\end{table*}

\section{Conclusion}\label{sec:conclusion}

In this survey we have analysed the Web of Agents (WoA) concept, unifying three decades of research from Multi-Agent Systems (MAS), the Semantic Web, and modern LLM-based systems under a single thesis: the locus of \emph{semantic effort} has migrated systematically across generations in chronological order, from platform-side coordination (MAS, Generation~I) through data-side annotation (Semantic Web, Generation~II) to model-side interpretation (LLM-era, Generation~III). The central Gen~II~$\rightarrow$~Gen~III transition within this trajectory, which we call the \textbf{semantics-in-data $\rightarrow$ semantics-in-models} shift (\S\ref{sec:evolution}), is predictive: each generation's adoption failures and present open problems follow from where it located the semantic effort. The four-dimensional comparative framework (\S\ref{sec:taxonomy}) makes the thesis operational, classifying sixteen representative systems (Table~\ref{tab:classification}) and revealing that each prior generation failed not from a single deficit but from \emph{mismatches between dimensions} (Lesson~1).

\paragraph{Lessons and challenges as durable contributions.} The cross-generational analysis yields \textbf{seven named lessons} paired with \textbf{seven generation-invariant challenges} (\S\ref{sec:findings}). Each lesson is grounded in cross-generational evidence and stated with a predictive consequence that remains evaluable beyond the May~2026 cut-off; each paired challenge is a structural problem that persists regardless of which specific protocol prevails. The pairing matrix (Figure~\ref{fig:lessons-challenges}) makes the dual structure visually explicit.

\paragraph{Institutional convergence supplies the surrounding ecosystem.} The November~2024--August~2026 institutional convergence we surveyed marks a qualitative shift in the surrounding ecosystem. Neutral governance has consolidated around the Linux Foundation's Agentic AI Foundation (AAIF), absorbing MCP, A2A, and the former ACP into a single non-proprietary stack \cite{A2A_LinuxFoundation_2025, AAIF_2025, ACP_Merger_2025}. The economic layer absent throughout the FIPA and Semantic Web eras has been supplied by incumbent payment networks (Visa's Trusted Agent Protocol, Mastercard's Agent Pay, and the Stripe--OpenAI Agentic Commerce Protocol) \cite{Visa_TAP_2025, Mastercard_AgentPay_2025, Stripe_OpenAI_ACP_2025}. Regulatory frameworks have begun moving from principles to enforceable obligations: the EU AI Act's phased schedule, the NIST CAISI agent-standards initiative, and the International AI Safety Report 2026 \cite{EUAI_2024, NIST_AgentStandards_2026, AISafetyReport_2026}. These developments do not resolve the generation-invariant challenges; they merely \emph{enable} the research that can.

\paragraph{What the thesis characterises next.} If the semantic-effort migration is directional rather than coincidental, an emerging fourth phase is already trading flexibility back for restored verifiability, namely \emph{semantics-in-verified-contracts}, in which signed tool manifests (MCP November~2025 specification), cryptographic Agent Card verification (A2A~v1.0), and runtime contract enforcement constrain LLM interpretation without losing its adaptive reach. Lessons~4 (Semantic-Effort Migration) and~6 (Capability--Verifiability Trade-off) jointly characterise this trajectory; whether it consolidates by 2028 will be visible in concrete falsifiable markers (signed-manifest adoption share among MCP servers; enforcement-gap displacing semantic ambiguity as the dominant residual problem), stated in Lesson~\ref{lesson:migration} (\S\ref{sec:findings}).

The full realisation of the Web of Agents now depends less on building more capable individual agents and more on engineering a trustworthy and resilient \emph{ecosystem} in which they can operate. Progress requires sustained engagement across the agent-research community, standards bodies, regulators, and commercial infrastructure providers, all of which are now assembling parts of this ecosystem in parallel.

\section*{Acknowledgments}

The authors thank the maintainers of OpenAlex for providing the bibliometric data underlying Figure~\ref{fig:pub-volume}.

\paragraph*{Use of AI tools.} In preparing this manuscript, the authors used AI assistance (large language model dialogue systems) for editorial cleanup of prose, consistency checking across drafts, and refinement of TikZ figure source. All technical content, citations, claims, analytical decisions, and final wording remain the authors' own; the AI assistance was confined to clerical-style support and did not generate primary scientific content. No AI system is credited as an author, in line with IEEE Access policy.


%\bibliographystyle{IEEEtran} 
%\bibliography{references}  

\begin{thebibliography}{100}
\providecommand{\url}[1]{#1}
\csname url@samestyle\endcsname
\providecommand{\newblock}{\relax}
\providecommand{\bibinfo}[2]{#2}
\providecommand{\BIBentrySTDinterwordspacing}{\spaceskip=0pt\relax}
\providecommand{\BIBentryALTinterwordstretchfactor}{4}
\providecommand{\BIBentryALTinterwordspacing}{\spaceskip=\fontdimen2\font plus
\BIBentryALTinterwordstretchfactor\fontdimen3\font minus \fontdimen4\font\relax}
\providecommand{\BIBforeignlanguage}[2]{{%
\expandafter\ifx\csname l@#1\endcsname\relax
\typeout{** WARNING: IEEEtran.bst: No hyphenation pattern has been}%
\typeout{** loaded for the language `#1'. Using the pattern for}%
\typeout{** the default language instead.}%
\else
\language=\csname l@#1\endcsname
\fi
#2}}
\providecommand{\BIBdecl}{\relax}
\BIBdecl

\bibitem{OpenAI_GPT45_2025}
\BIBentryALTinterwordspacing
{OpenAI}, ``Introducing {GPT}-4.5,'' OpenAI Blog, February 27, 2025, accessed: 2026-05-14. [Online]. Available: \url{https://openai.com/index/introducing-gpt-4-5/}
\BIBentrySTDinterwordspacing
\bibitem{Google_Gemini25Pro_2025}
\BIBentryALTinterwordspacing
{Google DeepMind}, ``{Gemini 2.5: Our newest Gemini model with thinking},'' Google DeepMind Blog, March 28, 2025, accessed: 2026-05-14. [Online]. Available: \url{https://blog.google/innovation-and-ai/models-and-research/google-deepmind/gemini-model-thinking-updates-march-2025/}
\BIBentrySTDinterwordspacing
\bibitem{Anthropic_Claude37Sonnet_2025}
\BIBentryALTinterwordspacing
{Anthropic}, ``{Claude 3.7 Sonnet and Claude Code},'' Anthropic Blog, February 24, 2025, accessed: 2026-05-14. [Online]. Available: \url{https://www.anthropic.com/news/claude-3-7-sonnet}
\BIBentrySTDinterwordspacing
\bibitem{xAI_Grok3_2025}
\BIBentryALTinterwordspacing
{xAI}, ``{Grok 3 Beta — The Age of Reasoning Agents},'' xAI Blog, February 19, 2025, accessed: 2026-05-14. [Online]. Available: \url{https://x.ai/news/grok-3}
\BIBentrySTDinterwordspacing
\bibitem{Mistral_Large2_2024}
\BIBentryALTinterwordspacing
{Mistral AI}, ``{Large Enough: Mistral Large 2},'' Mistral AI News, July 24, 2024, accessed: 2026-05-14. [Online]. Available: \url{https://mistral.ai/news/mistral-large-2407}
\BIBentrySTDinterwordspacing
\bibitem{Alibaba_Qwen3_2025}
\BIBentryALTinterwordspacing
{Alibaba DAMO Academy}, ``{Alibaba unveils Qwen 3: A family of hybrid AI reasoning models},'' April 28, 2025, accessed: 2026-05-14. [Online]. Available: \url{https://www.alibabacloud.com/blog/alibaba-introduces-qwen3-setting-new-benchmark-in-open-source-ai-with-hybrid-reasoning_602192}
\BIBentrySTDinterwordspacing
%//------------------------------------------------------------
\bibitem{wei_2022}
\BIBentryALTinterwordspacing
J.~Wei, Y.~Tay, R.~Bommasani, C.~Raffel, B.~Zoph, S.~Borgeaud, D.~Yogatama, M.~Bosma, D.~Zhou, D.~Metzler, E.~H. Chi, T.~Hashimoto, O.~Vinyals, P.~Liang, J.~Dean, and W.~Fedus, ``Emergent abilities of large language models,'' \emph{Transactions on Machine Learning Research}, August 2022, accessed: 2026-05-14. [Online]. Available: \url{https://openreview.net/forum?id=yzkSU5zdwD}
\BIBentrySTDinterwordspacing
\bibitem{shojaee_2025illusion}
\BIBentryALTinterwordspacing
P.~Shojaee, I.~Mirzadeh, K.~Alizadeh, M.~Horton, S.~Bengio, and M.~Farajtabar, ``The illusion of thinking: Understanding the strengths and limitations of reasoning models via the lens of problem complexity,'' in \emph{Advances in Neural Information Processing Systems (NeurIPS)}, 2025. [Online]. Available: \url{https://arxiv.org/abs/2506.06941}
\BIBentrySTDinterwordspacing
\bibitem{OpenAI_FunctionCalling_2023}
\BIBentryALTinterwordspacing
{OpenAI}, ``{Function calling and other API updates},'' OpenAI Blog, June~13, 2023, accessed: 2026-05-14. [Online]. Available: \url{https://openai.com/index/function-calling-and-other-api-updates/}
\BIBentrySTDinterwordspacing
\bibitem{Anthropic_2024_MCP}
\BIBentryALTinterwordspacing
{Anthropic}, ``{Model Context Protocol} ({MCP}) specification,'' November 2024, accessed: 2026-05-14. [Online]. Available: \url{https://modelcontextprotocol.io/introduction}
\BIBentrySTDinterwordspacing
\bibitem{Google_2025_A2A}
\BIBentryALTinterwordspacing
{Google}, ``{Agent-to-Agent} ({A2A}) protocol specification,'' 2025, accessed: 2026-05-14. [Online]. Available: \url{https://a2aproject.github.io/A2A/latest/}
\BIBentrySTDinterwordspacing
\bibitem{Weiss_1999}
G.~Weiss, Ed., \emph{Multiagent Systems: A Modern Approach to Distributed Artificial Intelligence}.\hskip 1em plus 0.5em minus 0.4em\relax Cambridge, MA: MIT Press, 1999.
\bibitem{Berners-Lee_2001}
T.~Berners-Lee, J.~Hendler, and O.~Lassila, ``The semantic web,'' \emph{Scientific American}, vol. 284, no.~5, pp. 34--43, May 2001.
\bibitem{Matthews_2005}
\BIBentryALTinterwordspacing
D.~Brickley, S.~Buswell, B.~Matthews, L.~Miller, D.~Reynolds, and M.~Wilson, ``Semantic web technologies,'' JISC Technology and Standards Watch, Tech. Rep. TSW 05-02, 2005, accessed: 2026-05-14. [Online]. Available: \url{https://epubs.stfc.ac.uk/manifestation/653/jisctsw_05_02pdf.pdf}
\BIBentrySTDinterwordspacing
\bibitem{Malik_2018}
\BIBentryALTinterwordspacing
N.~Malik and S.~K. Malik, ``Semantic web as the next generation smart web,'' \emph{Journal of Emerging Technologies and Innovative Research (JETIR)}, vol.~5, no.~6, pp. 580--594, 2018, accessed: 2026-05-14. [Online]. Available: \url{https://www.jetir.org/papers/JETIR1806326.pdf}
\BIBentrySTDinterwordspacing
\bibitem{Mika_2004}
\BIBentryALTinterwordspacing
P.~Mika and J.~Akkermans, ``Towards a new synthesis of ontology technology and knowledge management,'' \emph{The Knowledge Engineering Review}, vol.~19, no.~4, pp. 317--345, 2004, accessed: 2026-05-14. [Online]. Available: \url{https://doi.org/10.1017/S0269888905000305}
\BIBentrySTDinterwordspacing
\bibitem{Brickley_2004}
\BIBentryALTinterwordspacing
D.~Brickley and R.~V. Guha, ``{RDF} vocabulary description language 1.0: {RDF} schema,'' W3C Recommendation, February 2004, accessed: 2026-05-14. [Online]. Available: \url{https://www.w3.org/TR/2004/REC-rdf-schema-20040210/}
\BIBentrySTDinterwordspacing
\bibitem{McGuinness_2004}
\BIBentryALTinterwordspacing
D.~L. McGuinness and F.~van Harmelen, ``{OWL} web ontology language overview,'' W3C Recommendation, February 2004, accessed: 2026-05-14. [Online]. Available: \url{https://www.w3.org/TR/2004/REC-owl-features-20040210/}
\BIBentrySTDinterwordspacing

%//------------------------------------------------------------
\bibitem{Ciortea_2019}
\BIBentryALTinterwordspacing
A.~Ciortea, S.~Mayer, F.~Gandon, O.~Boissier, A.~Ricci, and A.~Zimmermann, ``A decade in hindsight: The missing bridge between multi-agent systems and the world wide web,'' in \emph{Proceedings of the 18th International Conference on Autonomous Agents and Multiagent Systems (AAMAS)}.\hskip 1em plus 0.5em minus 0.4em\relax Montreal, Canada: International Foundation for Autonomous Agents and Multiagent Systems, 2019, pp. 1659--1663, accessed: 2026-05-14. [Online]. Available: \url{https://hal.science/emse-02070625/}
\BIBentrySTDinterwordspacing
\bibitem{Sapkota_2025}
\BIBentryALTinterwordspacing
R.~Sapkota, K.~I. Roumeliotis, and M.~Karkee, ``AI agents vs. agentic AI: A conceptual taxonomy, applications and challenges,'' \emph{Information Fusion}, 2025, doi: 10.1016/j.inffus.2025.103599. [Online]. Available: \url{https://doi.org/10.1016/j.inffus.2025.103599}
\BIBentrySTDinterwordspacing
\bibitem{Acharya_2025}
\BIBentryALTinterwordspacing
D.~B. Acharya, K.~Kuppan, and B.~Divya, ``Agentic {AI}: Autonomous intelligence for complex goals---a comprehensive survey,'' \emph{IEEE Access}, vol.~13, pp. 18\,912--18\,936, 2025, doi: 10.1109/ACCESS.2025.3532853. [Online]. Available: \url{https://doi.org/10.1109/ACCESS.2025.3532853}
\BIBentrySTDinterwordspacing
\bibitem{Ferrag_2025}
\BIBentryALTinterwordspacing
M.~A. Ferrag, N.~Tihanyi, and M.~Debbah, ``{From LLM reasoning to autonomous AI agents: A comprehensive review},'' arXiv preprint arXiv:2504.19678, 2025, accessed: 2026-05-14. [Online]. Available: \url{https://arxiv.org/abs/2504.19678}
\BIBentrySTDinterwordspacing
\bibitem{Yang_2023}
\BIBentryALTinterwordspacing
L.~Wang, C.~Ma, X.~Feng, Z.~Zhang, H.~Yang, J.~Zhang, Z.~Chen, J.~Tang, X.~Chen, Y.~Lin, W.~X. Zhao, Z.~Wei, and J.-R. Wen, ``A survey on large language model based autonomous agents,'' \emph{Frontiers of Computer Science}, vol.~18, no.~6, art.~186345, 2024, doi: 10.1007/s11704-024-40231-1. [Online]. Available: \url{https://doi.org/10.1007/s11704-024-40231-1}
\BIBentrySTDinterwordspacing
\bibitem{Ehtesham_2025}
\BIBentryALTinterwordspacing
A.~Ehtesham, A.~Singh, G.~K. Gupta, and S.~Kumar, ``{A survey of agent interoperability protocols: Model Context Protocol (MCP), Agent Communication Protocol (ACP), Agent-to-Agent Protocol (A2A), and Agent Network Protocol (ANP)},'' arXiv preprint arXiv:2505.02279, 2025, accessed: 2026-05-14. [Online]. Available: \url{https://arxiv.org/abs/2505.02279}
\BIBentrySTDinterwordspacing
\bibitem{W3C_ANP_2025}
\BIBentryALTinterwordspacing
{Agent Network Protocol Contributors}, ``{Agent Network Protocol Technical White Paper: Towards an Open Internet of Agents},'' 2025, accessed: 2026-05-14. [Online]. Available: \url{https://agentnetworkprotocol.com/en/specs/01-agentnetworkprotocol-technical-white-paper/}
\BIBentrySTDinterwordspacing
%//------------------------------------------------------------
\bibitem{Schneider_2025}
J.~Schneider, ``{Generative to agentic AI: Survey, conceptualization, and challenges},'' \emph{arXiv preprint arXiv:2504.18875}, 2025. [Online]. Available: \url{https://arxiv.org/abs/2504.18875}
\bibitem{Sharma_2025}
R.~Sharma, M.~de~Vos, P.~Chari, R.~Raskar, and A.~Kermarrec, ``{Collaborative agentic AI needs interoperability across ecosystems},'' \emph{arXiv preprint arXiv:2505.21550}, 2025. [Online]. Available: \url{https://arxiv.org/abs/2505.21550}
\bibitem{PRISMA_2021}
\BIBentryALTinterwordspacing
M.~J. Page, J.~E. McKenzie, P.~M. Bossuyt, I.~Boutron, T.~C. Hoffmann, C.~D. Mulrow, L.~Shamseer, J.~M. Tetzlaff, E.~A. Akl, S.~E. Brennan, R.~Chou, J.~Glanville, J.~M. Grimshaw, A.~Hr{\'o}bjartsson, M.~M. Lalu, T.~Li, E.~W. Loder, E.~Mayo-Wilson, S.~McDonald, L.~A. McGuinness, L.~A. Stewart, J.~Thomas, A.~C. Tricco, V.~A. Welch, P.~Whiting, and D.~Moher, ``{The PRISMA 2020 statement: an updated guideline for reporting systematic reviews},'' \emph{BMJ}, vol.~372, art.~n71, 2021, doi: 10.1136/bmj.n71. [Online]. Available: \url{https://doi.org/10.1136/bmj.n71}
\BIBentrySTDinterwordspacing
\bibitem{Huhns_1987}
\BIBentryALTinterwordspacing
M.~N. Huhns, ``Distributed artificial intelligence,'' in \emph{Distributed artificial intelligence}, L.~Gasser and M.~N. Huhns, Eds.\hskip 1em plus 0.5em minus 0.4em\relax Morgan Kaufmann, 1989, pp. 1--23.
\BIBentrySTDinterwordspacing
\bibitem{Rao_1995}
A.~S. Rao and M.~P. Georgeff, ``{BDI agents: From theory to practice},'' in \emph{Proceedings of the First International Conference on Multi-Agent Systems (ICMAS)}.\hskip 1em plus 0.5em minus 0.4em\relax AAAI Press, 1995, pp. 312--319.
\bibitem{Finin_1994}
\BIBentryALTinterwordspacing
T.~Finin, R.~Fritzson, D.~McKay, and R.~McEntire, ``{KQML as an agent communication language},'' in \emph{Proceedings of the Third International Conference on Information and Knowledge Management (CIKM '94)}.\hskip 1em plus 0.5em minus 0.4em\relax ACM Press, 1994, pp. 456--463, accessed: 2026-05-14. [Online]. Available: \url{https://dl.acm.org/doi/10.1145/191246.191322}
\BIBentrySTDinterwordspacing
\bibitem{FIPA_2002}
\BIBentryALTinterwordspacing
{Foundation for Intelligent Physical Agents}, ``{FIPA ACL Message Structure Specification},'' FIPA Standard SC00061G, December 2002, accessed: 2026-05-14. [Online]. Available: \url{https://web.archive.org/web/2020*/http://www.fipa.org/specs/fipa00061/SC00061G.html}
\BIBentrySTDinterwordspacing
\bibitem{Bellifemine_1999}
\BIBentryALTinterwordspacing
F.~Bellifemine, A.~Poggi, and G.~Rimassa, ``{JADE: A FIPA-compliant agent framework},'' in \emph{Proceedings of the 4th International Conference on The Practical Application of Intelligent Agents and Multi-Agent Systems (PAAM'99)}, London, UK, 1999, pp. 97--108, accessed: 2026-05-14. [Online]. Available: \url{https://jade.tilab.com/papers/PAAM99.pdf}
\BIBentrySTDinterwordspacing
\bibitem{Lange_1998}
D.~B. Lange and M.~Oshima, \emph{{Programming and Deploying Java Mobile Agents with Aglets}}.\hskip 1em plus 0.5em minus 0.4em\relax Boston, MA: Addison-Wesley, 1998.
\bibitem{Jennings_1995}
M.~Wooldridge and N.~R. Jennings, ``Intelligent agents: Theory and practice,'' \emph{The Knowledge Engineering Review}, vol.~10, no.~2, pp. 115--152, 1995.
\bibitem{AutoGPT_2023}
{Significant-Gravitas}, ``{AutoGPT}: An autonomous {GPT} agent,'' GitHub repository, 2023, accessed: 2026-05-14. [Online]. Available: \url{https://github.com/Significant-Gravitas/AutoGPT}
\bibitem{Wooldridge_2009}
M.~Wooldridge, \emph{An introduction to {MultiAgent} systems}, 2nd~ed.\hskip 1em plus 0.5em minus 0.4em\relax Chichester, UK: John Wiley \& Sons, 2009.
\bibitem{Lesser_1989}
V.~R. Lesser, ``A retrospective view of FA/C distributed problem solving,'' \emph{IEEE Transactions on Systems, Man, and Cybernetics}, vol.~21, no.~6, pp. 1347--1362, Dec. 1991.
\bibitem{FIPA_History}
\BIBentryALTinterwordspacing
S.~Poslad, ``History of {FIPA},'' {Foundation for Intelligent Physical Agents}, Queen Mary, University of London, last updated 20 December 2005, accessed: 2026-05-14. [Online]. Available: \url{https://web.archive.org/web/2020*/http://www.fipa.org/subgroups/ROFS-SG-docs/History-of-FIPA.htm}
\BIBentrySTDinterwordspacing
%//------------------------------------------------------------
\bibitem{Poslad_2007}
S.~Poslad, ``{Specifying protocols for multi-agent systems interaction},'' \emph{ACM Transactions on Autonomous and Adaptive Systems}, vol.~2, no.~4, Article 15, pp. 15:1--15:24, November 2007.
\bibitem{Mascardi_2013}
V.~Mascardi, J.~Hendler, and L.~Papaleo, ``{Semantic web and declarative agent languages and technologies: Current and future trends},'' in \emph{Declarative Agent Languages and Technologies X}, ser. Lecture Notes in Computer Science, M.~Baldoni, L.~Dennis, V.~Mascardi, and W.~Vasconcelos, Eds.\hskip 1em plus 0.5em minus 0.4em\relax Berlin, Heidelberg: Springer, 2013, vol. 7784, pp. 197--202.
\bibitem{Manev_2014}
T.~Manev and S.~Filiposka, ``{Semantic aware multi-agent system advantages},'' \emph{International Journal of Informatics and Communication Technology (IJ-ICT)}, vol.~3, no.~1, pp. 1--12, February 2014.
\bibitem{Shafiq_2006}
O.~Shafiq, Y.~Ding, and D.~Fensel, ``{Bridging multi agent systems and web services: Towards interoperability between software agents and semantic web services},'' in \emph{Proceedings of the 10th {IEEE} International Enterprise Distributed Object Computing Conference ({EDOC} '06)}.\hskip 1em plus 0.5em minus 0.4em\relax Hong Kong, China: IEEE Computer Society, October 2006, pp. 85--96.
\bibitem{Cavallaro_2004}
F.~Cavallaro, F.~Giunchiglia, and I.~Liccardi, ``{A middleware architecture for semantic web services},'' in \emph{The Semantic Web: Research and Applications}, ser. Lecture Notes in Computer Science, C.~Bussler, J.~Davies, D.~Fensel, and R.~Studer, Eds., vol. 3053.\hskip 1em plus 0.5em minus 0.4em\relax Berlin, Heidelberg: Springer, 2004, pp. 67--81.
\bibitem{Pasha_2006}
\BIBentryALTinterwordspacing
M.~Pasha, S.~Rehman, A.~Ali, H.~F. Ahmad, and H.~Suguri, ``{Middleware between {OWL} and {FIPA} ontologies in the semantic grid environment},'' in \emph{Proceedings of the 2006 International Conference on Semantic Web and Web Services ({SWWS} '06)}.\hskip 1em plus 0.5em minus 0.4em\relax Las Vegas, NV, USA: CSREA Press, June 2006, pp. 30--35, accessed: 2026-05-14. [Online]. Available: \url{https://dblp.org/rec/conf/swws/PashaRAAS06.html}
\BIBentrySTDinterwordspacing
\bibitem{Beckett_2004}
\BIBentryALTinterwordspacing
F.~Manola and E.~Miller, ``{RDF} primer,'' {W3C} Recommendation, February 2004, accessed: 2026-05-14. [Online]. Available: \url{https://www.w3.org/TR/2004/REC-rdf-primer-20040210/}
\BIBentrySTDinterwordspacing
\bibitem{W3C_OWL2_2012}
\BIBentryALTinterwordspacing
{W3C OWL Working Group}, ``{OWL} 2 web ontology language document overview (second edition),'' W3C Recommendation, 11 December 2012, accessed: 2026-05-14. [Online]. Available: \url{https://www.w3.org/TR/2012/REC-owl2-overview-20121211/}
\BIBentrySTDinterwordspacing
\bibitem{Huang_2000}
Z.~Huang, A.~Eliëns, A.~van Ballegooij, and P.~de~Bra, ``A taxonomy of web agents,'' in \emph{Proceedings of the 11th International Workshop on Database and Expert Systems Applications ({DEXA}'00)}, A.~M. Toja, Ed.\hskip 1em plus 0.5em minus 0.4em\relax Los Alamitos, CA, USA: IEEE Computer Society, September 2000, pp. 765--769.
\bibitem{Shadbolt_BernersLee_Hall_2006}
N.~Shadbolt, T.~Berners-Lee, and W.~Hall, ``The semantic web revisited,'' \emph{IEEE Intelligent Systems}, vol.~21, no.~3, pp. 96--101, 2006.
\bibitem{Heindel_Weber_2020}
T.~Heindel and I.~Weber, ``Incentive alignment of business processes,'' in \emph{Business Process Management - 18th International Conference, BPM 2020, Seville, Spain, September 13-18, 2020, Proceedings}, ser. Lecture Notes in Computer Science, vol. 12168.\hskip 1em plus 0.5em minus 0.4em\relax Cham: Springer, 2020, pp. 114--131.
\bibitem{Simperl_Cuel_Stein_2013}
E.~Simperl, R.~Cuel, and M.~Stein, \emph{Incentive-Centric Semantic Web Application Engineering}, ser. Synthesis Lectures on the Semantic Web: Theory and Technology.\hskip 1em plus 0.5em minus 0.4em\relax San Rafael, CA: Morgan \& Claypool Publishers, 2013.
\bibitem{INSEMTIVE_2008}
\BIBentryALTinterwordspacing
K.~Siorpaes and E.~Simperl, Eds., \emph{Proceedings of the 1st Workshop on Incentives for the Semantic Web (INSEMTIVE 2008)} at ISWC 2008, Karlsruhe, Germany, October 26, 2008, accessed: 2026-05-14. [Online]. Available: \url{http://km.aifb.kit.edu/ws/insemtive2008/}
\BIBentrySTDinterwordspacing
%//------------------------------------------------------------
\bibitem{Guha_2016}
R.~V. Guha, D.~Brickley, and S.~Macbeth, ``{Schema.org}: Evolution of structured data on the web,'' \emph{Communications of the ACM}, vol.~59, no.~2, pp. 44--51, 2016.
%//------------------------------------------------------------
\bibitem{Facebook_2010}
\BIBentryALTinterwordspacing
{Facebook, Inc.}, ``The open graph protocol,'' 2010, accessed: 2026-05-14. [Online]. Available: \url{https://ogp.me/}
\BIBentrySTDinterwordspacing
\bibitem{bizer_2009}
\BIBentryALTinterwordspacing
C.~Bizer, J.~Lehmann, G.~Kobilarov, S.~Auer, C.~Becker, R.~Cyganiak, and S.~Hellmann, ``{DBpedia} -- A crystallization point for the web of data,'' \emph{Journal of Web Semantics}, vol.~7, no.~3, pp. 154--165, 2009.
\BIBentrySTDinterwordspacing
\bibitem{vrandecic_2014}
D.~Vrande\v{c}i\'{c} and M.~Kr{\"o}tzsch, ``Wikidata: A free collaborative knowledge base,'' \emph{Communications of the ACM}, vol.~57, no.~10, pp. 78--85, 2014.
\bibitem{Iliadis_2023}
A.~Iliadis, A.~Acker, W.~Stevens, and S.~B.~Kavakli, ``One schema to rule them all: How {Schema.org} models the world of search,'' \emph{Journal of the Association for Information Science and Technology}, vol.~76, no.~2, pp. 460--523, 2025.
\bibitem{Yao_2005}
Y.~Yao, ``{Web intelligence: New frontiers of exploration},'' in \emph{Proceedings of the 2005 International Conference on Active Media Technology (AMT 2005)}.\hskip 1em plus 0.5em minus 0.4em\relax IEEE, 2005, pp. 3--8.
%//------------------------------------------------------------
\bibitem{WI-IAT_2024}
\BIBentryALTinterwordspacing
{IEEE/WIC}, ``The 23rd {IEEE}/{WIC} international conference on web intelligence and intelligent agent technology ({WI-IAT}'24),'' in \emph{Proceedings of the 2024 {IEEE}/{WIC} International Conference on Web Intelligence and Intelligent Agent Technology ({WI-IAT})}.\hskip 1em plus 0.5em minus 0.4em\relax Bangkok, Thailand: IEEE, 2024, accessed: 2026-05-14. [Online]. Available: \url{https://dblp.org/db/conf/webi/webi2024.html}
\BIBentrySTDinterwordspacing
\bibitem{Domingos_2024}
D.~A. Domingos, B.~M. Faria, A.~M. Z-Flores, S.~Isotani, I.~I. Bittencourt, and J.~Cascalho, ``Web intelligence journal in perspective: An analysis of its two decades trajectory,'' \emph{arXiv preprint arXiv:2405.05129}, 2024, accessed: 2026-05-14. [Online]. Available: \url{https://arxiv.org/abs/2405.05129}
\bibitem{Moya_2007}
L.~J. Moya, ``{Towards a taxonomy of agents and multi-agent systems},'' in \emph{Proceedings of the 2007 Spring Simulation Multiconference}, San Diego, CA, USA: Society for Computer Simulation International, 2007, pp. 27--34.
\bibitem{Yao_2023}
S.~Yao, J.~Zhao, D.~Yu, N.~Du, I.~Shafran, K.~Narasimhan, and Y.~Cao, ``{ReAct}: Synergizing reasoning and acting in language models,'' in \emph{Proc. International Conference on Learning Representations (ICLR)}, 2023. [Online]. Available: \url{https://openreview.net/forum?id=WE_vluYUL-X}
\bibitem{alejandre_2018hatp}
\BIBentryALTinterwordspacing
R.~Lallement, L.~de~Silva, and R.~Alami, ``{HATP}: Hierarchical agent-based task planner,'' in \emph{Proc. 17th International Conference on Autonomous Agents and Multiagent Systems (AAMAS)}, Stockholm, Sweden, 2018, pp.~1823--1825, accessed: 2026-05-14. [Online]. Available: \url{http://www.ifaamas.org/Proceedings/aamas2018/pdfs/p1823.pdf}
\BIBentrySTDinterwordspacing
\bibitem{bai_2024twostep}
D.~Bai, I.~Singh, D.~Traum, and J.~Thomason, ``{TwoStep: Multi-agent task planning using classical planners and large language models},'' \emph{arXiv preprint arXiv:2403.17246}, 2024. [Online]. Available: \url{https://arxiv.org/abs/2403.17246}
\bibitem{bauters_2016probabilistic}
\BIBentryALTinterwordspacing
K.~Bauters, K.~McAreavey, J.~Hong, Y.~Chen, W.~Liu, L.~Godo, and C.~Sierra, ``Probabilistic planning in agentspeak using the pomdp framework,'' in \emph{Combinations of Intelligent Methods and Applications}, ser. Smart Innovation, Systems and Technologies, I.~Hatzilygeroudis, V.~Palade, and J.~Prentzas, Eds., Cham: Springer International Publishing, 2016, vol.~46, pp. 19--37.
\BIBentrySTDinterwordspacing
\bibitem{yao_2023treeofthoughts}
S.~Yao, D.~Yu, J.~Zhao, I.~Shafran, T.~L. Griffiths, Y.~Cao, and K.~Narasimhan, ``{Tree of thoughts}: Deliberate problem solving with large language models,'' in \emph{Advances in Neural Information Processing Systems (NeurIPS)}, vol.~36, 2023. [Online]. Available: \url{https://proceedings.neurips.cc/paper_files/paper/2023/hash/271db9922b8d1f4dd7aaef84ed5ac703-Abstract-Conference.html}
%///---------------------------------------------------------------------------------------------
\bibitem{LangChainRepo}
{langchain-ai}, ``{{LangChain}: A framework for developing applications powered by large language models},'' 2023, accessed: 2026-05-14. [Online]. Available: \url{https://github.com/langchain-ai/langchain}
\bibitem{MetaGPT_2024}
S.~Hong, M.~Zhuge, J.~Chen, X.~Zheng, Y.~Cheng, C.~Zhang, J.~Wang, Z.~Wang, S.~K.~S. Yau, Z.~Lin, L.~Zhou, C.~Ran, L.~Xiao, C.~Wu, and J.~Schmidhuber, ``{MetaGPT}: Meta programming for a multi-agent collaborative framework,'' in \emph{Proc. International Conference on Learning Representations (ICLR), Oral}, 2024. [Online]. Available: \url{https://openreview.net/forum?id=VtmBAGCN7o}
\bibitem{CrewAIRepo}
{CrewAI Contributors}, ``Crewai: Framework for orchestrating role-playing, autonomous {AI} agents,'' GitHub repository \url{https://github.com/crewAIInc/crewAI}, 2023, accessed: 2026-05-14.
\bibitem{nlweb}
{Microsoft}, ``{NLWeb}: Bringing conversational interfaces directly to the web,'' \url{https://github.com/nlweb-ai/NLWeb}, 2025, accessed: 2026-05-14.
\bibitem{Microsoft_MagenticUI_2025}
{Microsoft}, ``{Magentic-UI}: A research prototype of a human-centered interface powered by a multi-agent system,'' \url{https://github.com/microsoft/magentic-ui}, 2025, accessed: 2026-05-14.
\bibitem{GraphRAG_2024}
\BIBentryALTinterwordspacing
D.~Edge, H.~Trinh, N.~Cheng, J.~Bradley, A.~Chao, A.~Mody, S.~Truitt, D.~Metropolitansky, R.~O.~Ness, and J.~Larson, ``{From Local to Global: A GraphRAG Approach to Query-Focused Summarization},'' Microsoft Research technical report, 2024, also available on arXiv. [Online]. Available: \url{https://arxiv.org/abs/2404.16130}
\BIBentrySTDinterwordspacing
\bibitem{AnthropicComputer2024}
{Anthropic}, ``Build with {Claude},'' \emph{Anthropic Documentation}, 2024, accessed: 2026-05-14. [Online]. Available: \url{https://docs.anthropic.com/en/docs/using-the-computer}
\bibitem{OpenAI_Operator_2025}
\BIBentryALTinterwordspacing
{OpenAI}, ``{Introducing Operator},'' OpenAI Blog, January~23, 2025, accessed: 2026-05-14. [Online]. Available: \url{https://openai.com/index/introducing-operator/}
\BIBentrySTDinterwordspacing
\bibitem{SWE_Agent_2024}
\BIBentryALTinterwordspacing
J.~Yang, C.~E. Jimenez, A.~Wettig, K.~Lieret, S.~Yao, K.~Narasimhan, and O.~Press, ``{SWE-agent: Agent-Computer Interfaces Enable Automated Software Engineering},'' in \emph{Advances in Neural Information Processing Systems (NeurIPS)}, 2024. [Online]. Available: \url{https://arxiv.org/abs/2405.15793}
\BIBentrySTDinterwordspacing
\bibitem{Cognition_Devin_2024}
\BIBentryALTinterwordspacing
{Cognition AI}, ``{Introducing Devin, the first AI software engineer},'' Cognition AI Blog, March~12, 2024, accessed: 2026-05-14. [Online]. Available: \url{https://cognition.ai/blog/introducing-devin}
\BIBentrySTDinterwordspacing
\bibitem{Brown_2020}
\BIBentryALTinterwordspacing
T.~B. Brown, B.~Mann, N.~Ryder, M.~Subbiah, J.~Kaplan, P.~Dhariwal, A.~Neelakantan, P.~Shyam, G.~Sastry, A.~Askell, S.~Agarwal, A.~Herbert-Voss, G.~Krueger, T.~Henighan, R.~Child, A.~Ramesh, D.~M. Ziegler, J.~Wu, C.~Winter, C.~Hesse, M.~Chen, E.~Sigler, M.~Litwin, S.~Gray, B.~Chess, J.~Clark, C.~Berner, S.~McCandlish, A.~Radford, I.~Sutskever, and D.~Amodei, ``Language models are few-shot learners,'' in \emph{Advances in Neural Information Processing Systems (NeurIPS)}, vol.~33, 2020, pp.~1877--1901. [Online]. Available: \url{https://proceedings.neurips.cc/paper/2020/hash/1457c0d6bfcb4967418bfb8ac142f64a-Abstract.html}
\BIBentrySTDinterwordspacing
\bibitem{Bubeck_2023}
S.~Bubeck, V.~Chandrasekaran, R.~Eldan, J.~Gehrke, E.~Horvitz, E.~Kamar, P.~Lee, Y.~T. Lee, Y.~Li, S.~Lundberg, H.~Nori, H.~Palangi, M.~T. Ribeiro, and Y.~Zhang, ``Sparks of artificial general intelligence: Early experiments with {GPT-4},'' \emph{arXiv preprint arXiv:2303.12712}, 2023. [Online]. Available: \url{https://arxiv.org/abs/2303.12712}
\bibitem{Google_A2A_Repo}
{Google}, ``{Agent2Agent} ({A2A}) protocol repository,'' \url{https://github.com/a2aproject/A2A}, 2025, accessed: 2026-05-14.
\bibitem{OpenAI_Plugins_2023}
\BIBentryALTinterwordspacing
{OpenAI}, ``Introducing {ChatGPT} plugins,'' OpenAI Blog, March 2023, accessed: 2026-05-14. [Online]. Available: \url{https://openai.com/index/chatgpt-plugins/}
\BIBentrySTDinterwordspacing
\bibitem{Lewis_2020}
\BIBentryALTinterwordspacing
P.~Lewis, E.~P{\'{e}}rez, A.~Piktus, F.~Petroni, V.~Karpukhin, N.~Goyal, H.~K{\"{u}}ttler, S.~Welleck, M.~Komeili, W.~Yih, T.~Rockt{\"{a}}schel, S.~Riedel, and D.~Kiela, ``{Retrieval-augmented generation for knowledge-intensive NLP tasks},'' in \emph{Advances in Neural Information Processing Systems}, H.~Larochelle, M.~Ranzato, R.~Hadsell, M.~F. Balcan, and H.~Lin, Eds., vol.~33, 2020, pp. 9459--9474, accessed: 2026-05-14. [Online]. Available: \url{https://arxiv.org/abs/2005.11401}
\BIBentrySTDinterwordspacing
\bibitem{k8sborgblog}
B.~Burns, ``Borg, the predecessor to {K}ubernetes,'' \emph{Kubernetes Blog}, Apr. 2015, accessed: 2026-05-14. [Online]. Available: \url{https://kubernetes.io/blog/2015/04/borg-predecessor-to-kubernetes/}
\bibitem{gkeai}
{Google Cloud}, ``{AI} and {ML} orchestration on {GKE},'' \emph{Google Kubernetes Engine Documentation}, accessed: 2026-05-14. [Online]. Available: \url{https://cloud.google.com/kubernetes-engine/docs/integrations/ai-infra}
\bibitem{collabnix}
A.~Gupta, ``Agentic {AI} on {K}ubernetes: Advanced orchestration, deployment, and scaling strategies for autonomous {AI} systems,'' \emph{Collabnix}, May 2024, accessed: 2026-05-14. [Online]. Available: \url{https://collabnix.com/agentic-ai-on-kubernetes-advanced-orchestration-deployment-and-scaling-strategies-for-autonomous-ai-systems/}
\bibitem{Schick_2023}
T.~Schick, J.~Dwivedi-Yu, R.~Dess{\`{i}}, R.~Raileanu, M.~Lomeli, E.~Hambro, L.~Zettlemoyer, N.~Cancedda, and T.~Scialom, ``{Toolformer}: Language models can teach themselves to use tools,'' in \emph{Advances in Neural Information Processing Systems}, vol.~36, 2023, pp. 68\,539--68\,551. [Online]. Available: \url{https://arxiv.org/abs/2302.04761}
\bibitem{Christiano_2017}
\BIBentryALTinterwordspacing
P.~F. Christiano, J.~Leike, T.~B. Brown, M.~Martic, S.~Legg, and D.~Amodei, ``{Deep reinforcement learning from human preferences},'' in \emph{Advances in Neural Information Processing Systems}, I.~Guyon, U.~V. Luxburg, S.~Bengio, H.~Wallach, R.~Fergus, S.~Vishwanathan, and R.~Garnett, Eds., vol.~30, 2017, pp. 4299--4307, accessed: 2026-05-14. [Online]. Available: \url{https://proceedings.neurips.cc/paper/2017/hash/d5e2c0adad503c91f91df240d0cd4e49-Abstract.html}
\BIBentrySTDinterwordspacing
\bibitem{Ross_2011}
\BIBentryALTinterwordspacing
S.~Ross, G.~J. Gordon, and J.~A. Bagnell, ``{A reduction of imitation learning and structured prediction to no-regret online learning},'' in \emph{Proceedings of the Fourteenth International Conference on Artificial Intelligence and Statistics}, ser. Proceedings of Machine Learning Research, Z.~Ghahramani, M.~Welling, C.~Cortes, N.~D. Lawrence, and K.~Q. Weinberger, Eds., vol.~15.\hskip 1em plus 0.5em minus 0.4em\relax PMLR, 2011, pp. 627--635, accessed: 2026-05-14. [Online]. Available: \url{https://proceedings.mlr.press/v15/ross11a.html}
\BIBentrySTDinterwordspacing
\bibitem{Docker_2025}
\BIBentryALTinterwordspacing
{Docker Inc.}, ``The model context protocol: An emerging standard for {AI} agent-tool interactions,'' White Paper, Docker Inc., 2025, accessed: 2026-05-14. [Online]. Available: \url{https://www.docker.com/resources/the-model-context-protocol-white-paper/}
\BIBentrySTDinterwordspacing
\bibitem{MCP_Tasks_2025}
\BIBentryALTinterwordspacing
{Model Context Protocol Working Group}, ``{Model Context Protocol Specification, version 2025-11-25}: Tasks, OAuth 2.1, and Registry,'' November~25, 2025, accessed: 2026-05-14. [Online]. Available: \url{https://modelcontextprotocol.io/specification/2025-11-25}
\BIBentrySTDinterwordspacing
\bibitem{WebMCP_W3C_2025}
\BIBentryALTinterwordspacing
{W3C Web Machine Learning Community Group}, ``{WebMCP: Browser-Native Extension of the Model Context Protocol},'' W3C Community Group draft (deliverable accepted September~2025), accessed: 2026-05-14. [Online]. Available: \url{https://webmachinelearning.github.io/webmcp/}
\BIBentrySTDinterwordspacing
\bibitem{AGUI_2025}
\BIBentryALTinterwordspacing
{ag-ui-protocol contributors}, ``{AG-UI: The Agent-User Interaction Protocol},'' GitHub repository, May~2025 (initial release), accessed: 2026-05-14. [Online]. Available: \url{https://github.com/ag-ui-protocol/ag-ui}
\BIBentrySTDinterwordspacing
\bibitem{A2A_LinuxFoundation_2025}
\BIBentryALTinterwordspacing
{Linux Foundation}, ``{Linux Foundation Launches the Agent2Agent Protocol Project to Enable Secure, Intelligent Communication Between AI Agents},'' Press release, June~23, 2025, accessed: 2026-05-14. [Online]. Available: \url{https://www.linuxfoundation.org/press/linux-foundation-launches-the-agent2agent-protocol-project-to-enable-secure-intelligent-communication-between-ai-agents}
\BIBentrySTDinterwordspacing
\bibitem{ACP_Merger_2025}
\BIBentryALTinterwordspacing
K.~Blair and T.~Segal, ``{ACP Joins Forces with A2A under the Linux Foundation's LF AI \& Data},'' Linux Foundation AI \& Data Community Blog, August~29, 2025, accessed: 2026-05-14. [Online]. Available: \url{https://lfaidata.foundation/communityblog/2025/08/29/acp-joins-forces-with-a2a-under-the-linux-foundations-lf-ai-data/}
\BIBentrySTDinterwordspacing
\bibitem{AAIF_2025}
\BIBentryALTinterwordspacing
{Linux Foundation}, ``{Linux Foundation Announces the Formation of the Agentic AI Foundation},'' Press release, December~9, 2025, accessed: 2026-05-14. [Online]. Available: \url{https://www.linuxfoundation.org/press/linux-foundation-announces-the-formation-of-the-agentic-ai-foundation}
\BIBentrySTDinterwordspacing
\bibitem{A2A_v03_2025}
\BIBentryALTinterwordspacing
{A2A Working Group}, ``{Agent-to-Agent (A2A) Protocol Specification v0.3.0},'' GitHub release, July~30, 2025, accessed: 2026-05-14. [Online]. Available: \url{https://github.com/a2aproject/A2A/releases/tag/v0.3.0}
\BIBentrySTDinterwordspacing
\bibitem{A2A_v10_2026}
\BIBentryALTinterwordspacing
{A2A Working Group}, ``{Announcing A2A Protocol v1.0.0},'' A2A Project Documentation, March~12, 2026, accessed: 2026-05-14. [Online]. Available: \url{https://a2a-protocol.org/latest/announcing-1.0/}
\BIBentrySTDinterwordspacing
\bibitem{AutoGenRepo}
{Microsoft Research}, ``Autogen: Enabling next-gen llm applications via multi-agent conversation,'' GitHub repository \url{https://github.com/microsoft/autogen}, 2023, accessed: 2026-05-14.
\bibitem{LangChain_2023}
{LangChain Authors}, ``{LangChain}: A framework for building {LLM} applications with tooling,'' \url{https://docs.langchain.com/oss/python/langchain/overview}, 2023, accessed: 2026-05-14.
\bibitem{CrewAI_2024}
{CrewAI Contributors}, ``{{CrewAI}}: A multi-agent orchestration toolkit,'' 2024, accessed: 2026-05-14. [Online]. Available: \url{https://github.com/crewAIInc/crewAI}
\bibitem{AWSStrands_2025}
\BIBentryALTinterwordspacing
C.~Liguori, ``Introducing {S}trands {A}gents, an open source {AI} agents {SDK},'' \emph{AWS Open Source Blog}, May 16, 2025, accessed: 2026-05-14. [Online]. Available: \url{https://aws.amazon.com/blogs/opensource/introducing-strands-agents-an-open-source-ai-agents-sdk/}
\BIBentrySTDinterwordspacing
\bibitem{SemanticKernel_2023}
{Microsoft}, ``{Semantic Kernel}: Orchestration and plugins for {AI},'' \url{https://github.com/microsoft/semantic-kernel}, 2023, accessed: 2026-05-14.
\bibitem{SemanticKernelA2A_2024}
E.~Mattson, ``Integrating {S}emantic {K}ernel {P}ython with {G}oogle's {A2A} protocol,'' \emph{Microsoft Foundry Blog}, Apr.~17, 2025, accessed: 2026-05-14. [Online]. Available: \url{https://devblogs.microsoft.com/foundry/semantic-kernel-a2a-integration/}
\bibitem{LlamaIndexSite_2025}
{LlamaIndex Development Team}, ``{LlamaIndex}: Build knowledge assistants over your enterprise data,'' \url{https://www.llamaindex.ai/}, accessed: 2026-05-14.
\bibitem{SuperAGIRepo}
{TransformerOptimus}, ``{SuperAGI}: A dev-first open-source autonomous {AI} agent framework,'' \url{https://github.com/TransformerOptimus/SuperAGI}, 2023, accessed: 2026-05-14.
\bibitem{AmazonBedrock2023}
{Amazon Web Services}, ``Amazon bedrock agents,'' \url{https://aws.amazon.com/bedrock/agents/}, 2023, accessed: 2026-05-14.
\bibitem{AzureAIAgent2024}
{Microsoft Azure}, ``Azure {AI} Studio,'' \emph{Microsoft Azure}, 2024, accessed: 2026-05-14. [Online]. Available: \url{https://azure.microsoft.com/en-us/products/ai-studio}
\bibitem{GoogleAgentBuilder}
\BIBentryALTinterwordspacing
{Google Cloud}, ``{Vertex AI Agent Builder} overview,'' 2024, accessed: 2026-05-14. [Online]. Available: \url{https://cloud.google.com/vertex-ai/generative-ai/docs/agent-builder/overview}
\BIBentrySTDinterwordspacing

%/-------------------------------------------------
\bibitem{OpenAIAssistants2023}
{OpenAI}, ``Assistants {API},'' \emph{OpenAI API Documentation}, 2023, accessed: 2026-05-14. [Online]. Available: \url{https://platform.openai.com/docs/assistants/overview}
\bibitem{IBMWatson2023}
{IBM}, ``{IBM} watsonx Orchestrate,'' \emph{IBM}, 2023, accessed: 2026-05-14. [Online]. Available: \url{https://www.ibm.com/products/watsonx-orchestrate}

%/--------------------------------------------------------------------
\bibitem{AWSBedrockMarketplace}
\BIBentryALTinterwordspacing
{Amazon Web Services}, ``{AI} foundation model marketplace - {Amazon Bedrock Marketplace},'' 2024, accessed: 2026-05-14. [Online]. Available: \url{https://aws.amazon.com/bedrock/marketplace/}
\BIBentrySTDinterwordspacing
\bibitem{OpenAIGPTStore}
\BIBentryALTinterwordspacing
{OpenAI}, ``Introducing the {GPT Store},'' January 2024, accessed: 2026-05-14. [Online]. Available: \url{https://openai.com/index/introducing-the-gpt-store/}
\BIBentrySTDinterwordspacing
\bibitem{SWE_Bench_2024}
\BIBentryALTinterwordspacing
C.~E. Jimenez, J.~Yang, A.~Wettig, S.~Yao, K.~Pei, O.~Press, and K.~Narasimhan, ``{SWE-bench: Can Language Models Resolve Real-World GitHub Issues?},'' in \emph{Proc. International Conference on Learning Representations (ICLR)}, 2024. [Online]. Available: \url{https://openreview.net/forum?id=VTF8yNQM66}
\BIBentrySTDinterwordspacing
\bibitem{Asai_2024_SelfRAG}
A.~Asai, Z.~Wu, Y.~Wang, A.~Sil, and H.~Hajishirzi, ``{Self-RAG}: Learning to retrieve, generate, and critique through self-reflection,'' in \emph{Proc. International Conference on Learning Representations (ICLR)}, 2024.
\bibitem{Petrova_CARRAG_2025}
T.~Petrova, D.~Koriakov, and R.~State, ``{CAR-RAG}: Category-aware hybrid retrieval-augmented generation for hallucination mitigation,'' in \emph{Proc.\ {IEEE} Int.\ Conf.\ Big Data ({BigData})}, 2025, pp.\ 1547--1554.
\bibitem{Rehm_2020_Interoperability}
\BIBentryALTinterwordspacing
G.~Rehm, D.~Galanis, P.~Labropoulou, S.~Piperidis, M.~Welß, R.~Usbeck, J.~Köhler, M.~Deligiannis, K.~Gkirtzou, J.~Fischer, C.~Chiarcos, N.~Feldhus, J.~Moreno-Schneider, F.~Kintzel, E.~Montiel, V.~Rodríguez~Doncel, J.~P. McCrae, D.~Laqua, I.~P. Theile, C.~Dittmar, K.~Bontcheva, I.~Roberts, A.~Vasiljevs, and A.~Lagzdiņš, ``Towards an interoperable ecosystem of {AI} and {LT} platforms: A roadmap for the implementation of different levels of interoperability,'' in \emph{Proc. 1st International Workshop on Language Technology Platforms (IWLTP)} (co-located with LREC~2020), Marseille, France: European Language Resources Association, 2020, pp.~96--107. [Online]. Available: \url{https://aclanthology.org/2020.iwltp-1.15/}
\BIBentrySTDinterwordspacing
\bibitem{Yekollu_2024_AgentMarketplace}
N.~Yekollu, R.~Jain, and S.~G. Patil, ``Agent marketplace,'' 2024. [Online]. Available: \url{https://gorilla.cs.berkeley.edu/blogs/11_agent_marketplace.html}
\bibitem{ETHOS_2024}
T.~J.~Chaffer, J.~Goldston, B.~Okusanya, and Gemach~D.A.T.A.~I, ``On the ETHOS of AI Agents: An Ethical Technology and Holistic Oversight System,'' arXiv preprint arXiv:2412.17114, 2024.
\bibitem{W3C_VC_2025}
\BIBentryALTinterwordspacing
{World Wide Web Consortium}, ``{Verifiable Credentials Data Model v2.0},'' World Wide Web Consortium, W3C Recommendation, May 2025, accessed: 2026-05-14. [Online]. Available: \url{https://www.w3.org/TR/vc-data-model-2.0/}
\BIBentrySTDinterwordspacing
\bibitem{KYA_Paradigm_2025}
\BIBentryALTinterwordspacing
T.~J.~Chaffer, ``{Know Your Agent: Governing AI Identity on the Agentic Web},'' SSRN Working Paper, February~2025, doi: 10.2139/ssrn.5162127. [Online]. Available: \url{https://doi.org/10.2139/ssrn.5162127}
\BIBentrySTDinterwordspacing

% =====================================================================
% Added in v4 second pass: computer-use, software-engineering agents,
% GraphRAG, OpenAI Function Calling
% =====================================================================
\bibitem{Huang_2025}
K.~Huang, V.~S. Narajala, J.~Yeoh, J.~Ross, R.~Raskar, Y.~Harkati, J.~Huang, I.~Habler, and C.~Hughes, ``{A Novel Zero-Trust Identity Framework for Agentic AI: Decentralized Authentication and Fine-Grained Access Control},'' \emph{arXiv preprint arXiv:2505.19301}, May 2025. [Online]. Available: \url{https://arxiv.org/abs/2505.19301}
\bibitem{Clemm_2022}
A.~Clemm, L.~Ciavaglia, L.~Z. Granville, and J.~Tantsura, ``Intent-based networking - concepts and definitions,'' Internet Research Task Force (IRTF), RFC 9315, October 2022. [Online]. Available: \url{https://www.ietf.org/rfc/rfc9315.html}
\bibitem{Liu_2024_PQCSurvey}
T.~Liu, G.~S. Ramachandran, and R.~Jurdak, ``Post-quantum cryptography for internet of things: A survey on performance and optimization,'' \emph{arXiv preprint arXiv:2401.17538}, 2024. [Online]. Available: \url{https://arxiv.org/html/2401.17538v1/}
\bibitem{Noothigattu_2018}
R.~Noothigattu, S.~Gaikwad, E.~Awad, S.~Dsouza, I.~Rahwan, P.~Ravikumar, and A.~D. Procaccia, ``A voting-based system for ethical decision making,'' in \emph{Proceedings of the Thirty-Second AAAI Conference on Artificial Intelligence}, 2018, pp.~1587--1594. [Online]. Available: \url{https://ojs.aaai.org/index.php/AAAI/article/view/11512}


%/---------------------------------------------------------
\bibitem{Anthropic_2025}
\BIBentryALTinterwordspacing
Anthropic, ``How we built our multi-agent research system,'' Anthropic Engineering Blog, June 2025, accessed: 2026-05-14. [Online]. Available: \url{https://www.anthropic.com/engineering/multi-agent-research-system}
\BIBentrySTDinterwordspacing
\bibitem{Sadykhov_2023_DeTEcT}
\BIBentryALTinterwordspacing
R.~Sadykhov, G.~Goodell, D.~de~Montigny, M.~Schoernig, and P.~Treleaven, ``Decentralized token economy theory ({DeTEcT}): Token pricing, stability and governance for token economies,'' \emph{Frontiers in Blockchain}, vol.~6, art.~1298330, 2023, doi: 10.3389/fbloc.2023.1298330. [Online]. Available: \url{https://doi.org/10.3389/fbloc.2023.1298330}
\BIBentrySTDinterwordspacing
\bibitem{Yermack_2015}
D.~Yermack, ``Is bitcoin a real currency? An economic appraisal,'' in \emph{Handbook of Digital Currency}, D.~K.~C.~Lee, Ed. Amsterdam: Elsevier, 2015, pp. 31--43.
\bibitem{Visa_TAP_2025}
\BIBentryALTinterwordspacing
{Visa Inc.}, ``{Visa Introduces Trusted Agent Protocol: An Ecosystem-Led Framework for AI Commerce},'' Press release, October~14, 2025, accessed: 2026-05-14. [Online]. Available: \url{https://usa.visa.com/about-visa/newsroom/press-releases/visa-intelligent-commerce.html}
\BIBentrySTDinterwordspacing
\bibitem{Mastercard_AgentPay_2025}
\BIBentryALTinterwordspacing
{Mastercard}, ``{Mastercard Unveils Agent Pay, Pioneering Agentic Payments Technology to Power Commerce in the Age of AI},'' Press release, April~29, 2025, accessed: 2026-05-14. [Online]. Available: \url{https://www.mastercard.com/global/en/news-and-trends/press/2025/april/mastercard-unveils-agent-pay-pioneering-agentic-payments-technology-to-power-commerce-in-the-age-of-ai.html}
\BIBentrySTDinterwordspacing
\bibitem{Stripe_OpenAI_ACP_2025}
\BIBentryALTinterwordspacing
{Stripe} and {OpenAI}, ``{Developing an Open Standard for Agentic Commerce},'' Stripe Blog, September~29, 2025, accessed: 2026-05-14. [Online]. Available: \url{https://stripe.com/blog/developing-an-open-standard-for-agentic-commerce}
\BIBentrySTDinterwordspacing
\bibitem{Akerlof_1970}
G.~A. Akerlof, ``The market for `lemons': Quality uncertainty and the market mechanism,'' \emph{The Quarterly Journal of Economics}, vol.~84, no.~3, pp. 488--500, 1970.
\bibitem{Sabater_2005}
J.~Sabater and C.~Sierra, ``Review on computational trust and reputation models,'' \emph{Artificial Intelligence Review}, vol.~24, no.~1, pp. 33--60, 2005. [Online]. Available: \url{https://doi.org/10.1007/s10462-004-0041-5}
%/---------------------------------------------------------
\bibitem{Josang_2007}
A.~J{\o}sang, R.~Ismail, and C.~Boyd, ``A survey of trust and reputation systems for online service provision,'' \emph{Decision Support Systems}, vol.~43, no.~2, pp. 618--644, 2007. [Online]. Available: \url{https://doi.org/10.1016/j.dss.2005.05.019}
\bibitem{Huynh_2006}
T.~D. Huynh, N.~R. Jennings, and N.~R. Shadbolt, ``An integrated trust and reputation model for open multi-agent systems,'' \emph{Autonomous Agents and Multi-Agent Systems}, vol.~13, no.~2, pp. 119--154, 2006. [Online]. Available: \url{https://doi.org/10.1007/s10458-005-6825-4}
\bibitem{Kamvar_2003}
S.~D. Kamvar, M.~T. Schlosser, and H.~Garcia-Molina, ``The EigenTrust algorithm for reputation management in P2P networks,'' in \emph{Proceedings of the 12th International Conference on World Wide Web (WWW '03)}.\hskip 1em plus 0.5em minus 0.4em\relax New York, NY, USA: ACM, 2003, pp. 640--651. [Online]. Available: \url{https://doi.org/10.1145/775152.775242}
\bibitem{Friedman_2001} E.~Friedman and P.~Resnick, ``The social cost of cheap pseudonyms,'' \emph{Journal of Economics \& Management Strategy}, vol.~10, no.~2, pp. 173--199, 2001. [Online]. Available: \url{https://smg.media.mit.edu/library/FriedmanResnick.pseudonyms.pdf} (accessed: 2026-05-14).
\bibitem{Dogan_2023}
{\"O}.~Do{\u{g}}an and H.~Karacan, ``A blockchain-based e-commerce reputation system built with verifiable credentials,'' \emph{IEEE Access}, vol.~11, pp. 49227--49238, 2023. [Online]. Available: \url{https://doi.org/10.1109/ACCESS.2023.3274707}
\bibitem{Jurca_2005}
R.~Jurca and B.~Faltings, ``Truthful feedback for sanctioning reputation mechanisms,'' in \emph{Proceedings of the 6th ACM Conference on Electronic Commerce (EC '05)}.\hskip 1em plus 0.5em minus 0.4em\relax New York, NY, USA: ACM, 2005, pp. 190--199.
%/---------------------------------------------------------
\bibitem{Dimitriou_2021} A.~Battah, Y.~Iraqi, and E.~Damiani, ``Blockchain-based reputation systems: Implementation challenges and mitigation,'' \emph{Electronics}, vol.~10, no.~3, p. 289, 2021. [Online]. Available: \url{https://www.mdpi.com/2079-9292/10/3/289} (accessed: 2026-05-14).
\bibitem{Masse_2016} M.~Masse, \emph{REST API Design Rulebook: Designing Consistent Web Services}.\hskip 1em plus 0.5em minus 0.4em\relax O'Reilly Media, 2016.
\bibitem{Heshmatisafa_2023}
S.~Heshmatisafa and M.~Seppänen, ``Exploring API-driven business models: Lessons learned from Amadeus's digital transformation,'' \emph{Digital Business}, vol.~3, no.~1, p.~100055, 2023. [Online]. Available: \url{https://doi.org/10.1016/j.digbus.2023.100055}
\bibitem{Erradi_2022_Cycle}
Z.~Hong, S.~Guo, R.~Zhang, P.~Li, Y.~Zhan, and W.~Chen, ``Cycle: Sustainable Off-Chain Payment Channel Network with Asynchronous Rebalancing,'' in \emph{2022 52nd Annual IEEE/IFIP International Conference on Dependable Systems and Networks (DSN)}, 2022, pp.~41--53. [Online]. Available: \url{https://ieeexplore.ieee.org/document/9833795}

% =====================================================================
% Added in v4 (2026-05-14): 2025--2026 institutional convergence references
% =====================================================================
\bibitem{AISafetyReport_2026}
\BIBentryALTinterwordspacing
Y.~Bengio et al., ``{International AI Safety Report 2026},'' International AI Safety Report Secretariat, February~2026, accessed: 2026-05-14. [Online]. Available: \url{https://internationalaisafetyreport.org/publication/international-ai-safety-report-2026}
\BIBentrySTDinterwordspacing
\bibitem{Liu_2023} Y.~Liu, G.~Deng, Y.~Li, K.~Wang, T.~Zhang, Y.~Liu, H.~Wang, Y.~Zheng, and Y.~Liu, ``Prompt injection attack against LLM-integrated applications [HouYi framework],'' arXiv preprint arXiv:2306.05499, 2023. [Online]. Available: \url{https://arxiv.org/abs/2306.05499} (accessed: 2026-05-14).

%/---------------------------------------------------------
\bibitem{Koh_2017}
\BIBentryALTinterwordspacing
P.~W. Koh and P.~Liang, ``Understanding black-box predictions via influence functions,'' in \emph{Proceedings of the 34th International Conference on Machine Learning}, ser. Proceedings of Machine Learning Research, D.~Precup and Y.~W. Teh, Eds., vol.~70.\hskip 1em plus 0.5em minus 0.4em\relax PMLR, 2017, pp. 1885--1894, accessed: 2026-05-14. [Online]. Available: \url{http://proceedings.mlr.press/v70/koh17a/koh17a.pdf}
\BIBentrySTDinterwordspacing
\bibitem{OWASP_GenAI_2025}
\BIBentryALTinterwordspacing
{OWASP Foundation}, ``{OWASP Top 10 for Large Language Model Applications},'' 2024, version 1.1, accessed: 2026-05-14. [Online]. Available: \url{https://owasp.org/www-project-top-10-for-large-language-model-applications/}
\BIBentrySTDinterwordspacing
\bibitem{UETA_1999}
\BIBentryALTinterwordspacing
{National Conference of Commissioners on Uniform State Laws}, ``{Uniform Electronic Transactions Act},'' 1999, accessed: 2026-05-14. [Online]. Available: \url{http://euro.ecom.cmu.edu/program/law/08-732/Transactions/ueta.pdf}
\BIBentrySTDinterwordspacing
\bibitem{Goldwasser_1989}
S.~Goldwasser, S.~Micali, and C.~Rackoff, ``The knowledge complexity of interactive proof systems,'' \emph{SIAM Journal on Computing}, vol.~18, no.~1, pp. 186--208, 1989.
\bibitem{Cherif_2023}
\BIBentryALTinterwordspacing
A.~Nait~Cherif, Y.~Achir, M.~Youssfi, M.~Elgarej, and O.~Bouattane, ``{Zero-Knowledge Proofs and OAuth 2.0 for Anonymity and Security in Distributed Systems},'' in \emph{E3S Web of Conferences}, vol. 469, 2023, p. 00085. [Online]. Available: \url{https://doi.org/10.1051/e3sconf/202346900085}
\BIBentrySTDinterwordspacing
%/--------------------------------------------------------
\bibitem{Gabizon_2019}
A.~Gabizon, Z.~J.~Williamson, and O.~Ciobotaru, ``PLONK: Permutations over Lagrange-bases for Oecumenical Noninteractive Arguments of Knowledge,'' Cryptology ePrint Archive, Report 2019/953, 2019. [Online]. Available: \url{https://eprint.iacr.org/2019/953}
\bibitem{Lior_2022}
\BIBentryALTinterwordspacing
A.~Lior, ``{Insuring AI: The Role of Insurance in Artificial Intelligence Regulation},'' \emph{Harvard Journal of Law \& Technology}, vol.~35, no.~2, pp.~469--529, 2022, accessed: 2026-05-14. [Online]. Available: \url{https://jolt.law.harvard.edu/assets/articlePDFs/v35/2.-Lior-Insuring-AI.pdf}
\BIBentrySTDinterwordspacing
\bibitem{Toth_2019}
\BIBentryALTinterwordspacing
K.~C.~Toth, ``{Agent-Based Digital Identity Architecture},'' in \emph{Proceedings of the Pacific Northwest Software Quality Conference (PNSQC)}, 2019, pp.~1--12, accessed: 2026-05-14. [Online]. Available: \url{https://www.pnsqc.org/docs/Toth_Agent-BasedDigital_Final.pdf}
\BIBentrySTDinterwordspacing
\bibitem{EU_ID_2024}
\BIBentryALTinterwordspacing
{European Commission}, ``{European Digital Identity Wallet: Security and Privacy Features},'' European Commission, 2024, accessed: 2026-05-14. [Online]. Available: \url{https://ec.europa.eu/digital-building-blocks/sites/display/EUDIGITALIDENTITYWALLET/Security+and+Privacy}
\BIBentrySTDinterwordspacing
\bibitem{Pantiukhov_2024}
P.~Pantiukhov, D.~Koriakov, T.~Petrova, J.~H. Alves, V.~K. Gurbani, and R.~State, ``Enhanced {DeFi} security on {XRPL} with zero-knowledge proofs and speaker verification,'' in \emph{Proc.\ {IEEE} Int.\ Conf.\ Expo Real Time Commun.\ at {IIT}}, 2024.
\bibitem{EUAI_2024}
\BIBentryALTinterwordspacing
{European Parliament and Council of the European Union}, ``{Regulation (EU) 2024/1689 of the European Parliament and of the Council of 13 June 2024 laying down harmonised rules on artificial intelligence and amending Regulations (EC) No 300/2008, (EU) No 167/2013, (EU) No 168/2013, (EU) 2018/858, (EU) 2018/1139 and (EU) 2019/2144 and Directives 2014/90/EU, (EU) 2016/797 and (EU) 2020/1828 (Artificial Intelligence Act)},'' Official Journal of the European Union, pp.~1--98, Jul. 2024. Regulation (EU) 2024/1689; Accessed: 2026-05-14. [Online]. Available: \url{https://eur-lex.europa.eu/eli/reg/2024/1689/oj/eng}
\BIBentrySTDinterwordspacing
\bibitem{NIST_AgentStandards_2026}
\BIBentryALTinterwordspacing
{Center for AI Standards and Innovation (CAISI), U.S. National Institute of Standards and Technology}, ``{AI Agent Standards Initiative},'' Announced February~17, 2026, accessed: 2026-05-14. [Online]. Available: \url{https://www.nist.gov/caisi/ai-agent-standards-initiative}
\BIBentrySTDinterwordspacing

\end{thebibliography}

\clearpage
\appendix
\section{Comparison of Representative LLM-Based Agent Frameworks}\label{app:frameworks}

\begin{table*}[!ht]
 \caption{Comparison of representative LLM-based agent frameworks (cross-referenced from \S\ref{sec:llm_frameworks}). AutoGen, SuperAGI, and LlamaIndex follow comparable layered patterns and are not separately tabulated.}
 \label{tab:frameworks_comparison}
 \centering
 \scriptsize
 \begin{tabularx}{\textwidth}{|
  p{0.15\textwidth}
| Y
| Y
| Y
| Y
|}
  \hline
  \textbf{Characteristic}
   & \textbf{AutoGPT (2023) (Single LLM Agent)}
   & \textbf{LangChain (2022) (LLM App Framework)}
   & \textbf{CrewAI (2023) (Role-based Multi-Agent)}
   & \textbf{MetaGPT (2023) (Multi-Agent System)} \\
  \hline
  Primary focus
   & Fully autonomous goal-driven agent; demonstrate LLM autonomy in executing tasks end-to-end.
   & Developer library for building LLM-powered applications (chains, agents, chatbots, etc.) with modular components.
   & Lightweight framework for orchestrating role-playing autonomous agents with a designated orchestrator and shared task context.
   & Research framework for collaborative multi-agent workflows, simulating a team of specialized LLM agents to tackle complex projects. \\
  \hline
  Level of autonomy
   & High. Runs continuously without user intervention, deciding and executing next actions until goal completion; single-agent recursive loop (self-feedback).
   & Configurable. Not inherently autonomous; supports synchronous chains or custom agent loops defined by developer.
   & High. Each agent runs autonomously within an assigned role; the orchestrator coordinates task hand-off without per-step human input.
   & High (multi-agent). Agents autonomously exchange information and outputs; once initialized, system self-drives through task hand-offs. \\
  \hline
  Tool/Plugin integration
   & Built-in plugins for web search, web browsing, file I/O, and code execution; extensible via custom plugin development.
   & Extensive support for tools/APIs. Pre-built connectors for search, databases, and custom APIs.
   & Custom tools per agent role; integrates with LangChain tool ecosystem and MCP servers.
   & Focuses on inter-agent communication and artifact generation. Limited general external API/plugin integration by default. \\
  \hline
  Scalability \& deployment
   & Intended as a personal agent. Local process loops on API calls. Manual parallelism required for scaling; lacks built-in monitoring or failover.
   & Library-level scalability in user-managed infrastructure. No inherent multi-agent coordination; developer-managed deployment.
   & Library-level deployment; lightweight runtime suitable for both prototypes and production workloads on cloud infrastructure.
   & Research prototype. Resource-intensive multi-agent setup. Not optimized for high-throughput tasks; requires significant computational resources. \\
  \hline
  Task orchestration
   & Single-agent loop with GPT-driven planning and execution via ReAct-like feedback.
   & Developer-managed orchestration via chains or custom control flows in code.
   & Orchestrator agent assigns tasks to role-specialised agents and aggregates their outputs; sequential or hierarchical workflows.
   & Role-based pipeline using a publish/subscribe message bus. SOP-driven handoff between specialized agents. \\
  \hline
  MCP/A2A integration
   & Experimental community efforts for A2A; no official support yet. Potential for MCP tool invocation via JSON-RPC.
   & Supports emerging standards. Listed as A2A partner. Potential MCP client for tool calls through generic tool integration interface.
   & MCP support via community connectors; A2A integration on roadmap.
   & Implements a custom message-passing protocol analogous to A2A. Potential future integration with MCP/A2A for greater interoperability. \\
  \hline
 \end{tabularx}
\end{table*}

\section*{List of Acronyms}
\label{app:acronyms}

\begin{IEEEdescription}[\IEEEsetlabelwidth{GraphRAG}\IEEEusemathlabelsep]
\item[A2A] Agent-to-Agent Protocol.
\item[AAIF] Agentic AI Foundation.
\item[AAMAS] Autonomous Agents and Multi-Agent Systems.
\item[ACL] Agent Communication Language.
\item[ACM] Association for Computing Machinery.
\item[ACP] Agent Communication Protocol (IBM/Cisco; merged with A2A 2025); also Agentic Commerce Protocol (Stripe/OpenAI 2025).
\item[AG-UI] Agent-User Interaction Protocol.
\item[AISR] International AI Safety Report.
\item[ANP] Agent Network Protocol.
\item[API] Application Programming Interface.
\item[BDI] Belief-Desire-Intention.
\item[CAISI] Center for AI Standards and Innovation (NIST).
\item[DAO] Decentralized Autonomous Organization.
\item[DF] Directory Facilitator.
\item[DID] Decentralized Identifier.
\item[DOM] Document Object Model.
\item[ETHOS] Ethical Technology and Holistic Oversight System.
\item[FIPA] Foundation for Intelligent Physical Agents.
\item[FIRE] Flexible Integrated Reputation and Trust.
\item[GraphRAG] Graph-aware Retrieval-Augmented Generation.
\item[HATP] Hierarchical Agent-based Task Planner.
\item[HITL] Human-in-the-Loop.
\item[HTTP] Hypertext Transfer Protocol.
\item[JADE] Java Agent DEvelopment Framework.
\item[JSON] JavaScript Object Notation.
\item[JSON-RPC] JSON Remote Procedure Call.
\item[KQML] Knowledge Query and Manipulation Language.
\item[KYA] Know Your Agent.
\item[LLM] Large Language Model.
\item[MAS] Multi-Agent Systems.
\item[MCP] Model Context Protocol.
\item[NIST] National Institute of Standards and Technology.
\item[OWASP] Open Web Application Security Project.
\item[OWL] Web Ontology Language.
\item[P2P] Peer-to-Peer.
\item[PDDL] Planning Domain Definition Language.
\item[PRISMA] Preferred Reporting Items for Systematic Reviews and Meta-Analyses.
\item[RAG] Retrieval-Augmented Generation.
\item[RDF] Resource Description Framework.
\item[REST] Representational State Transfer.
\item[RL] Reinforcement Learning.
\item[RPC] Remote Procedure Call.
\item[SPARQL] SPARQL Protocol and RDF Query Language.
\item[SSE] Server-Sent Events.
\item[SSI] Self-Sovereign Identity.
\item[SW] Semantic Web.
\item[SWE] Software Engineering (as in SWE-Agent, SWE-bench).
\item[TAP] Trusted Agent Protocol (Visa).
\item[UDDI] Universal Description, Discovery, and Integration.
\item[VC] Verifiable Credential.
\item[W3C] World Wide Web Consortium.
\item[WebMCP] Browser-native extension of the Model Context Protocol.
\item[WI] Web Intelligence.
\item[WI-IAT] Web Intelligence and Intelligent Agent Technology.
\item[WoA] Web of Agents.
\item[XPath] XML Path Language.
\item[ZKP] Zero-Knowledge Proof.
\end{IEEEdescription}

\end{document}